\documentclass{article}
\usepackage[final,main]{neurips_2026}
\usepackage[english]{babel}
\usepackage{amssymb, amsmath, amsthm}
\usepackage{hyperref}
\usepackage{xcolor}
\usepackage{nicefrac}
\usepackage[margin=1cm]{caption}
\usepackage{natbib}
\usepackage{bbm}
\usepackage[scr]{rsfso}
\usepackage{algorithm}
\usepackage{algpseudocode}
\usepackage{times}
\usepackage{graphicx}
\usepackage{stackrel}
\usepackage{cases}
\usepackage[shortlabels]{enumitem}

\renewcommand{\le}{\leqslant}
\renewcommand{\leq}{\leqslant}
\renewcommand{\ge}{\geqslant}
\renewcommand{\geq}{\geqslant}

\renewcommand{\tilde}{\widetilde}
\renewcommand{\hat}{\widehat}

\newcommand{\eps}{\varepsilon}
\newcommand{\ups}{\upsilon}

\newcommand{\dd}{{\mathrm{d}}}
\newcommand{\Var}{{\rm Var}}

\newcommand{\polylog}{{\rm polylog}}

\newcommand{\sfD}{\mathsf{D}}

\newcommand{\sfp}{\mathsf{p}}
\newcommand{\sfP}{\mathsf{P}}
\newcommand{\sfq}{\mathsf{q}}

\newcommand{\NN}{\mathsf{NN}}

\newcommand{\E}{\mathbb E}
\newcommand{\N}{\mathbb N}
\newcommand{\p}{\mathbb P}
\newcommand{\R}{\mathbb R}

\newcommand{\1}{\mathbbm 1}

\newcommand{\bk}{{\mathbf k}}

\newcommand{\cA}{{\mathcal{A}}}

\newcommand{\cD}{{\mathcal{D}}}

\newcommand{\cF}{{\mathcal{F}}}

\newcommand{\cH}{{\mathcal{H}}}
\newcommand{\cJ}{{\mathcal{J}}}
\newcommand{\cK}{{\mathcal{K}}}
\newcommand{\cL}{{\mathcal{L}}}

\newcommand{\cN}{{\mathcal{N}}}
\newcommand{\cO}{{\mathcal{O}}}

\newcommand{\cT}{{\mathcal{T}}}

\newcommand{\floor}[1]{{\lfloor #1 \rfloor}}

\def\argmin{\operatornamewithlimits{argmin}}

\newcommand{\relu}{\mathrm{ReLU}}

\newcommand{\KL}{{\mathrm{KL}}}

\newcommand{\sdata}{\sigma_{\mathrm{data}}}

\newcommand{\integral}[1]{\int\limits_{#1}}

\newtheorem{Th}{Theorem}[section]
\newtheorem{Lem}[Th]{Lemma}
\newtheorem{Def}[Th]{Definition}

\newtheorem{As}[Th]{Assumption}

\newcommand{\myendproof}{\hfill$\square$}

\title{On approximation and estimation of Schrödinger potentials in a low-dimensional data model}

\author{
Artsiom Patarusau\\
MBZUAI, HSE University
\And
Nikita Puchkin\\
HSE University
\And
Konstantin Yakovlev\\
HSE University
}

\date{}

\begin{document}

\maketitle

\begin{abstract}
    We examine generative modelling approaches based on the construction of Schr\"odinger bridges between Gaussian noise and a target distribution. It is known that the solution of the dynamic Schr\"odinger problem is a diffusion process with a drift associated with Doob's h-transform of a Schr\"odinger potential. Although its accurate restoration from finite samples is crucial for reliable, high-quality data generation, the existing literature lacks theoretical guarantees regarding this question. In our work, we establish theoretical upper bounds on the complexity of Schr\"odinger potential approximation and estimation via neural networks. These bounds are mainly determined by the effective dimension of the target distribution. To our knowledge, this is the first result demonstrating that generative modelling methods based on Schr\"odinger bridges and stochastic optimal control remain consistent when the ambient dimension grows almost as fast as $\sqrt{n}$.
\end{abstract}

\section{Introduction}

Denoising diffusion \citep{ho20,song21} and flow-based \citep{liu23,lipman23} models showed themselves as a powerful tool for modelling complex multimodal distributions. The idea of using diffusion processes for data generation and translation was further developed in approaches based on Schr\"odinger bridges \citep{debortoli21, shi23, korotin24, gushchin24} and stochastic optimal control (SOC) \citep{domingoenrich25,rapakoulias25,puchkin25}, which have recently gained much popularity. In general,
the Schr\"odinger bridge between two probability distributions $\sfP_0$ and $\sfP_T$ on $\R^D$ is a stochastic process $\{X_t^* : 0 \leq t \leq T\}$ such that $X_0^* \sim \sfP_0$, $X_T^* \sim \sfP_T$, and $X_t$ deviates from a given base (or reference) process $\{X_t^0 : 0 \leq t \leq T\}$ as less as possible (in terms of the Kullback-Leibler divergence between the corresponding path measures). In what follows, we will assume that the source and the target distributions are absolutely continuous with the densities $\sfp_0$ and $\sfp_T$, respectively, and $\{X_t^0 : 0 \leq t \leq T\}$ is the Ornstein-Uhlenbeck (OU) process
\begin{equation}
    \label{eq:base_process}
    \dd X_t^0 = -b X_t^0 \dd t + \sqrt{\gamma} \, \dd W_t,
    \quad 0 < t < T,
    \quad X_0 \sim \sfp_0.
\end{equation}
Here $b > 0$ and $\gamma > 0$ are some parameters and $W_t$ is the standard Brownian motion in $\R^D$.
Such a choice of the reference process is motivated by exponentially fast mixing and a recent paper \citep{puchkin26} demonstrating advantages of the OU process over the standard scaled Wiener process $\{\sqrt{\gamma} \; \dd W_t : 0 \leq t \leq T\}$ in unconditional generation and data-to-data translation tasks.
It is known that (see, for instance, Theorem 2.4 in \cite{leonard13}) the joint density of $X_0^*$ and $X_T^*$ is given by 
\begin{equation}
    \label{eq:coupling}
    \pi^*(x, y) = \nu_0^*(x) \, \sfq_T(y \,\vert\, x) \, \nu_T^*(y),
\end{equation}
where
\begin{equation}
    \label{eq:markov_kernel}
    \sfq_t(y \, |\, x) = (\sqrt{2\pi \gamma} \sigma_t)^{-D} \exp\left\{-\frac{\|y - e^{- b t}x\|^2}{2\gamma\sigma_t^2}\right\},
    \quad \sigma_t^2 = \frac{1 - e^{-2bt}}{2b},
\end{equation}
is the transition density of the base process and $(\nu_0^*, \nu_T^*)$ is a uniquely determined (up to positive multiplicative constants) pair of $\sigma$-finite measures on $\R^D$ referred to as \emph{Schr\"odinger potentials}. The right potential $\nu_T^*$ plays a crucial role in generative modelling, since the solution of the Schr\"odinger bridge problem (as well as the optimally controlled process in the SOC approach) admits a representation
\[
    \dd X_t^* = -b X_t^* \, \dd t + \gamma \, \nabla \log \cT_{T - t}[\nu_T^*](X_t^*) \, \dd t + \sqrt{\gamma} \, \dd W_t,
    \quad 0 < t < T,
\]
where $\cT_t$ stands for the Ornstein-Uhlenbeck operator. For any measurable function $f$ on $\R^D$, it is defined as
\begin{equation}
    \label{eq:ou_operator}
    \cT_t[f](x) = \integral{\R^D} f(y) \, \sfq_t(y \, | \, x) \, \dd y,
    \quad x \in \R^D.
\end{equation}
Unfortunately, as it will become clear a bit later, 
theoretical study of $\nu_T^*$ remains quite limited.

\paragraph{Related work.}
The factorization \eqref{eq:coupling} is the core of  the Sinkhorn algorithm, which was extensively studied in the literature. Some significant results on its rates of convergence appeared in the last years \citep{conforti23, chizat24, chiarini24}. However, in generative modelling, the target density $\sfp_T$ is accessible only through i.i.d. samples $Y_1, \dots, Y_n$, while \cite{conforti23, chizat24, chiarini24} assume that the source and the target distributions are known. One of the ways to overcome this issue is to apply the Sinkhorn algorithm to the empirical measure associated with the training sample, as, for instance, in \cite{pooladian25}. This approach has an evident drawback that the estimated potential (or the corresponding drift) requires additional smoothing. In \cite{pooladian25}, the authors use nonparametric kernel-type estimates for this purpose, which usually suffer from the curse of dimensionality.

Several recent papers \citep{korotin24, puchkin25, puchkin26, belomestny26} suggested procedures for direct estimation of the continuous potential $\nu_T^*$ or a related quantity $\sfp_T / \nu_T^*$ based on empirical risk minimization. The framework of \citep{puchkin25, belomestny26} naturally incorporates neural networks, which proved their ability to adapt to the intrinsic data structure and to succeed in high-dimensional practical tasks. In addition, these methods directly optimize the corresponding objectives, avoiding splitting into smaller subproblems, which is usual for iterative proportional \citep{debortoli21, vargas21, chen22} or Markovian \citep{chen23, peluchetti23b} fitting.

\paragraph{Rates of convergence and the curse of dimensionality.}
The ERM-based approach \citep{korotin24, puchkin25, belomestny26} is supported by rigorous theoretical study. In \cite{korotin24}, \cite{puchkin25}, and \cite{puchkin26}, the authors elaborate on the statistical error and report their results in terms of the excess risk and complexity of the learnable class of potentials, leaving analysis of the approximation error out of the scope.
This gap was recently addressed by \cite{belomestny26}, who studied approximation of the ratio $\sfp_T / \nu_T^*$ in the $L^2(\sfp_T)$ norm.
However, our work focuses on approximating $\log \nu_T^*$ with respect to KL divergence rather than the ratio itself, addressing a distinct problem.
To our knowledge, \cite{belomestny26} were the first ones who derived non-asymptotic $\smash{\cO(\log^D(n) / \sqrt{n})}$ rates of convergence. Unfortunately, the unfavourable dependence on the ambient dimension $D$ is inevitable, unless we impose additional assumptions on $\sfp_T$. While the ability of deep generative models to adapt to the low intrinsic dimension of the underlying distribution and to escape the curse of dimensionality was extensively studied in the context of denoising diffusion models \citep{chen23, tang24, azangulov24, yakovlev25, yakovlev25b}, \emph{there are no results on approximation or estimation of Schr\"odinger potentials leveraging the data structure}. The present paper aims to make the first step towards filling this gap.

\paragraph{Contributions.}
We consider an affine-subspace model, which has recently got considerable attention in the context of score estimation and generative diffusion models \citep{sasaki16, chen23, oko23}.
Our main contributions are following.
\begin{itemize}
    \item We prove that (Theorem \ref{th:approx_main}), for any $\eps > 0$, there is a feedforward neural network with polylogarithmic in $\eps^{-1}$ number of weights approximating the right log-potential $\log \nu_T^*$, such that the KL divergence between the corresponding endpoint marginals does not exceed $\eps$.
    \item We establish an $\cO(\mathrm{polylog}(n)/n)$ rate of convergence for the problem of density estimation in the affine-subspace model (in terms of the KL divergence, see Theorem \ref{th:estim_main}), thereby substantially improving over the $\cO(n^{-2/(d+5)})$ bound of~\cite{chen23} and the $\cO(n^{-1/2})$ rate of \cite{pooladian25}. In addition, our bound does not depend on the stopping time provided that the underlying data distribution possesses intrinsic variability.
    \item The proofs of Theorems \ref{th:approx_main} and \ref{th:estim_main} rely on sharp bounds on the $\|\cdot\|_\infty$-norms of higher-order derivatives of $\log \sfp_T$ and $\log \nu_T^*$ (see Lemmata \ref{lem:log_density_der} and \ref{lem:log_int_der_gaus}), which could be of independent interest.
\end{itemize}

\paragraph{Notation.}
We use the following notation throughout the paper. For a vector $v$ and a tensor $A$ with real entries, $\|v\|_\infty$ and $\|A\|_\infty$ denote the maximum absolute value of their entries. Similarly, for a vector $v$ and a matrix $A$, the number of non-zero entries is denoted by $\|v\|_0$ and $\|A\|_0$, respectively.
We often replace $\max{\{a, b}\}$ and $\min{\{a, b}\}$ by shorter $a \vee b$ and $a \wedge b$, respectively. 
The operation $\otimes$ stands for the outer product.
For a function $f : \Omega \to \R$, $\|f\|_{L^\infty(\Omega)} = \sup_{x \in \Omega} |f(x)|$. For $s \geq 1$, the $L_s(\sfp)$-norm of a measurable function $f$ is $\|f\|_{L_s(\sfp)} = (\int |f(x)|^s \sfp(x)\, \dd x)^{1/s}$. For two probability densities $\sfp$ and $\sfq$ on $\R^d$, such that $\sfp \ll \sfq$, the Kullback–Leibler (KL) divergence is $\KL(\sfp, \sfq) = \int \sfp(x) \log(\sfp(x)/\sfq(x)) \, \dd x$.
For any $s \geq 1$, the Orlicz $\psi_s$-norm of a random variable is defined as
\begin{align*}
    \|\xi\|_{\psi_s} = \inf \left\{u > 0 \, : \, \E e^{|\xi|^s/u^s} \leq 2 \right\} .
\end{align*}
For a random vector $X \in \R^d$, we define $\|X\|_{\psi_2} = \sup_{v \in S^{d-1}} \|\langle v, X \rangle\|_{\psi_2}$, where $S^{d-1} = \{v \in \R^d : \|v\| = 1\}$ is the Euclidean unit sphere.
For some $\Omega \subseteq \R^r$, we define the class of $\beta$-Hölder functions with parameter $H$ as
\begin{align*}
    \cH^\beta(\Omega, H) = \left\{f : \Omega \to \R : \sum_{|\bk| < \beta} \|\partial^\bk f\|_{L^\infty(\Omega)} + \sum_{|\bk| = \floor{\beta}} \sup_{x \not= y}\frac{|\partial^\bk f(x) - \partial^\bk f(y)|}{\|x - y\|_\infty^{\beta - \floor{\beta}}} \leq H \right\} .
\end{align*}
Finally, $f \lesssim g$ and $g \gtrsim f$ are equivalent to $f = \cO(g)$, and $f \asymp g$ means that $f \lesssim g \lesssim f$.

\paragraph{Paper structure.}
The rest of the paper is organized as follows. Section~\ref{sec:prelim} introduces the Schrödinger bridge problem, our notation, and the function classes we will use. Section~\ref{sec:approx_main} states the main approximation result and its statistical corollary, including a comparison with~\cite{chen23}. Section~\ref{sec:limitations} discusses limitations and an extension to factorisable density models. Section~\ref{sec:proof_insights} provides high-level insights of our proof strategy. The proofs of all auxiliary results are deferred to the appendix.

\section{Problem setup}
\label{sec:prelim}

We consider a problem of unconditional data generation from Gaussian noise. That is, given i.i.d. samples $Y_1, \dots, Y_n$ drawn according to an unknown probability density $\sfp_T$ over $\R^D$, our goal is to produce a new sample from approximately the same distribution. We use the Schr\"odinger bridge framework for this purpose. Let $\sfp_0$ be the density of the standard Gaussian distribution $\cN(0, I_D)$ in $\R^D$. Given a reference stochastic process $\{X_t^0 : 0 \leq t \leq T\}$ obeying the dynamics \eqref{eq:base_process}, we aim to construct $\{X_t^* : 0 \leq t \leq T\}$ of the form
\[
    \dd X_t^* = -b X_t^* \, \dd t + u(X_t^*, t) \, \dd t + \sqrt{\gamma} \, \dd W_t,
    \quad 0 < t < T,
\]
such that $X_0^* \sim \sfp_0$, $X_T^* \sim \sfp_T$, and the Kullback-Leibler divergence between the path measures of $\{X_t^* : 0 \leq t \leq T\}$ and $\{X_t^0 : 0 \leq t \leq T\}$ is as small as possible. It is known that the optimal $u(x, t)$ is given by
\begin{equation}
    \label{eq:optimal_control}
    u(x, t) = \gamma \, \nabla \log \cT_{T - t}[\nu_T^*](x),
\end{equation}
where $\nu_T^*$ and $\cT_{T - t}$ stand for the right Schr\"odinger potential and the Ornstein-Uhlenbeck operator defined in \eqref{eq:coupling} and \eqref{eq:ou_operator}, respectively. Let us note that $u(x, t)$ from \eqref{eq:optimal_control} also solves a stochastic optimal control problem \cite{daipra91}.

In our paper, we are interested in quantitative bounds on complexity of approximation and estimation of $\nu_T^*$ in the presence of low-dimensional data structure via neural networks, in particular, via feedforward neural networks with ReLU activation functions. Let us recall that the ReLU activation is given by $\relu(x) = 0 \vee x$.
For a vector $b \in \R^r$, we define the shifted activation function $\relu_b(x) = \smash{\big( \relu(x_1 - b_1), \dots, \relu(x_r - b_r) \big)}^\top$, where $x \in \R^r$.
Given the number of layers $L \in \N$ and an architecture vector $W = (W_0, \dots, W_L) \in \R^{L + 1}$, a feedforward neural network $f : \R^{W_0} \to \R^{W_L}$ is a composition of linear and nonlinear transforms of the form
\begin{align}
\label{eq:f_nn_def}
    f(x) = -b_L + A_L \circ \relu_{b_{L - 1}} \circ A_{L - 1} \circ \ldots \circ A_2 \circ \relu_{b_1} \circ A_1 \circ x, 
    \quad x \in \R^{W_0} .
\end{align}
Here, $A_j \in \R^{W_j \times W_{j - 1}}$ is a weight matrix and $b_j \in \R^{W_j}$ is a bias vector for each $j \in \{1, \dots, L\}$.
Finally, we introduce the class of feedforward neural networks with at most $S \in \N$ non-zero weights, maximum width $\|W\|_\infty \in \N$, and weight magnitudes bounded by $B > 0$ as follows:
\begin{align*}
    \NN(L, W, S, B) = \left\{f \text{ as in } \eqref{eq:f_nn_def} : \sum_{j = 1}^L \big( \|A_j\|_0 + \|b_j\|_0 \big) \leq S,
    \; \max_{1 \leq j \leq L} \big\{ \|A_j\|_\infty \vee \|b_j\|_\infty \big\} \leq B \right\} .
\end{align*}
Once we constructed $\widetilde \nu_T$ approximating $\nu_T^*$, we can consider the corresponding diffusion
\[
    \dd \widetilde X_t = -b \widetilde X_t \, \dd t + \gamma \, \nabla \log \cT_{T - t}[\widetilde \nu_T](\widetilde X_t) \, \dd t + \sqrt{\gamma} \, \dd W_t,
    \quad 0 < t < T,
\]
where $\widetilde X_0 \sim \sfp_0$. We measure the quality of $\widetilde\nu_T$ with the Kullback-Leibler divergence between the corresponding endpoint marginals of $X_T^*$ and $\widetilde X_T$. Let us note that, in view of \eqref{eq:coupling}, it holds that
\[
    \sfp_0(x)
    = \int\limits_{\R^D} \pi^*(x, y) \, \dd y
    = \nu_0^*(x) \int\limits_{\R^D} \sfq_T(y \,\vert\, x) \, \nu_T^*(y) \, \dd y
    = \nu_0^*(x) \, \cT_{T}[\nu_T^*](x)
\]
and then
\[
    \sfp_T(y)
    = \int\limits_{\R^D} \pi^*(x, y) \, \dd x
    = \nu_T^*(y) \int\limits_{\R^D} \sfq_T(y \,\vert\, x) \, \nu_0^*(x) \, \dd x
    = \nu_T^*(y) \int\limits_{\R^D} \frac{\sfq_T(y \,\vert\, x)}{\cT_{T}[\nu_T^*](x)} \, \sfp_0(x) \, \dd x.
\]
Similarly, the marginal density of $\widetilde X_T$ equals to
\begin{align}
\label{eq:cand_pot_margin_dens}
    \widetilde \sfp_T(y) = \widetilde \nu_T(y) \int\limits_{\R^D} \frac{\sfq_T(y \,\vert\, x)}{\cT_{T}[\widetilde \nu_T](x)} \, \sfp_0(x) \, \dd x.
\end{align}
Hence, the performance of $\widetilde \nu_T$ is measured with
\begin{align*}
    \KL(\sfp_T, \widetilde \sfp_T)
    &
    = \int\limits_{\R^D} \log\left(\frac{\nu_T^*(y)}{\widetilde \nu_T(y)}\right) \sfp_T(y) \, \dd y
    + \int\limits_{\R^D} \log\left( \; \int\limits_{\R^D} \frac{\sfq_T(y \,\vert\, x)}{\cT_{T}[\nu_T^*](x)} \, \sfp_0(x) \, \dd x \right)\sfp_T(y)\,\dd y
    \\&\quad
    - \int\limits_{\R^D} \log\left( \; \int\limits_{\R^D} \frac{\sfq_T(y \,\vert\, x)}{\cT_{T}[\widetilde \nu_T](x)} \, \sfp_0(x) \, \dd x \right)\sfp_T(y)\,\dd y.
\end{align*}
In the next section, we address the following question.
\begin{quote}
    \emph{
    \textbf{Q1}:
    Given a precision parameter $\eps > 0$, how large should be the parameters $L$, $W$, $S$, $B$ of an approximating neural network to ensure that $\KL(\sfp_T, \widetilde\sfp_T) \leq \eps$?
    }
\end{quote}

Besides approximation guarantees, we are interested in non-asymptotic rates of convergence for Schr\"odinger potential estimation. In \cite{puchkin25}, the authors studied theoretical properties an empirical risk minimizer
\begin{equation}
    \label{eq:erm}
    \log \widehat\nu_T \in \argmin\limits_{\psi \in \Psi} \left\{ -\frac1n \sum\limits_{i = 1}^n \psi(Y_i) - \frac1n \sum\limits_{i = 1}^n \log\left( \; \int\limits_{\R^D} \frac{\sfq_T(Y_i \,\vert\, x)}{\cT_{T}[e^\psi](x)} \, \sfp_0(x) \, \dd x \right)  \right\}  ,
\end{equation}
where $Y_1, \dots, Y_n \sim \sfp_T$ are i.i.d. samples and $\Psi$ is a reference class of potentials. In particular, they proved an oracle inequality for the excess risk of $\widehat\nu_T$ and specified the estimation error. However, they left the question of approximation out of the scope of their paper. This brings us to the following question.
\begin{quote}
    \emph{
    \textbf{Q2}:
    Given a training sample of size $n$, what are the rates of convergence of the estimate \eqref{eq:erm}?
    }
\end{quote}

\section{Main results}
\label{sec:approx_main}

In this section, we present our main theoretical results on approximation and estimation of the right Schrödinger log-potential. We introduce the following assumption on the data distribution, which exhibits an underlying low-dimensional structure.

\begin{As}
\label{asn:linear_subspace}
    The data distribution $Y \sim \sfp_T$ is generated by the following model:
    \begin{align*}
        Y = G Z + \sdata \xi,
    \end{align*}
    where $\sdata \in (0, 1)$ and $G \in \R^{D \times d}$ has orthonormal columns $G^\top G = I_d$.
    The random elements $Z \sim \mu$ and $\xi \sim \cN(0, I_D)$ are independent, with $Z \in \R^d$ being centered and sub-Gaussian with $\|Z\|_{\psi_2} \leq \tau < +\infty$.
\end{As}

Assumption \ref{asn:linear_subspace} means that the data distribution is concentrated near $d$-dimensional unknown linear subspace, where the \emph{effective dimension} $d$ can be significantly smaller than the ambient dimension $D$. Similar models were considered in recent papers on score estimation and denoising diffusion models \citep{sasaki16, chen23, oko23}. The parameter $\sdata > 0$ measures intrinsic data variability.
In practice, it is unlikely that data points lie exactly on a low-dimensional subspace or a manifold but they may concentrate around a low-dimensional set. In many practical settings access to noise-free measurements is either expensive or impossible, see, for instance, \cite{daras24}. Our model therefore includes an irreducible noise and still preserves the intrinsic low‑dimensional structure of the data. In addition, the requirement $\sdata > 0$ is crucial for the stochastic optimal control approach to generative modelling, because, according to \cite{daipra91}, the marginal endpoint distribution of the base process $\{X_t^0 : 0 \leq t \leq T\}$ must dominate
the target density $\sfp_T$.

In view of Assumption \ref{asn:linear_subspace}, we have that the target density admits a decomposition
\begin{align}
\label{eq:p_T_decomp}
    \log \sfp_T(y) = \log \sfp_{||}(G^\top y) - \frac{D - d}{2} \log(2\pi\sdata^2) - \frac{\|(I_{D} - GG^\top)y\|^2}{2\sdata^2},
\end{align}
where the on-support density is given by
\begin{align}
\label{eq:rho_par_def}
    \sfp_{||}(u) &= (2\pi \sdata^2)^{-d/2} \integral{\R^d} \exp\left\{-\frac{\|u - z\|^2}{2\sdata^2}\right\} \dd \mu(z) ,
    \quad u\in \R^d .
\end{align}
It is important to note that we impose mild conditions on the mixing measure $\mu$. For instance, \cite{chen23} requires that $\nabla \log \sfp_{||}(u)$ must be globally Lipschitz. This yields that the log-density Hessian $\nabla^2 \log \sfp_{||}(u)$ must have a uniformly bounded operator norm. Direct computations show that
\begin{align}
    \label{eq:on-support_log-density_hessian}
    \nabla^2 \log \sfp_{||}(u)
    &= -\frac{I_d}{\sdata^2}
    + \frac{1}{2\sdata^4 Z^2(u)}
      \integral{\R^d}\integral{\R^d}
      (z_1-z_2)(z_1-z_2)^\top \notag\\
    &\qquad\cdot
      \prod_{j=1}^2
      \exp\left\{\frac{u^\top z_j}{\sdata^2}
                  -\frac{\|z_j\|^2}{2\sdata^2}\right\}
      \,\dd\mu(z_1)\,\dd\mu(z_2),
\end{align}
where
\[
    Z(u)=\integral{\R^d}
    \exp\left\{\frac{u^\top z}{\sdata^2}
                  -\frac{\|z\|^2}{2\sdata^2}\right\}\,\dd\mu(z).
\]
While the right-hand side of \eqref{eq:on-support_log-density_hessian} is obviously bounded on $\R^d$ in the case when $\mu$ is Gaussian or has a bounded support, it is not clear, whether this claim holds for a general sub-Gaussian measure $\mu$. For this reason, Assumption \ref{asn:linear_subspace} allows us to consider a more general framework compared to \cite{chen23}.

We also impose a mild assumption on the potential $\nu_T^*$ assuming that it is bounded from above.

\begin{As}
\label{asn:bounded_log_pot}
    There exists a constant $M \geq 1$ such that $\log \nu_T^*(y) \leq M$ for all $y \in \R^D$.
    In addition, it holds that $\cT_\infty[\log \nu_T^*] = 0$.    
\end{As}

Note that Assumption \ref{asn:bounded_log_pot} is consistent with the analysis of Schrödinger potential estimation in \cite{puchkin25} and \cite{puchkin26}. The requirement $\cT_\infty[\log \nu_T^*] = 0$ is not restrictive at all. Let us remind a reader that the potentials $\nu_0^*$ and $\nu_T^*$ are defined up to multiplicative constant. Hence, the condition $\cT_\infty[\log \nu_T^*] = 0$ should be considered as a normalization.

We now establish that, under the assumptions formulated above, the complexity of the right Schrödinger log-potential approximation is mainly determined by the effective dimension $d$, rather than by the ambient one.
The key observation is that the log-potential admits the following decomposition.

\begin{Lem}[log-potential decomposition]
\label{lem:log_pot_decomp}
    Assume $\sfp_T$ satisfies Assumption \ref{asn:linear_subspace}, and let $\sfp_0$ be the standard Gaussian density on $\R^D$. Let $\rho_{||}(u)$ and $\sfq_{||}(u \,|\, x)$ be the densities of the Gaussian distributions $\cN(0, I_d)$ and $\cN(e^{-bT} x, \gamma \sigma_T^2 I_d)$, where $\sigma_T^2$ is defined in \eqref{eq:markov_kernel}, respectively. For any measurable function $f: \R^d \rightarrow \R$, introduce
    \begin{equation}
        \label{eq:on-support_ou}
        \cT_{||}[f](x) = \int\limits_{\R^d} f(u) \sfq_{||}(u \,|\, x) \, \dd u,
    \end{equation}
    and let $\nu_{||}^* : \R^d \rightarrow (0, +\infty)$ be a solution of the equation
    \begin{equation}
        \label{eq:log_pot_equation}
        \log \nu_{||}^*(u)
        = \log \sfp_{||}(u) - \log \integral{\R^d} \frac{\sfq_{||}(u \, | \, x) \rho_{||}(x) \, \dd x}{\cT_{||}[\nu_{||}^*](x)},
        \quad u \in \R^d,
    \end{equation}
    such that $\int_{\R^d} \log \nu_{||}^*(u) e^{-\|u\|^2 / (2\gamma \sigma_{\infty}^2)} \, \dd u = 0$.
    Then, the log-potential $\log \nu_T^*$ admits a decomposition
    \[
        \log \nu_T^*(y) = \log \nu_{||}^*(G^\top y) - \frac{\|(I_D - GG^\top)y\|^2 - \gamma\sigma_{\infty}^2(D - d)}{2} \left( \frac{1}{\sdata^2 - \sigma^2} - \frac{1}{\gamma \sigma_T^2} \right) ,
    \]
    where $\sigma_\infty^2 = (2b)^{-1}$ and $\sigma$ is the positive root of the quadratic equation $\gamma\sigma_T^2 \sigma = e^{-bT}(\sdata^2 - \sigma^2)$.
    Furthermore, $\cT_\infty[\log \nu_T^*] = 0$.

\end{Lem}

The proof of Lemma \ref{lem:log_pot_decomp} is postponed to Appendix \ref{sec:lem_log_pot_decomp_proof}.
We would like to note that $\smash{\nu_{||}^*}$ is well-defined, since it is the right Schr\"odinger potential for the marginals $\rho_{||}(x)$, $\sfp_{||}(u)$ and the Markov kernel $\sfq_{||}(u \,|\, x)$.
Importantly, Lemma \ref{lem:log_pot_decomp} shows that the approximation of the right Schrödinger log-potential $\log \nu_T^*$ can be achieved by approximating $\log \nu_{||}^*$, a $d$-dimensional function, thus avoiding the curse of dimensionality associated with the ambient $D$-dimensional space.
Nevertheless, existing approximation results from the literature do not apply to $\log \nu_{||}^*$.
Overcoming this obstacle is the main technical challenge of the present paper. When $D>d$, the restriction of $\log\nu_T^*$ to the orthogonal complement of $\operatorname{Im}(G)$ is bounded from above if and only if
$\sdata^2\leq\gamma\sigma_T^2+e^{-2bT}$.
Indeed, the quadratic coefficient is non-negative precisely when
$\sigma\leq e^{-bT}$, which is equivalent to this inequality by the equation defining $\sigma$.
This is a necessary condition for Assumption \ref{asn:bounded_log_pot}
when $D>d$, and it can be ensured by a suitable choice of the reference
parameters $\gamma$, $b$, and $T$.

\begin{Th}
\label{th:approx_main}
    Grant Assumptions \ref{asn:linear_subspace} and \ref{asn:bounded_log_pot} and assume that $\sfp_0$ is the standard Gaussian density on $\R^D$.
    Let us fix an arbitrary $\eps \in (0,1)$ and suppose that
    \begin{equation}
    \label{eq:kl_bT_cond}
        bT \gtrsim M + \log\left( M + \log\frac{e}\eps \right) + \log\log\frac{e}{\sdata^{2}}.
    \end{equation}
    Then there exists an approximation $\tilde\nu_T$ of the form
    \[
        \log\tilde\nu_T(y)=\log\tilde\nu_{||}(G^\top y)
        -\frac{\|(I_D-GG^\top)y\|^2-\gamma\sigma_\infty^2(D-d)}{2}
        \left(\frac1{\sdata^2-\sigma^2}-\frac1{\gamma\sigma_T^2}\right),
    \]
    where $\sigma$ is the positive root of $\gamma\sigma_T^2\sigma=e^{-bT}(\sdata^2-\sigma^2)$ and $\sigma_\infty^2=(2b)^{-1}$, such that
    \[
        \cT_\infty[\log\tilde\nu_{||}]=0,
        \quad -A(1+\|z\|^2)\leq\log\tilde\nu_{||}(z) \leq 2M,
        \quad z\in\R^d,
    \]
    with $A\asymp\sdata^{-2}$, and the corresponding terminal density satisfies $\KL(\sfp_T,\tilde\sfp_T)\leq\eps$.
    Moreover, $\log\tilde\nu_{||}\in\NN(L,W,S,B)$ with
    \[
        L\vee\log B
        \lesssim \left( M + \log\frac{e}\eps \right)^2 \log\left( \frac{e M}{\sdata^{2}} + \frac{e}{\sdata^{2}} \log\frac{e}\eps \right),
    \]
    and
    \[
        \|W\|_\infty\vee S
        \lesssim \sdata^{-2d} \left( M + \log\frac{e}\eps \right)^{3d+5} \log^{d+1}\left( \frac{e M}{\sdata^{2}} + \frac{e}{\sdata^{2}} \log\frac{e}\eps \right).
    \]
    All hidden constants depend only on $\gamma,b,\tau$, and $d$.
\end{Th}

The proof of Theorem \ref{th:approx_main} is postponed to Appendix \ref{sec:th_approx_main_proof}.
Theorem \ref{th:approx_main} establishes that the right Schrödinger log-potential $\log \nu_T^*$ can be approximated within the accuracy $\eps$ in terms of the KL divergence between the corresponding endpoint marginals via a quite small neural network with only \smash{$\cO\big(\polylog(1/\eps) \big)$} weights. It is a consequence of the favourable bounds on the higher-order derivatives of the log-potential (see Lemmata \ref{lem:log_density_der} and \ref{lem:log_int_der_gaus}).
To the best of our knowledge, Theorem \ref{th:approx_main} provides the first approximation result of such kind.
A very recent paper \cite{belomestny26} considered a related problem but focused on the approximation of $\sfp_T / \nu_T^*$ in the $L^2(\sfp_T)$ norm. Our goal is different.
We also would like to note that the condition \eqref{eq:kl_bT_cond} should be considered as a condition on the product $bT$, rather than on the time horizon $T$. We still can take $T = 1$ but it will require larger values of $b$.

We proceed with theoretical guarantees on accuracy of estimation of the potential $\nu_T^*$ from $n$ i.i.d. samples $Y_1, \dots, Y_n$ drawn according to $\sfp_T$. Lemma \ref{lem:log_pot_decomp} and Theorem \ref{th:approx_main} suggest us to use a class of neural networks of the following form.

\begin{Def}
\label{def:log_pot_class}
We denote by $\cF_\NN(L, W, S, B, \sigma_{\min})$ the class of neural networks of the form
\begin{align*}
    \psi(y) = \varphi(V^\top y) - a(\sigma) \left( \|(I_D - VV^\top) y\|^2 - \gamma\sigma_\infty^2(D - d) \right),
\end{align*}
where $\varphi \in \NN(L, W, S, B)$, the matrix $V \in \R^{D \times d}$ has orthonormal columns, $0 < \sigma_{\min} \leq \sigma \leq 1$ for some $\sigma_{\min} \leq \sdata$,
\begin{align*}
    a(\sigma) = \frac{1}{2}\left(\frac{1}{\sigma^2 - \tilde\sigma^2} - \frac{1}{\gamma\sigma_T^2}\right),
\end{align*}
and $\tilde \sigma$ is the positive root to the equation $\gamma\sigma_T^2 \tilde\sigma = e^{-bT}(\sigma^2 - \tilde\sigma^2)$.
We further assume that
\begin{align*}
    -A(1 + \|z\|^2) \leq \varphi(z) \leq 2M,
    \quad \text{for all } z \in \R^d ,
\end{align*}
where the constant $A \geq 1$ is sufficiently large and satisfies $A \lesssim \sigma_{\min}^{-2}$. The hidden constant behind $\lesssim$ depends on $\gamma, b, \tau$, and $d$ only.
Moreover, we suppose that $\cT_\infty[\varphi] = 0$.
\end{Def}

In Definition \ref{def:log_pot_class}, since $\sdata$ is unknown, we make it learnable and use the decomposition given by Lemma \ref{lem:log_pot_decomp}.
Crucially, we assume that we have an access to a lower bound to $\sdata$, which we denote by $\sigma_{\min} > 0$.
Moreover, since the linear subspace given by $G$ from Assumption \ref{asn:linear_subspace} is also unknown, we introduce a learnable matrix $V$ with orthonormal columns.
Finally, we emphasize that the log-potential $\psi$ satisfies the normalization constraint $\cT_\infty[\psi] = 0$.
The closed interval for $\sigma$ ensures that the empirical minimum is attained.
We now formulate our main result on the estimation of the Schrödinger potential.

\begin{Th}
\label{th:estim_main}
    Assume the conditions of Theorem \ref{th:approx_main}. Suppose that $bT \gtrsim \log\log n + M + D\sigma_{\min}^{-2}$.
    Then there are such
    \[
        L \vee\log B \lesssim \big( M + \log(en) \big)^2 \log\left( \frac{eD (M + \log(en)) }{\sigma_{\min}^{2}} \right)
    \]
    and
    \[
        \|W\|_\infty\vee S
        \lesssim\sigma_{\min}^{-2d} \big( M + \log(en) \big)^{3d+5} \log^{d+1}\left( \frac{eD (M + \log(en)) }{\sigma_{\min}^{2}} \right)
    \]
    that, for any $\delta \in (0, 1)$, the density $\widehat \sfp_T$, associated with an empirical risk minimizer $\log \widehat\nu_T$ (see \eqref{eq:erm}) over the class $\Psi = \cF_\NN(L, W, S, B, \sigma_{\min})$ (see Definition \ref{def:log_pot_class}), satisfies
    \begin{align*}
        \KL(\sfp_T, \widehat\sfp_T)
        &
        \lesssim
        \frac{D (M + D \sigma_{\min}^{-2}) \log^2(en)\log(e/\delta)}{n\sigma_{\min}^{2d}}
        \\&\quad
        \cdot \big( M + \log(en) \big)^{3d + 9} \log^{d+3} \left( \frac{eD (M + \log(en)) }{\sigma_{\min}^{2}} \right)
        \\&
        = \tilde\cO\left(\frac{D M^{3d + 9} \log(e / \delta)}{n \sigma_{\min}^{2d}} \left(M + \frac{D}{\sigma_{\min}^{2}} \right) \right)
    \end{align*}
    with probability at least $(1 - \delta)$.
    Here, the hidden constants depend on $\gamma$, $b$, $\tau$ and $d$ only.
    
\end{Th}

The proof of Theorem \ref{th:estim_main} is moved to Appendix \ref{sec:th_estim_main_proof}.
Let us elaborate on comparison of Theorem \ref{th:estim_main} with relevant results in the existing literature.
By a straightforward application of the Girsanov theorem,
Theorem 2 in \cite{chen23} yields that
\begin{align}
\label{eq:chen_kl_bound}
    \E [\KL(\sfp_{T - t_0}, \widehat\sfp_{T - t_0})] = \tilde\cO\left(\frac{1}{t_0}\left(n^{-\frac{2}{d + 5}} + D n^{-\frac{d + 3}{d + 5}}\right)\right),
\end{align}
where $t_0$ is a stopping time.
By contrast, Theorem \ref{th:estim_main} significantly improves the dependence on the sample size.
The underlying reason is the approximation theory developed in Theorem \ref{th:approx_main}.
Our construction guarantees that the number of non-zero parameters scales polylogarithmically with $\varepsilon^{-1}$, thereby enabling the faster convergence rate in the sample size.
While the results of \cite{yakovlev25} extend beyond linear subspace assumption of \cite{chen23}, their bound with the smoothness parameter $\beta \asymp \log n$ does not recover the \smash{$\widetilde\cO(n^{-1})$} convergence rate.
This is due to the fact that the dependence on the ambient dimension $D$ is worse than in Theorem \ref{th:estim_main}.
Moreover, the bounds \eqref{eq:chen_kl_bound} and those of \cite{yakovlev25} explode as $t_0 \to 0$.
The same issue appears in the analysis of diffusion and flow-based generative models \citep{oko23,yakovlev25,fukumizu25}, even when the target distribution is absolutely continuous.
On the other hand, the bound established in Theorem \ref{th:estim_main} circumvents this limitation and eliminates the need for stopping time $t_0$.
The dependence on the noise level appears only through \smash{$\sigma_{\min}^2$}, which plays a role analogous to $t_{0}$.

Notably, \cite{pooladian25} obtained that expected squared total variation distance between the path measures, corresponding to $\{X_t^* : 0 \leq t \leq T\}$ and the process with the estimated drift $\{\widehat X_t : 0 \leq t \leq T\}$, scales as $\cO(n^{-1/2} + n^{-1}(1 - \tau)^{-k - 2})$, where $1 - \tau$ denotes the stopping time and $k$ represents the dimension of the manifold supporting the target distribution.
Our Theorem \ref{th:estim_main} correctly captures the dependence on $\sigma_{\min}$ through the effective dimension $d$.
Notably, unlike \cite{pooladian25}, our result avoids the slow $n^{-1/2}$ convergence rate, achieving improved statistical efficiency.

The above findings indicate possible gaps in theoretical guarantees for generative diffusion models and flow-based models.
We suppose that these problems are primarily related to the properties of the corresponding loss functions (in denoising score matching and flow matching), which blow up as $t_0$ approaches zero, rather than to the properties of the score function and the velocity field, which may remain bounded.
We emphasize that this comparison does not imply that the approaches, relying on construction of Schr\"odinger bridges or stochastic optimal control, inherently outperform diffusion- and flow-based methods.
Our analysis focuses solely on upper bounds, while lower bounds remain an open question in this context.

\section{Extensions and future directions}
\label{sec:limitations}

In view of \eqref{eq:p_T_decomp}, the target density $\sfp_T$ can be represented as a product of two probability densities on orthogonal spaces. Such a model is quite popular in statistics in the context of non-Gaussian component analysis and linear dimension reduction (see, for instance, \cite{blanchard06a} and the references therein). It has a natural extension when $\sfp_T$ is represented as a product of $J \geq 2$ components: 
\begin{equation}
    \label{eq:fac_den_model}
    \sfp_T(y) = \prod_{j=1}^J \sfp^{(j)}_T(S_j y), \quad y \in \R^D,
\end{equation}
where the matrices $S_j \in \R^{d_j \times D}$, $j \in \{1, \ldots, J\}$, $d_1 + \ldots + d_J = D$, are such that
\[
    S_j S_j^\top = I_{d_j}
    \quad \text{for every $1 \leq j \leq J$}
    \quad \text{and} \quad
    S_j S_k^\top = 0 \quad \text{for all $j \neq k$},
\]
and, for each $j \in \{1, \ldots, J\}$, $\sfp^{(j)}_T$ is a probability density on $\R^{d_j}$. In this case, it is also possible to approximate and estimate the Schr\"odinger potential $\nu_T^*$, because it is also factorizable. We provide the corresponding result below.

\begin{Lem}
\label{lem:factorizable_log_pot_decomp}
    Assume that $\sfp_T$ admits a decomposition \eqref{eq:fac_den_model}, and let $\sfp_0$ be the density of $\cN(0, I_D)$.
    For each $j \in \{1, \ldots, J\}$, let \smash{$\rho^{(j)}(\tilde y_j)$ and $\sfq_T^{(j)}(\tilde y_j \, | \, \tilde x_j)$} denote the densities of the Gaussian distributions $\cN(0, I_{d_j})$ and $\cN(e^{-bT} \tilde x_j, \gamma\sigma_T^2 I_{d_j})$, respectively.
    Suppose that for every $j \in \{1, \ldots, J\}$ the solution of the Schr\"odinger subproblem on $\R^{d_j}$ with initial density \smash{$\rho^{(j)}$, target density $\sfp^{(j)}_T$, and transition kernel $\sfq_T^{(j)}$} exists and let \smash{$(\nu_0^{(j)}, \nu_T^{(j)})$} be the corresponding potentials.
    After a suitable choice of the multiplicative normalizations of the subproblem potentials, the Schr\"odinger log-potentials admit the decompositions
    \begin{align}
    \label{eq:decomposition_main}
        \log \nu_0^*(x) = \sum_{j=1}^J \log \nu_0^{(j)}(S_j x)
        \quad \text{and} \quad
        \log \nu_T^*(y) = \sum_{j=1}^J \log \nu_T^{(j)}(S_j y),
        \quad x, y \in \R^D .
    \end{align}
\end{Lem}
The proof of Lemma \ref{lem:factorizable_log_pot_decomp} is postponed to Appendix \ref{sec:lem_factorizable_log_pot_decomp_proof}.
Further extensions of Assumption \ref{asn:linear_subspace} can include situation when the support of the mixing measure $\mu$ is a union of affine spaces \citep{chakraborty26} or nonlinear \cite{yakovlev25}. The nonlinear model with low intrinsic dimension has a practical impact, because real-world data sets often exhibit nonlinear underlying structures \citep{brown23}.

\section{Proof Insights}
\label{sec:proof_insights}

In this section, we provide some intuition behind the proofs of our main results, Theorems \ref{th:approx_main} and \ref{th:estim_main}. Lemma \ref{lem:log_pot_decomp} essentially reduces the problem to approximation and estimation of the on-support log-potential \smash{$\log \nu_{||}^*$} defined in \eqref{eq:log_pot_equation}. We rely on a well-known result of \cite{schmidt2020} (Lemma \ref{lem:smidt_hieber}) to approximate the summands in the right-hand side of \eqref{eq:log_pot_equation} via feedforward neural networks with ReLU activations. However, the most challenging and the most technical part of the proof is to provide sharp bounds on $\|\cdot\|_\infty$-norms of higher-order derivatives of
\[
    \log \sfp_{||}(u)
    \quad \text{and} \quad
    \log\integral{\R^d}\frac{\sfq_{||}(u\,|\,x)\rho_{||}(x)\,\dd x}{\cT_{||}[\nu_{||}^*](x)}.
\]
We provide the corresponding results below and believe that they could be of independent interest.
\begin{Lem}
\label{lem:log_density_der}
    Assume that the target distribution $\sfp_T$ satisfies Assumption~\ref{asn:linear_subspace}, and let $\sfp_{||}$ be the on-support density defined in \eqref{eq:rho_par_def}. Then, for every integer $k \geq 2$ and every $u \in \R^d$,
    \begin{align*}
        \left\| \nabla^k \left( \log \sfp_{||}(u) + \frac{\|u\|^2}{2\sdata^2}\right)\right\|_\infty
        \leq \left( \frac{36 (\|u\| + 2\tau \sqrt{d} + \sdata) \sqrt{C}}{\sdata^2} \right)^k k!,
    \end{align*}
    where $C \geq 1$ is an absolute constant.
\end{Lem}
\begin{Lem}
\label{lem:log_int_der_gaus}
    Let $\rho_{||}$ and $\sfq_{||}(\cdot \,|\, x)$ be the densities of $\cN(0, I_d)$ and $\cN(e^{-bT} x, \gamma \sigma_T^2 I_d)$, respectively. Let $\nu_{||} : \R^d \to (0,\infty)$ be measurable and satisfy $\cT_\infty[\log\nu_{||}] = 0$ and $\log\nu_{||} \leq M$, and assume that $bT$ obeys
    \begin{align}
    \label{eq:der_bt_cond}
        8 e^{2 - bT} \leq \gamma \sigma_T^2 \sqrt{d}
        \quad \text{and} \quad
        14e^2\sqrt{d} \, e^{-bT/2} \leq \log 2 .
    \end{align}
    Then, for every integer $k \geq 3$ and every $y \in \R^d$,
    \begin{align*}
        \left\| \nabla^k \log \integral{\R^d} \frac{\sfq_{||}(y \,|\, x)\rho_{||}(x) \, \dd x}{\cT_{||}[\nu_{||}](x)} \right\|_\infty
        \leq \exp\!\left\{\!k(2M+3) + \frac{k\,e^{-2bT}\|y\|^2}{2(\gamma\sigma_T^2)^2}\!\right\}\!
        \left( \frac{3 \cdot 2^{2 + d/2}\, e^{-bT}\sqrt{C}}{\gamma\sigma_T^2}\right)^{\!\!k}\! k! ,
    \end{align*}
    where $\cT_{||}$ is defined in \eqref{eq:on-support_ou} and $C \geq 1$ is an absolute constant.
\end{Lem}
The proofs of Lemmata \ref{lem:log_density_der} and \ref{lem:log_int_der_gaus} are moved to Appendices \ref{sec:lem_log_density_der_proof} and \ref{sec:lem_log_int_der_gaus_proof}, respectively.
The main obstacle is that the density $\rho_{||}(x)$ as well as the mixing measure $\mu$ have unbounded supports. This makes the proof significantly more complicated compared to, for instance, \citep{yakovlev25} (Lemma 4.1). In the proofs of Lemmata \ref{lem:log_density_der} and \ref{lem:log_int_der_gaus}, we derive explicit expressions for the $k$-th order derivatives and reduce the problem to upper bounds on the expectations of homogeneous shift-invariant polynomials of sub-Gaussian random variables of degree $k$ (see Lemma \ref{lem:gaus_expect}). The proof of Lemma \ref{lem:log_density_der} significantly exploits the following fact from probability theory, which, to our knowledge, was not presented in the literature before.

We would like to relate the results of Lemmata \ref{lem:log_density_der} and \ref{lem:log_int_der_gaus} to those in the literature. 
In view of \eqref{eq:log_pot_equation}, Lemma \ref{lem:log_density_der}, and Lemma \ref{lem:log_int_der_gaus}, we obtain bounds on higher-order derivatives of $\log \nu_{||}(u)$.
This problem also arises in entropic optimal transport.
In particular, \citet[Proposition 1]{genevay19} derives bounds on derivatives of the Scr\"odinger potentials when the source and target distributions are compactly supported. In \citet[Proposition 1]{mena19}, the authors extend their result to the case of sub-Gaussian marginals.
Similarly to \cite{mena19}, we consider unbounded supports, but, unlike the aforementioned works, we carefully track the dependence on the derivative order.

\begin{Lem}
    \label{lem:cond_density_bound}
    Let $X$ and $Y$ be independent random vectors such that $X \sim \cN(0, \sigma^2 I_d)$, $\sigma > 0$, and $\E Y = 0$ and $\E \|Y\|^2 \le d\tau^2$. Set $S = X + Y$. Then, for any $s \in \R^d$, the conditional distribution of $X$ given $S = s$ is sub-Gaussian. Moreover, its conditional $\psi_2$-norm and the conditional expectation $\mu(s) = \E[X \mid S = s]$ satisfy the inequalities
    \begin{align*}
        \|X - \mu(s)\|_{\psi_2(\cdot \mid S=s)} \le 6R + 6\sigma
        \quad \text{and} \quad
        \|\mu(s)\| \le 3R + 3\sigma,
    \end{align*}
    where $R = \|s\| + 2 \tau \sqrt{d}$.
\end{Lem}
We provide the proof of Lemma \ref{lem:cond_density_bound} in Appendix \ref{sec:lem_cond_density_bound_proof}. Notably, the random vector $Y$ in Lemma \ref{lem:cond_density_bound} is not necessarily sub-Gaussian, it should just have a finite second moment.
We would like to note that similar result appears in the entropic optimal transport literature.
Specifically, \citet[Lemma 2]{pooladian24} show that the random vector $X - \mu(s)$ is sub-Gaussian when $S = s$ is fixed.
However, in contrast to us, the authors do not track how the $\psi_2$-norm scales with the problem parameters.

Lemmata \ref{lem:log_density_der}, \ref{lem:log_int_der_gaus}, and \ref{lem:cond_density_bound} are the main ingredients of the proof of Theorem \ref{th:approx_main}. Once we quantified the complexity of Schr\"odinger potential approximation, the proof of Theorem \ref{th:estim_main} is not very hard. It relies on technical lemmata from \cite{puchkin25} and on the localization argument for unbounded empirical processes developed in \cite{yakovlev25b}. Combination of this results with Theorem \ref{th:approx_main} yields the desired rates of convergence.

\begin{ack}
    The study was implemented in the framework of the Basic Research Program at HSE University (HSE-BR-2026-002).
\end{ack}

\bibliographystyle{abbrvnat}
\bibliography{references}

\newpage
\appendix

\section{Proofs of the results from Section \ref{sec:approx_main}}

\subsection{Proof of Lemma \ref{lem:log_pot_decomp}}
\label{sec:lem_log_pot_decomp_proof}

If $d=D$, the perpendicular component is absent and the claim follows
directly from \eqref{eq:log_pot_equation}. In the remainder of the proof,
assume that $d<D$.

First, observe that the density of $X_T$ satisfies
\begin{align*}
    \log \sfp_T(y) + \frac{D}{2} \log(2\pi \sdata^2)
    =
    -\frac{\|(I_D - GG^\top)y\|^2}{2\sdata^2} + \log \integral{\R^d} \exp\left(-\frac{\|G^\top y - u\|^2}{2\sdata^2}\right) \, \dd\mu(u) .
\end{align*}
We now show that the Ornstein-Uhlenbeck operator also admits a similar decomposition.
Let $G_\perp \in \R^{D\times(D-d)}$ be the matrix that completes $G$ to an
orthonormal basis of $\R^D$.
Let us find the log-potential in the form $\log \nu_T^*(y) = \log \nu_{||}^*(G^\top y) + \log \nu_\perp^*(G_\perp^\top y)$.
Then, it holds that
\begin{align*}
    \cT_T[\nu_T^*](x)
    &=\E_{\xi\sim\cN(e^{-bT}x,\gamma\sigma_T^2I_D)}
       [\nu_{||}^*(G^\top\xi)\nu_\perp^*(G_\perp^\top\xi)]\\
    &=\E_\xi[\nu_{||}^*(G^\top\xi)]\,
      \E_\xi[\nu_\perp^*(G_\perp^\top\xi)],
\end{align*}
where we used the fact that $G^\top \xi$ and $G_\perp^\top \xi$ are independent.
Consequently, defining
\begin{align*}
    \cT_{||}[\nu_{||}](u) &= \E_{\xi \sim \cN(e^{-bT}u, \sigma_T^2 \gamma I_d)}[\nu_{||}(\xi)],
    \qquad u \in \R^d,
    \\
    \cT_\perp[\nu_\perp](v) &= \E_{\xi \sim \cN(e^{-bT}v, \sigma_T^2 \gamma I_{D - d})} [\nu_\perp(\xi)],
    \qquad v \in \R^{D - d},
\end{align*}
we conclude that
\begin{align*}
    \cT_T[\nu_T^*](x) = \cT_{||}[\nu_{||}^*](G^\top x) \cdot \cT_\perp[\nu_\perp^*](G_\perp^\top x) .
\end{align*}
We now define transition densities
\begin{align*}
    \log \sfq_{||}(u' \, | \, u)
    &= -\frac{d}{2}\log(2\pi\gamma\sigma_T^2) -\frac{\|u'- e^{-bT} u\|^2}{2\gamma\sigma_T^2},
    \quad u', u \in \R^d,
    \\
    \log \sfq_\perp(v' \, | \, v)
    &= -\frac{D - d}{2} \log(2\pi\gamma\sigma_T^2) - \frac{\|v' - e^{-bT}v\|^2}{2\gamma\sigma_T^2},
    \quad v', v \in \R^{D - d} .
\end{align*}
Let us also define the standard Normal density in $\R^d$ and $\R^{D - d}$ as $\rho_{||}$ and $\rho_\perp$ respectively.
Thus, it follows that
\begin{align*}
    \log \integral{\R^D} \frac{\sfq_T(y \, | \, x) \sfp_0(x) \, \dd x}{\cT_T[\nu_T^*](x)}
    = \log \integral{\R^d} \frac{\sfq_{||}(G^\top y \, | \, x) \rho_{||}(x) \, \dd x}{\cT_{||}[\nu_{||}^*](x)}
    + \log \integral{\R^{D - d}} \frac{\sfq_\perp(G_\perp^\top y \, | \, x) \rho_\perp(x) \, \dd x}{\cT_\perp[\nu_\perp^*](x)} .
\end{align*}
Therefore, we conclude that the potentials $\nu_{||}^*$ and $\nu_\perp^*$ can be found from the corresponding Schrödinger systems:
\begin{align*}
    \log \nu_{||}^*(z)
    &= -\frac{d}{2}\log(2\pi\sdata^2) + \log \integral{\R^d} \exp\left(-\frac{\|z - u\|^2}{2\sdata^2}\right) \, \dd \mu(u)
    - \log \integral{\R^d}\frac{\sfq_{||}(z \, | \, x) \rho_{||}(x) \, \dd x}{\cT_{||}[\nu_{||}^*](x)},
    \\
    \log \nu_\perp^*(v) &= -\frac{D - d}{2}\log(2\pi\sdata^2) - \frac{\|v\|^2}{2\sdata^2}
    - \log \integral{\R^{D - d}} \frac{\sfq_\perp(v \, | \, x) \rho_\perp(x) \, \dd x}{\cT_\perp[\nu_\perp^*](x)} .
\end{align*}
Here, $z \in \R^d$ and $v \in \R^{D - d}$.
Note that $\nu_\perp^*$ solves the Schrödinger problem between two Gaussian distributions $\cN(0, I_{D - d})$ and $\cN(0, \sdata^2 I_{D - d})$.
This problem has a closed-form solution, which we formalize in the following lemma.

\begin{Lem}
\label{prop:sch_pot_gaus_gaus}
    Let the base process be given by \eqref{eq:base_process}
    Let also $\sfp_0 = \cN(0, I_D)$ and $\sfp_T = \cN(\mu_T, \sdata^2 I_D)$ for some $\mu_T \in \R^D$ and $\sdata>0$.
    Then, the coupling between $X_0^*$ and $X_T^*$ in the optimally controlled process has the form
    \begin{align*}
        \begin{pmatrix}
            X_0^* \\ X_T^*
        \end{pmatrix}
        \sim \cN\left(
        \begin{pmatrix}
            0 \\ \mu_T
        \end{pmatrix},
        \begin{pmatrix}
            I_D & \sigma I_D \\ \sigma I_D & \sdata^2 I_D
        \end{pmatrix}
        \right),
        \quad \frac{\sigma}{\sdata} = \frac{-1 + \sqrt{1 + 4C^2}}{2C},
        \quad C = \frac{\sdata e^{-bT}}{\gamma \sigma_T^2} .
    \end{align*}
    In addition, for all $x, y \in \R^D$, the Schrödinger potentials are given by
    \begin{align*}
        &\log \nu_T^*(y) = -\frac{\|y\|^2}{2}\left( \frac{1}{\sdata^2 - \sigma^2} - \frac{1}{\gamma \sigma_T^2} \right) + \frac{y^\top \mu_T}{\sdata^2 - \sigma^2} 
    \end{align*}
    and
    \begin{align*}
        \log \nu_0^*(x) &= -\frac{\|x\|^2}{2}\left(\frac{\sdata^2 - e^{-bT}\sigma}{\sdata^2 - \sigma^2}\right) - \frac{x^\top \mu_T \, \sigma}{\sdata^2 - \sigma^2}
        - \frac{\|\mu_T\|^2}{2(\sdata^2 - \sigma^2)}
        \\&\quad
        -\frac{D}{2} \log\left(\frac{2\pi(\sdata^2 - \sigma^2)}{\gamma \sigma_T^2}\right) .
    \end{align*}
    
\end{Lem}

The proof of Lemma \ref{prop:sch_pot_gaus_gaus} is postponed to Appendix \ref{sec:prop_sch_pot_gaus_gaus_proof}.
Therefore, applying Lemma \ref{prop:sch_pot_gaus_gaus}, we obtain the explicit form of $\log \nu_\perp^*$:
\begin{align*}
    \log \nu_\perp^*(v) = -\frac{\|v\|^2 - \gamma\sigma_\infty^2(D - d)}{2} \left(\frac{1}{\sdata^2 - \sigma^2} - \frac{1}{\gamma \sigma_T^2}\right) .
\end{align*}
This log-potential satisfies the normalization condition $\cT_\infty[\log \nu_\perp^*] = 0$.
Finally, we observe that $\|G_\perp^\top y\|^2 = \|(I_D - G G^\top) y\|^2$, for all $y \in \R^D$.
The proof is now complete.
    
\myendproof

\subsection{Proof of Theorem \ref{th:approx_main}}
\label{sec:th_approx_main_proof}

Throughout this proof, constants depend only on $\gamma,b,\tau$, and $d$.
By Lemma \ref{lem:potential_lower_bound}, for a sufficiently large $A\asymp\sdata^{-2}$,
\begin{equation}
    \label{eq:approx_log_paral_bounds}
    -A(1+\|z\|^2)\leq\log\nu_{||}^*(z)\leq M,
    \quad \cT_\infty[\log\nu_{||}^*]=0.
\end{equation}
We use the following approximation result.

\begin{Lem}
    \label{lem:approx_log_pot_compact}
    Grant Assumptions \ref{asn:linear_subspace} and \ref{asn:bounded_log_pot}, with $M\geq1$, and let $\sfp_0$ be the standard Gaussian density.
    Fix $R\geq e$ and suppose that
    \begin{equation}
        \label{eq:approx_compact_bT_cond}
        bT\gtrsim M+\log R.
    \end{equation}
    Let $\operatorname{clip}_R$ denote coordinatewise clipping to $[-R,R]^d$.
    For every $\eta\in(0,1)$ there exists $\tilde f\in\NN(L,W,S,B)$ such that
    \begin{equation*}
        \left\|\tilde f - \log\nu_{||}^* \circ \operatorname{clip}_R  \right\|_{L^\infty(\R^d)} \leq \eta.
    \end{equation*}
    Its architecture satisfies
    \[
        L\vee\log B\lesssim\log^2(e/\eta)\log(\sdata^{-2}R),
        \quad
        \|W\|_\infty\vee S\lesssim
        \log^{2d+5}(e/\eta)(R^2\sdata^{-2})^d\log(\sdata^{-2}R).
    \]
\end{Lem}

The proof is postponed to Appendix \ref{sec:lem_approx_log_pot_compact_proof}.
Choose
\begin{align}
\label{eq:approx_eta_def}
    \eta \asymp \eps\exp\{-C(1+M)\},
\end{align}
where $C>0$ is sufficiently large.
Set
\begin{equation}
\label{eq:approx_R_def}
    R^2 \asymp \log\left(\frac{e(M + A)}{\eta}\right),
\end{equation}
with the hidden constant sufficiently large, so that $R\geq e$.
Then
\[
    \log\frac{e}\eta \lesssim M + \log\frac{e}\eps,
    \quad
    R^2\lesssim \left( M + \log\frac{e}\eps \right) \log\left( \frac{e M}{\sdata^{2}} + \frac{e}{\sdata^{2}} \log\frac{e}\eps \right),
\]
and
\[
    \log(\sdata^{-2}R)\lesssim \log\left( \frac{e M}{\sdata^{2}} + \frac{e}{\sdata^{2}} \log\frac{e}\eps \right).
\]
Condition \eqref{eq:kl_bT_cond} therefore implies \eqref{eq:approx_compact_bT_cond}.
Let us apply Lemma \ref{lem:approx_log_pot_compact} with these $R$ and $\eta$.
Since $\|\operatorname{clip}_R x\|\leq\|x\|$, we have that
\[
    -A(1 + \|x\|^2) - \eta
    \leq \tilde f(x)
    \leq M + \eta
    \quad \text{ for all $x \in \R^d$.}
\]
Both the distribution with density $\sfp_{||}$ and the stationary Gaussian distribution have sub-Gaussian tails.
Thus, by the Cauchy-Schwarz inequality, the Gaussian concentration bound \citep[Theorem 1.19]{rigollet2023}, \eqref{eq:approx_log_paral_bounds}, and \eqref{eq:approx_R_def},
\begin{align*}
    \|\tilde f-\log\nu_{||}^*\|_{L^1(\sfp_{||})}
    &\lesssim \|\tilde f - \log \nu_{||}^*\|_{L^\infty([-R, R]^d)} + \|\tilde f - \log \nu_{||}^*\|_{L^2(\sfp_{||})} \sqrt{\p_{\xi \sim \sfp_{||}}(\|\xi\|_\infty \geq R)}
    \\& \lesssim \eta + (M + A) \exp\{-c R^2\}
    \lesssim \eta .
\end{align*}
Similarly, we deduce that $\cT_\infty[|\tilde f-\log\nu_{||}^*|] \lesssim \eta$.
Define the normalized approximation by
\[
    \log\tilde\nu_{||}=\tilde f-\cT_\infty[\tilde f],
\]
which leaves the associated terminal density unchanged.
The change in the last bias is at most $\cO(\eta)$, by \eqref{eq:approx_log_paral_bounds}.
Consequently,
\begin{equation}
    \label{eq:norm_tilde_pot_cond}
    \|\log\tilde\nu_{||}-\log\nu_{||}^*\|_{L^1(\sfp_{||})}
    +\cT_\infty[|\log\tilde\nu_{||}-\log\nu_{||}^*|]
    \lesssim\eta\leq1.
\end{equation}
By reducing the absolute constant in the definition of $\eta$, and by increasing once and for all the constant $A$, we also obtain
\begin{equation*}
    -A(1+\|x\|^2)\leq\log\tilde\nu_{||}(x)\leq 2M,
    \quad \text{ for all } x \in \R^d.
\end{equation*}
Next, observe that
\begin{align*}
    \KL(\sfp_T, \tilde\sfp_T)
    &\leq \integral{\R^D}\sfp_T(y) \left| \log\nu_T^*(y) - \log\tilde\nu_T(y) \right| \, \dd y
    \\&\quad
    + \integral{\R^D}\sfp_T(y) \left| \log \integral{\R^D}\frac{\sfq_T(y \, | \, x) \sfp_0(x) \, \dd x}{\cT_T[\nu_T^*](x)} - \log \integral{\R^D}\frac{\sfq_T(y \, | \, x) \sfp_0(x) \, \dd x}{\cT_T[\tilde\nu_T](x)} \right| \, \dd y .
\end{align*}
In view of Lemma \ref{lem:log_pot_decomp}, we have that
\begin{align*}
    \KL(\sfp_T, \tilde\sfp_T)
    &\leq \integral{\R^d} \sfp_{||}(z) \left| \log\nu_{||}^*(z) - \log\tilde\nu_{||}(z) \right| \, \dd z
    \\&\quad
    + \integral{\R^d} \sfp_{||}(z) \left| \log\integral{\R^d}\frac{\sfq_{||}(z \, | \, u)\rho_{||}(u) \, \dd u}{\cT_{||}[\nu_{||}^*](u)} - \log\integral{\R^d}\frac{\sfq_{||}(z \, | \, u)\rho_{||}(u) \, \dd u}{\cT_{||}[\tilde\nu_{||}](u)} \right| \, \dd z .
\end{align*}
Now we need the following technical result, which compares the log-integrals generated by two potentials.

\begin{Lem}[\cite{puchkin25}, Lemma B.2, corrected]
\label{lem:log_integr_diff_aux}
    Let $f_0,f_1:\R^d\to\R$ satisfy $\cT_\infty[f_i]=0$ and
    \[
        -A\|x\|^2-B\leq f_i(x)\leq M,
        \quad i\in\{0,1\},\quad x\in\R^d,
    \]
    where $A\geq0$, $M\geq0$, and $B\geq M\vee1$.
    If $bT\geq(5+\log d)\vee\log(160b/\gamma)$, then
    \begin{align*}
        &\left|\log\int_{\R^d}\frac{\sfq_{||}(y\mid x)\rho_{||}(x)}{\cT_{||}[e^{f_0}](x)}\,\dd x
        -\log\int_{\R^d}\frac{\sfq_{||}(y\mid x)\rho_{||}(x)}{\cT_{||}[e^{f_1}](x)}\,\dd x\right|\\
        &\quad\lesssim
        \bigl(\cT_\infty[|f_1-f_0|]\bigr)^{1/\cK(T)}
        (d^2+\|y\|^2)^{1-1/\cK(T)}\\
        &\qquad\cdot\exp\left\{\cO\!\left(d+M+
        \bigl(M+\log(e+A+B)\bigr)\sqrt d\,e^{-bT}
        +e^{-bT}(1+\|y\|^2)\right)\right\}.
    \end{align*}
    Here $\cK(T)$ is defined in \eqref{eq:ck}, and the constants depend only on $\gamma$ and $b$.
\end{Lem}

Applying Lemma \ref{lem:log_integr_diff_aux} with upper bound $2M$ and with $B=A\vee2M$, we obtain that
\begin{align*}
    \KL(\sfp_T, \tilde\sfp_T)
    &
    \lesssim \|\log \nu_{||}^* - \log \tilde\nu_{||}\|_{L^1(\sfp_{||})}
    \\&\quad
    + \left(\cT_\infty[|\log \nu_{||}^* - \log \tilde\nu_{||}|]\right)^{1 / \cK(T)}
    \\&\quad
    \cdot \exp\left\{\cO(M + (M + \log (e + A + M)) e^{-bT})\right\}
    \\&\quad
    \cdot \E \left[(1 + \|\xi\|^2)^{1 - 1 / \cK(T)} \exp\{\cO( e^{-bT}(1 + \|\xi\|^2)) \} \right] ,
\end{align*}
where $\xi \sim \sfp_{||}$.
Condition \eqref{eq:kl_bT_cond}, the bound \eqref{eq:cK_T_bound}, and the sub-Gaussian tails of $\sfp_{||}$ yield
\[
    \KL(\sfp_T,\tilde\sfp_T)
    \lesssim \eta + \exp\{\cO(M)\}\eta^{1/\cK(T)}.
\]
Moreover,
\[
    \left(1-\frac1{\cK(T)}\right)\log(1/\eta)
    \lesssim e^{-bT} \left( M + \log\frac{e}\eps \right)
    \lesssim1.
\]
It follows that $\eta^{1/\cK(T)}\lesssim\eta$, and, hence, $\KL(\sfp_T,\tilde\sfp_T)\leq\eps$.
Finally, substituting \eqref{eq:approx_eta_def} and \eqref{eq:approx_R_def} into Lemma \ref{lem:approx_log_pot_compact}, yields $\log \tilde\nu_{||} \in \NN(L, W, S, B)$ with
\[
    L\vee\log B
    \lesssim \left( M + \log\frac{e}\eps \right)^2 \log\left( \frac{e M}{\sdata^{2}} + \frac{e}{\sdata^{2}} \log\frac{e}\eps \right),
\]
and
\[
    \|W\|_\infty\vee S
    \lesssim \sdata^{-2d} \left( M + \log\frac{e}\eps \right)^{3d+5} \log^{d+1}\left( \frac{e M}{\sdata^{2}} + \frac{e}{\sdata^{2}} \log\frac{e}\eps \right).
\]
The proof is finished.

\myendproof

\subsection{Proof of Theorem \ref{th:estim_main}}
\label{sec:th_estim_main_proof}

The proof of Theorem \ref{th:estim_main} uses key technical lemmata from \cite{puchkin25} and employs machinery for unbounded empirical processes developed in \cite{yakovlev25b}.
First, recall from Definition \ref{def:log_pot_class} that any $\psi \in \cF_{\NN}(L, W, S, B, \sigma_{\min})$ is of the form
\begin{equation}
\label{eq:psi}
    \psi(y) = \varphi(V^\top y)
    - a(\sigma) \left( \|(I_D-VV^\top)y\|^2 - \gamma\sigma_\infty^2(D-d) \right),
\end{equation}
where $V\in\R^{D\times d}$ has orthonormal columns and $-A(1 + \|z\|^2)\leq \varphi(z) \leq 2M$, for every $z \in \R^d$.
Proceeding as in \eqref{eq:p_T_decomp} and Lemma \ref{lem:log_pot_decomp}, we deduce that
\begin{align}
\label{eq:sup_psi_log_p_decomp}
    \log \sfp_T^\psi(y) = \log \sfp_{||}^\varphi(V^\top y) - \frac{D - d}{2} \log(2\pi \sigma^2) - \frac{\|(I_D - VV^\top)y\|^2}{2\sigma^2} .
\end{align}
Here, $\log \sfp_{||}^\varphi$ is given by
\begin{align}
\label{eq:sup_psi_norm_ld_density}
    \log \sfp_{||}^\varphi(u) = \varphi(u) + \log \int \frac{\sfq_{||}(u \, | \, x) \rho_{||}(x) \, \dd x}{\cT_{||}[e^\varphi](x)},
    \quad u \in \R^d .
\end{align}
In view of \eqref{eq:cand_pot_margin_dens}, for every $\psi \in \cF_{\NN}(L, W, S, B, \sigma_{\min})$ define the probability density
\begin{align}
\label{eq:cand_pot_NN_margin_dens}
    \sfp_T^\psi(y) = \int \frac{ \sfq_T(y \, | \, x) e^{\psi(y)}}{\cT_T[e^\psi](x)} \sfp_0(x) \, dx,
    \quad y \in \R^D .
\end{align}


The following lemma controls the $\psi_1$-norm of the envelope of the log-likelihood ratios induced by $\cF_{\NN}(L, W, S, B, \sigma_{\min})$.

\begin{Lem}
\label{lem:psi_norm_sup}
Let $\cF_{\NN}$ and $\sfp_T^\psi$ be defined as in Definition \ref{def:log_pot_class} and \eqref{eq:cand_pot_NN_margin_dens}, respectively.
Assume further that $Y \sim \sfp_T$ and $bT\gtrsim1$.
Then, it follows that
\begin{align*}
    &(i) \quad \left\|\sup_{\psi\in\cF_{\NN}} \left|\log\sfp_T^\psi(Y) - \log\sfp_T(Y)\right| \right\|_{\psi_1}
    \lesssim M + D\sigma_{\min}^{-2}
    \\&
    (ii) \quad \sup_{\psi \in \cF_{\NN}} \left\{ \log \sfp_T(y) - \log \sfp_T^\psi(y) \right\} \lesssim M + D\sigma_{\min}^{-2} + \sigma_{\min}^{-2} \|y\|^2,
    \quad \text{for all $y \in \R^D$} .
\end{align*}
\end{Lem}

The proof of Lemma \ref{lem:psi_norm_sup} is deferred to Appendix \ref{sec:lem_psi_norm_sup_proof}.
This lemma allows us to apply the result of \citet[Theorem D.3]{yakovlev25b}, extending the standard localization technique \cite{bartlett05} empirical processes with bounded $\psi_1$-diameter, to the excess loss class
\begin{align*}
    \cL = \{ y \mapsto \log \sfp_T(y) - \log \sfp_T^\psi(y) + n^{-1} \, : \, \psi \in \cF_{\NN}(L, W, S, B) \}.
\end{align*}
First, we have to show that $\cL$ satisfies a Bernstein condition.

\begin{Lem}
\label{lem:bernstein_log_ratio}
Suppose that Assumptions \ref{asn:linear_subspace} and \ref{asn:bounded_log_pot} hold, that $\sfp_0$ is the standard Gaussian density on $\R^D$, and that $\cF_{\NN}(L, W, S, B, \sigma_{\min})$ is the class in Definition \ref{def:log_pot_class}. Assume also that $bT\gtrsim1$.
Then, for every integer $n \geq 2$ and every $\psi \in \cF_{\NN}(L, W, S, B, \sigma_{\min})$, it holds that
\begin{align*}
    \Var_{Y\sim\sfp_T}
    \left[\log\frac{\sfp_T(Y)}{\sfp_T^\psi(Y)}\right]
    \leq \E \left( \log\frac{\sfp_T(Y)}{\sfp_T^\psi(Y)} \right)^2
    \lesssim
    \left(M + \frac{D}{\sigma_{\min}^{2}} \right) \log n \left( \KL(\sfp_T, \sfp_T^\psi) + \frac{1}{n} \right) .
\end{align*}
The implicit constants may depend on $\gamma$, $b$, $\tau$, and $d$, but not on $n$, $bT$, $D$, or $\sigma_{\min}$.
\end{Lem}

According to Lemma \ref{lem:bernstein_log_ratio} and the Cauchy-Schwarz inequality, it holds that
\begin{align*}
    \E(\log \sfp_T(Y) - \log \sfp_T^\psi(Y) + n^{-1})^2
    &\leq 2 \, \E(\log \sfp_T^\psi(Y) - \log \sfp_T(Y))^2 + \frac{2}{n^{2}}
    \\&
    \lesssim 2 \left(M + \frac{D}{\sigma_{\min}^{2}} \right) \log n \left(\KL(\sfp_T, \sfp_T^\psi) + \frac1n \right) + \frac{2}{n^{2}},
\end{align*}
for any $\psi \in \cF_{\NN}(L, W, S, B, \sigma_{\min})$.
The proof of Lemma \ref{lem:bernstein_log_ratio} essentially relies on Lemmata C.1 and C.2 from \cite{puchkin25}.
We defer it to Appendix \ref{sec:lem_bernstein_log_ratio_proof} and aim to evaluate the covering number of the excess loss class with respect to the empirical $L^2$-norm.

\begin{Lem}
\label{lem:impr_loss_prox}
Grant Assumptions \ref{asn:linear_subspace} and
\ref{asn:bounded_log_pot}. Suppose that $\sfp_0$ is the standard Gaussian density on $\R^D$, and that $bT \gtrsim 1$.
Let $\psi_1, \psi_2 \in \cF_{\NN}(L, W, S, B, \sigma_{\min})$ (see Definition \ref{def:log_pot_class}) have the form
\begin{align*}
    \psi_j(y)
    =\varphi_j(V_j^\top y)
    -a(\sigma_j)\left(
        \|(I_D-V_jV_j^\top)y\|^2
        -\gamma\sigma_\infty^2(D-d)
    \right),\quad j\in\{1,2\} .
\end{align*}
Fix $0 < \eps < 1$ and assume that $\|V_1 - V_2\|_\infty \vee |\sigma_1^2 - \sigma_2^2| \leq \eps$.
In addition, assume that the underlying weight matrices and biases of $\varphi_1$ and $\varphi_2$ are close in the sense that
\begin{align*}
    \max_{1 \leq l \leq L} (\|A_l^{(1)} - A_l^{(2)}\|_\infty \vee \|b_l^{(1)} - b_l^{(2)}\|_\infty) \leq \eps .
\end{align*}
Put
\begin{align}
\label{eq:loss_prox_lambda_def}
    \Lambda_\varphi = (d + 1) L (B \vee 1)^{L} (\|W\|_\infty + 1)^{L} .
\end{align}
Then, for every $y \in \R^D$, it holds that
\begin{align*}
    &\left|\log \sfp_T^{\psi_1}(y) - \log \sfp_T^{\psi_2}(y) \right|
    \lesssim D(\Lambda_\varphi + \sigma_{\min}^{-4})(1 + \|y\|^2) \eps
    \\&\quad
    + \exp\{ \cO(M + e^{-bT}(M + \log(e+A+M) + \|y\|^2)) \} \Lambda_\varphi \eps^{1/2} .
\end{align*}
\end{Lem}

We provide the proof of Lemma \ref{lem:impr_loss_prox} in Appendix \ref{sec:lem_impr_loss_prox_proof}.
Since an empirical risk minimizer is selected over the class of neural networks with at most $S$ non-zero weights, we observe that there are at most
\[
    \sum_{s = 0}^S \binom{L\|W\|_\infty (\|W\|_\infty + 1)}{s}
    \leq (S + 1) L^S (\|W\|_\infty + 1)^{2S}
\]
admissible sparsity patterns.
Furthermore, we need additional $Dd + 1$ parameters to parametrize $V \in \R^{D \times d}$ and $\sigma$.
Hence, the total number of parameters is $(S + Dd + 1)$.
For each sparsity pattern, let us partition the bounded parameter set into cubes and choose one admissible representative from every nonempty cube.
This keeps the envelope and normalization constraints intact.
Lemma~\ref{lem:impr_loss_prox} then yields
\begin{align}
\label{eq:cov_num_excess_loss_aux}
    &
    \log\cN\big(\tau, \cL, L^2(\sfP_n) \big)
    \lesssim (S + Dd + 1) M
    \notag
    \\&\qquad
    + (S + Dd + 1) e^{-bT} \left(M + \log(e + A + M) + \max\limits_{1\leq i\leq n}\|Y_i\|^2 \right)
    \\&\qquad
    + (S + Dd + 1) L \log\left( \frac{eDL(\|W\|_\infty+1)(B\vee1)\sigma_{\min}^{-2}
    \big(1 + \max\limits_{1\leq i\leq n}\|Y_i\|^2 \big)}{\tau}\right).
    \notag
\end{align}
Initially this bound holds at sufficiently small radii. Increasing the constants and the random scale $\sfD_n$ below extends it to every radius up to the empirical diameter. In other words, the covering number $\cN\big(\ups, \cL, L^2(\sfP_n) \big)$ satisfies the inequality
\[
    \cN\big(\ups, \cL, L^2(\sfP_n) \big)
    \leq \cA \left( \frac{\sfD_n}\ups \right)^\zeta
    \quad \text{for all $\ups \in (0, \cD(\cL, L^2(\sfP_n))$}
\]
with
\begin{align*}
    \zeta
    &
    \asymp (S + Dd + 1) L,
    \\
    \log \cA
    &
    \lesssim (S + Dd + 1) \left[ M + L \log\left( \frac{eDL(\|W\|_\infty + 1) (B\vee1)}{\sigma_{\min}^{2}} \left( M + \frac{D}{\sigma_{\min}^{2}} \right)  \right) \right],
    \\
    \log \sfD_n
    &
    \asymp 1 + \log\left( M + \frac{D}{\sigma_{\min}^{2}} \right) + \log\left( 1 + \max\limits_{1\leq i\leq n}\|Y_i\|^2 \right)
    \\&
    \quad
    + e^{-bT} \left( M + \log(e + A + M) + \max\limits_{1\leq i\leq n}\|Y_i\|^2 \right).
\end{align*}
We choose the absolute constant in $\sfD_n$ large enough that it dominates
the empirical diameter of $\cL$. Such a choice is possible because the
quadratic bounds in Lemma~\ref{lem:psi_norm_sup} give
\[
    \cD(\cL, L^2(\sfP_n)) \lesssim \left( M + \frac{D}{\sigma_{\min}^{2}} \right) \left(1 + \max\limits_{1\leq i\leq n}\|Y_i\|^2 \right).
\]
This and Lemmata \ref{lem:psi_norm_sup} and \ref{lem:bernstein_log_ratio} ensure that the assumptions of Theorem D.3 from \cite{yakovlev25b} are satisfied.
The sub-Gaussian representation $Y = GZ + \sdata\xi$ implies
\[
    \left\|\max\limits_{1\leq i\leq n}\|Y_i\|^2 \right\|_{\psi_1}
    \lesssim (d\tau^2 + D\sdata^2)\log(en)
    \lesssim D\log(en).
\]
Furthermore, Jensen's inequality gives
$\|\log(1+\|Y\|^2)\|_{\psi_1}\lesssim\log(eD)$.
The maximal inequality in \cite[Lemma E.3]{yakovlev25b} therefore yields
\[
 \bigl(\E\log^3\sfD_n\bigr)^{1/3}
 \lesssim \log\left(eM + \frac{eD}{\sigma_{\min}^2 } \right) + \log(eD) \log(en)
 +e^{-bT} \bigl(M + \log(e+A+M) + D \log(en) \bigr).
\]
The condition on $bT$ in the theorem statement makes the last term bounded.
Due to Theorem D.3 from \cite{yakovlev25b}, we conclude that, with probability at least $1 - \delta$,
\begin{align}
\label{eq:estim_corrected_oracle}
    \KL(\sfp_T,\widehat\sfp_T)
    &\notag
    \lesssim \inf_{\psi\in\cF_\NN(L, W, S, B, \sigma_{\min})}\KL(\sfp_T,\sfp_T^\psi)
    \\&\quad\notag
    + M \left(M + \frac{D}{\sigma_{\min}^2 } \right) (S + Dd + 1) \frac{\log^2(en)}n
    \\&\quad
    + L \left(M + \frac{D}{\sigma_{\min}^2 } \right) (S + Dd + 1) \frac{\log^2(en)}n
    \\&\quad\notag
    \cdot \log\left(\frac{eDL (\|W\|_\infty + 1)(B\vee1)}{\sigma_{\min}^{2}} \left(M + \frac{D}{\sigma_{\min}^2 } \right) \right)
    \\&\quad\notag
    + \left(M + \frac{D}{\sigma_{\min}^2 } \right) \frac{\log^2(en) \log(e / \delta)}n.
\end{align}
Let us apply Theorem~\ref{th:approx_main} with accuracy $\eps = 1 / (2n)$ and replace
$\sdata$ by its lower bound $\sigma_{\min}$ in the architecture budgets.
We can choose
\[
    L \vee\log B \lesssim \big( M + \log(en) \big)^2 \log\left( \frac{eD (M + \log(en)) }{\sigma_{\min}^{2}} \right)
\]
and
\[
    \|W\|_\infty\vee S
    \lesssim \sigma_{\min}^{-2d} \big( M + \log(en) \big)^{3d+5} \log\left( \frac{eD (M + \log(en)) }{\sigma_{\min}^{2}} \right)^{d+1}
\]
in such a way that the corresponding class $\cF_\NN(L, W, S, B, \sigma_{\min})$ satisfies the inequality
\[
    \inf_{\psi\in\cF_\NN(L, W, S, B, \sigma_{\min})}\KL(\sfp_T,\sfp_T^\psi) \leq \frac1{2n}.
\]
Then
\[
    (S + Dd + 1)
    \lesssim D \sigma_{\min}^{-2d} \big( M + \log(en) \big)^{3d+5} \log\left( \frac{eD (M + \log(en)) }{\sigma_{\min}^{2}} \right)^{d + 1}
\]
and
\begin{align*}
    &
    \log\left(\frac{eDL (\|W\|_\infty + 1)(B\vee1)}{\sigma_{\min}^{2}} \left(M + \frac{D}{\sigma_{\min}^2 } \right) \right)
    \\&
    \lesssim \big( M + \log(en) \big)^{2} \log\left( \frac{eD (M + \log(en)) }{\sigma_{\min}^{2}} \right).
\end{align*}
Hence, with probability at least $(1 - \delta)$, it holds that
\begin{align*}
    \KL(\sfp_T, \widehat\sfp_T)
    &
    \lesssim
    \frac{D (M + D \sigma_{\min}^{-2}) \log^2(en)\log(e/\delta)}{n\sigma_{\min}^{2d}}
    \\&\quad
    \cdot \big( M + \log(en) \big)^{3d + 9} \log^{d+3} \left( \frac{eD (M + \log(en)) }{\sigma_{\min}^{2}} \right).
\end{align*}
\myendproof

\subsection{Proof of Lemma \ref{prop:sch_pot_gaus_gaus}}
\label{sec:prop_sch_pot_gaus_gaus_proof}

Let $\pi^*$ be the joint density of $(X_0^*, X_T^*)$.
Then, according to \cite{leonard13} and \eqref{eq:markov_kernel}, we have that, for all $x, y \in \R^D$,
\begin{align}
\label{eq:log_coupl_general}
    \notag
    \log \pi^*(x, y)
    &= \log \nu_0^*(x) + \log \sfq_T(y \, | \, x) + \log \nu_T^*(y)
    \\&
    = \log \nu_0^*(x) - \frac{D}{2} \log(2\pi\gamma\sigma_T^2) - \frac{\|y - e^{-bT}x\|^2}{2\gamma\sigma_T^2}
    + \log \nu_T^*(y) .
\end{align}
Let us verify that the form of the coupling described in claim of the proposition has the above form.
From now on, assume that $0 < \sigma < \sdata$.
Then, taking into account that
\begin{align*}
    \det
    \begin{pmatrix}
        I_D & \sigma I_D \\ \sigma I_D & \sdata^2 I_D
    \end{pmatrix}
    = \det \left( \begin{pmatrix}1 & \sigma \\ \sigma & \sdata^2\end{pmatrix} \otimes I_D \right)
    = (\sdata^2 - \sigma^2)^D ,
\end{align*}
we deduce that
\begin{align*}
    \log \pi^*(x, y) = -D \log(2\pi) - \frac{D}{2}\log(\sdata^2 - \sigma^2) - \frac{1}{2}
    \begin{pmatrix}
        x \\ y - \mu_T
    \end{pmatrix}^\top
    \begin{pmatrix}
        I_D & \sigma I_D \\ \sigma I_D & \sdata^2 I_D
    \end{pmatrix}^{-1}
    \begin{pmatrix}
        x \\ y - \mu_T
    \end{pmatrix} .
\end{align*}
Applying the Schur complement argument, we obtain that
\begin{align}
\label{eq:gaus_pi_star}
    \notag
    &\log \pi^*(x, y) = -D \log(2\pi) - \frac{D}{2}\log(\sdata^2 - \sigma^2)
    -\frac{1}{2}\|x\|^2 \left(1 - \frac{\sigma^2}{\sdata^2}\right)^{-1}
    \\&
    + x^\top (y - \mu_T) \left(1 - \frac{\sigma^2}{\sdata^2} \right)^{-1} \frac{\sigma}{\sdata^2}
    - \frac{1}{2} \|y - \mu_T\|^2 \left(\sdata^{-2} + \frac{\sigma^2}{\sdata^4} \left(1 - \frac{\sigma^2}{\sdata^2}\right)^{-1} \right) .
\end{align}
Therefore, it suffices to take $\sigma$ such that the terms with $x^\top y$ coincide.
Specifically, let
\begin{align}
\label{eq:sigma_quadr_eq}
    \left(1 - \frac{\sigma^2}{\sdata^2} \right)^{-1} \frac{\sigma}{\sdata^2} = \frac{e^{-bT}}{\gamma \sigma_T^2} .
\end{align}
Solving this quadratic equation, we conclude that
\begin{align*}
    \frac{\sigma}{\sdata} = \frac{-1 + \sqrt{1 + 4C^2}}{2C} .
\end{align*}
Note that $0 < \sigma < \sdata$ and, therefore, the first claim follows.
Equations \eqref{eq:log_coupl_general} and \eqref{eq:gaus_pi_star} determine the right Schrödinger log-potential up to an additive constant:
\begin{align*}
    \log \nu_T^*(y)
    &
    = \log \pi^*(x, y) - \log\sfq_T(y \, | \, x) - \log \nu_0^*(x)
    \\&
    \propto -\frac{\|y\|^2}{2}\left( \frac{1}{\sdata^2 - \sigma^2} - \frac{1}{\gamma \sigma_T^2} \right) + \frac{y^\top \mu_T}{\sdata^2 - \sigma^2} .
\end{align*}
Therefore, we can choose $\nu_T^*$ as shown on the right hand side of the above expression, and the second claim follows.
Then, the left Schrödinger potential is expressed as
\begin{align*}
    \log \nu_0^*(x)
    &
    = -\frac{\|x\|^2}{2}\left(\frac{\sdata^2}{\sdata^2 - \sigma^2} - \frac{e^{-2bT}}{\gamma\sigma_T^2}\right) -
    \frac{x^\top \mu_T \, \sigma}{\sdata^2 - \sigma^2}
    - \frac{\|\mu_T\|^2}{2(\sdata^2 - \sigma^2)}
    \\&\quad
    -\frac{D}{2} \log\left(\frac{2\pi(\sdata^2 - \sigma^2)}{\gamma \sigma_T^2}\right) .
\end{align*}
Consequently, using \eqref{eq:sigma_quadr_eq}, we obtain
\begin{align*}
    \log \nu_0^*(x)
    &
    = -\frac{\|x\|^2}{2}\left(\frac{\sdata^2 - e^{-bT}\sigma}{\sdata^2 - \sigma^2}\right) - \frac{x^\top \mu_T \, \sigma}{\sdata^2 - \sigma^2}
    \\&\quad
    - \frac{\|\mu_T\|^2}{2(\sdata^2 - \sigma^2)}
    -\frac{D}{2} \log\left(\frac{2\pi(\sdata^2 - \sigma^2)}{\gamma \sigma_T^2}\right) .
\end{align*}
This completes the proof.

\myendproof

\subsection{Proof of Lemma \ref{lem:approx_log_pot_compact}}
\label{sec:lem_approx_log_pot_compact_proof}
Throughout the proof, the hidden constants depend only on
$\gamma$, $b$, $\tau$, and $d$.
The implicit constant in \eqref{eq:approx_compact_bT_cond} is chosen sufficiently large.
In particular, $\gamma\sigma_T^2 \asymp 1$,
\eqref{eq:der_bt_cond} holds, $e^{2 M - bT}R \lesssim 1$.
Recall that, by Lemma \ref{lem:log_pot_decomp}, the actual
log-potential decomposes as
\begin{align*}
    \log\nu_{||}^*(u)
    &=f_1(u)-f_2(u)
      +\frac d2\log(2\pi\gamma\sigma_T^2),
    \\
    f_1(u)
    &=\log\sfp_{||}(u)
      +\frac{\|u\|^2}{2\gamma\sigma_T^2},
    \\
    f_2(u)
    &=\log\integral{\R^d}
      \exp\left\{
        \frac{e^{-bT}u^\top x}{\gamma\sigma_T^2}
        -\frac{e^{-2bT}\|x\|^2}{2\gamma\sigma_T^2}
      \right\}
      \frac{\rho_{||}(x)\,\dd x}
           {\cT_{||}[\nu_{||}^*](x)}.
\end{align*}
Consequently, it suffices to approximate $f_1$ and $f_2$ separately.

\medskip
\noindent
\textbf{Step 1: log-density approximation.}
\quad
We first approximate $f_1$ on $[-R,R]^d$.
Our proof uses the following approximation result.

\begin{Lem}[\cite{schmidt2020}, Theorem 5, adapted]
\label{lem:smidt_hieber}
    For any function $f\in \cH^\beta([0, 1]^d, H)$ and any integers $m$ and $N \geq (\beta \vee 1)^d \vee 2e^d$, there exists $\tilde f \in \NN(L, W, S, B)$ with
    \begin{align*}
        L \lesssim m \log(d + \beta),
        \quad S \lesssim (d + \beta + 1)^{d + 3} N m,
        \quad \|W\|_\infty \lesssim (d + \beta) N,
        \quad \log B \lesssim \log(H + 1),
    \end{align*}
    such that 
    \begin{align*}
        \|\tilde f - f\|_{L^\infty([0, 1]^d)}
        \lesssim (H + 1)(d^2 + \beta^2)6^d N 2^{-m} + H \, 3^\beta N^{-\beta / d} .
    \end{align*}
\end{Lem}

This version of Lemma \ref{lem:smidt_hieber} follows by applying the cited theorem to $f/(H\vee1)$ and rescaling the output.
Let $\beta \geq 2$ be an integer, which will be specified later in the proof.
For a smooth function $f$ on $[0,1]^d$, its $\beta$-H\"older norm is bounded by
\[
    C_d\binom{d+\beta}{d}
    \max_{|\bk|\leq\beta} \|\partial^\bk f\|_{L^\infty([0,1]^d)}
    \lesssim \beta^d \max_{|\bk|\leq\beta} \|\partial^\bk f\|_{L^\infty([0,1]^d)} ,
\]
where $C_d$ depends only on $d$.
The affine change of variables $u=-R+2Rx$ multiplies derivatives of order $|\bk|$ by $(2R)^{|\bk|}$.

For the derivatives of orders zero and one, we argue directly.
Since $Z$ is centered (see Assumption \ref{asn:linear_subspace}), Jensen's inequality and \eqref{eq:rho_par_def} yield
\[
    -\frac d2\log(2\pi\sdata^2)
    -\frac{\|u\|^2+\E\|Z\|^2}{2\sdata^2}
    \leq\log\sfp_{||}(u)
    \leq-\frac d2\log(2\pi\sdata^2).
\]
Using $\E\|Z\|^2\lesssim d\tau^2$ and $R\geq e$, we obtain
\begin{align}
\label{eq:approx_f1_compact}
    \|f_1\|_{L^\infty([-R,R]^d)} \lesssim\sdata^{-2}R^2.
\end{align}
Based on \eqref{eq:rho_par_def}, we have that $\nabla\log\sfp_{||}(u)=\sdata^{-2}\bigl(\E[Z\mid Z+\sdata\zeta=u]-u\bigr)$ and, by Lemma \ref{lem:cond_density_bound},
\[
    \max_{|\bk|=1}
    \|\partial^\bk f_1\|_{L^\infty([-R,R]^d)}
    \lesssim\sdata^{-2}R.
\]
For every integer $k\geq2$, Lemma \ref{lem:log_density_der} and the correction of the quadratic term in $f_1$ imply that
\[
    \max_{|\bk|=k}
    \|\partial^\bk f_1\|_{L^\infty([-R,R]^d)}
    \leq (C\sdata^{-2}R)^k k!.
\]
Consequently, the function $x\mapsto f_1(-R+2Rx)$ belongs to $\cH^\beta([0,1]^d,H^{(1)})$, where 
\begin{equation}
\label{eq:approx_h1_def}
    H^{(1)}
    \lesssim \beta^d (C\sdata^{-2}R^2)^\beta\beta!.
\end{equation}

\medskip
\noindent
\textbf{Step 2: log-integral term approximation.}
\quad
We next approximate $f_2$ on $[-R,R]^d$.
We first bound the derivatives of orders $|\bk| \in \{0,1,2\}$, which are not covered by Lemma \ref{lem:log_int_der_gaus}.

The normalization $\cT_\infty[\log\nu_{||}^*]=0$ and Jensen's inequality, yield
\begin{align*}
    \log\cT_{||}[\nu_{||}^*](x)
    &=
    \log\cT_\infty\left[
        \nu_{||}^*(\cdot)\,
        \sfq_{||}(\cdot\,|\,x)\,
        (2\pi\gamma\sigma_\infty^2)^{d/2}
        \exp\left\{
            \frac{\|\cdot\|^2}{2\gamma\sigma_\infty^2}
        \right\}
    \right]
    \\
    &\geq
    \cT_\infty[\log\nu_{||}^*]
    +\cT_\infty\left[
        \log\sfq_{||}(\cdot \,|\,x)
        +\frac d2\log(2\pi\gamma\sigma_\infty^2)
        +\frac{\|\cdot\|^2}{2\gamma\sigma_\infty^2}
    \right]
    \\
    &=
    \frac d2\log\frac{\sigma_\infty^2}{\sigma_T^2}
    -\frac{d\gamma\sigma_\infty^2
            +e^{-2bT}\|x\|^2}{2\gamma\sigma_T^2}
    +\frac d2
    \\
    &=
    -\frac d2\left(
        \frac{\sigma_\infty^2}{\sigma_T^2}
        -1+\log\frac{\sigma_T^2}{\sigma_\infty^2}
    \right)
    -\frac{e^{-2bT}\|x\|^2}{2\gamma\sigma_T^2}.
\end{align*}
Using $\log\nu_{||}^*\leq M$, we obtain
\begin{equation}
\label{eq:approx_f2_weight_bounds}
    e^{-M}
    \leq
    \frac1{\cT_{||}[\nu_{||}^*](x)}
    \leq
    C\exp\left\{
       \frac{e^{-2bT}\|x\|^2}{2\gamma\sigma_T^2}
    \right\}.
\end{equation}
The lower bound in \eqref{eq:approx_f2_weight_bounds} and Gaussian integration yield
\begin{align*}
    e^{f_2(u)}
    \geq
    e^{-M}
    \left(
       1+\frac{e^{-2bT}}{\gamma\sigma_T^2}
    \right)^{-d/2}
    \exp\left\{
       \frac{e^{-2bT}\|u\|^2}
       {2\gamma\sigma_T^2
          (\gamma\sigma_T^2+e^{-2bT})}
    \right\}.
\end{align*}
Combining this with the upper bound in \eqref{eq:approx_f2_weight_bounds}, and completing the square in the corresponding Gaussian integrals, we obtain
\begin{align*}
    \|\nabla f_2(u)\|
    &\lesssim
    e^{M-bT}
    (1+e^{-bT}\|u\|)
    \exp\{Ce^{-2bT}\|u\|^2\},
    \\
    \|\nabla^2f_2(u)\|_\infty
    &\lesssim
    e^{M-2bT}
    (1+e^{-bT}\|u\|)^2
    \exp\{Ce^{-2bT}\|u\|^2\}.
\end{align*}
Since $e^{-bT}R\lesssim1$ and $e^{2M-bT}\lesssim1$, it follows that
\begin{equation*}
    \max_{1\leq|\bk|\leq2}
    \|\partial^\bk f_2\|_{L^\infty([-R,R]^d)}
    \lesssim1.
\end{equation*}
To control the zeroth-order term, we use the normalization
of the potential.
The global gradient bound above, integrated along the segment from $0$ to $u$, and condition $e^{M-bT}\lesssim1$ imply that $\cT_\infty[|f_2-f_2(0)|]\lesssim1$.
On the other hand, the decomposition of the potential gives
\[
    \cT_\infty[f_2]
    =\cT_\infty[f_1] +\frac d2\log(2\pi\gamma\sigma_T^2).
\]
From \eqref{eq:approx_f1_compact} together with the Gaussian concentration bound \citep[Theorem 1.19]{rigollet2023} we obtain $\cT_\infty[|f_1|]\lesssim\sdata^{-2}$.
Consequently, $|f_2(0)|\lesssim\sdata^{-2}$ and
\[
    \|f_2\|_{L^\infty([-R,R]^d)}
    \lesssim\sdata^{-2}+R.
\]
Thus, Lemma \ref{lem:log_int_der_gaus} implies that, for every $|\bk| \geq 3$,
\begin{align*}
    \max_{|\bk|=k}
    \|\partial^\bk f_2\|_{L^\infty([-R,R]^d)}
    \leq
    \left[
       \frac{Ce^{-bT}}{\gamma\sigma_T^2}
       \exp\left\{
          2M+3+
          \frac{e^{-2bT}dR^2}
               {2(\gamma\sigma_T^2)^2}
       \right\}
    \right]^k k!
    \leq C^k k!,
\end{align*}
where the last inequality uses \eqref{eq:approx_compact_bT_cond}.
Using the bounds for orders $|\bk| \in \{0,1,2\}$, we deduce that
\begin{equation*}
    \max_{|\bk|\leq\beta}
    \|\partial^\bk f_2\|_{L^\infty([-R,R]^d)}
    \leq C(\sdata^{-2}+R)+C^\beta\beta!.
\end{equation*}
Therefore, after the change of variables, we have
$x\mapsto f_2(-R+2Rx)\in\cH^\beta([0,1]^d,H^{(2)})$ with
\begin{equation}
\label{eq:approx_h2_def}
    H^{(2)}
    \lesssim \beta^d (C\sdata^{-2}R^2)^\beta\beta!.
\end{equation}
Apply Lemma \ref{lem:smidt_hieber} to the two rescaled functions $f_1$ and $f_2$ with common parameters
\[
    \beta=\left\lceil C\log(e/\eta)\right\rceil,
    \quad
    N = \left\lceil
       (C\sdata^{-2}R^2\beta)^d
    \right\rceil
    \vee\left\lceil2e^d\right\rceil,
\]
where the constants are sufficiently large.
Choose an integer $m\geq1$ such that $\beta^2N2^{-m} \leq3^\beta N^{-\beta/d}$ and $m\lesssim\beta\log N$.
Such a choice is possible by taking
\[
    m=
    \max\left\{
       1,\,
       \left\lceil
       \log_2\bigl(
          \beta^2N^{1+\beta/d}3^{-\beta}
       \bigr)
       \right\rceil
    \right\}.
\]
Hence, by \eqref{eq:approx_h1_def}, \eqref{eq:approx_h2_def}, and Stirling's formula, the approximation error for each rescaled function is bounded by
\begin{align*}
    C\beta^d
    (C\sdata^{-2}R^2)^\beta\beta!\,
    3^\beta N^{-\beta/d}
    \lesssim
    \beta^{d+1/2}
    \left(
       \frac{C\sdata^{-2}R^2\beta}{N^{1/d}}
    \right)^\beta
    \leq\eta/2.
\end{align*}
Composing the resulting networks with the linear map $u\mapsto(u+R)/(2R)$ gives neural networks $\tilde f_1,\tilde f_2$ such that $\|\tilde f_j-f_j\|_{L^\infty([-R,R]^d)}\leq\eta/2$ for $j = 1, 2$.
Moreover, $N\lesssim (\sdata^{-2}R^2)^d \bigl(\log(e/\eta)\bigr)^d$ and
\[
    m\lesssim
    \log(e/\eta)
    \left(
       \log(\sdata^{-2}R)
       +\log\log(e/\eta)
    \right).
\]
Consequently, by Lemma \ref{lem:smidt_hieber}, for $j=1,2$ we have that $\tilde f_j \in \NN(L^{(j)}, W^{(j)}, S^{(j)}, B^{(j)})$ with
\begin{align}
\label{eq:approx_f1_cfg}
    &L^{(j)}
    \lesssim
      \log^2(e/\eta)\log(\sdata^{-2}R),
    \quad
    \log B^{(j)}
    \lesssim
      \log^2(e/\eta)\log(\sdata^{-2}R),
    \\
\label{eq:approx_f2_cfg}
    &S^{(j)}\vee\|W^{(j)}\|_\infty
    \lesssim
      \bigl(\log(e/\eta)\bigr)^{2d+5}
      (\sdata^{-2}R^2)^d
      \log(\sdata^{-2}R).
\end{align}

\medskip
\noindent
\textbf{Step 3: ensuring regularity on the entire space.}
\quad
For every $x\in\R$, define
\[
    f_{\mathrm{clip}}(x)
    =\min\{\max\{x,-R\},R\}
    =-R+\relu(x+R)-\relu(x-R).
\]
This function is represented exactly by a ReLU network of constant depth and width, with weights bounded by $R$.
By parallel stacking, we extend $f_{\mathrm{clip}}$ to act component-wise on $\R^d$.
Define the final approximation by
\[
    \tilde f(x)
    =
    \tilde f_1(f_{\mathrm{clip}}(x))
    -
    \tilde f_2(f_{\mathrm{clip}}(x))
    +
    \frac d2\log(2\pi\gamma\sigma_T^2).
\]
Therefore, we conclude that
\begin{align*}
    \|\tilde f-\log\nu_{||}^*\|_{L^\infty([-R,R]^d)}
    \leq
    \|\tilde f_1-f_1\|_{L^\infty([-R,R]^d)}
    +
    \|\tilde f_2-f_2\|_{L^\infty([-R,R]^d)}
    \leq\eta.
\end{align*}
Furthermore, $f_{\mathrm{clip}}(x)\in[-R,R]^d$ for every
$x\in\R^d$, and hence $\|\tilde f -\log\nu_{||}^* \circ f_{\mathrm{clip}} \|_{L^\infty(\R^d)} \leq\eta$.
It follows that, for all $x\in\R^d$,
\[
    \min_{u\in[-R,R]^d}\log\nu_{||}^*(u)-\eta
    \leq\tilde f(x)
    \leq
    \max_{u\in[-R,R]^d}\log\nu_{||}^*(u)+\eta
    \leq M+\eta.
\]
Finally, standard concatenation and parallelization of ReLU neural networks \citep{nakada20} together with \eqref{eq:approx_f1_cfg} and \eqref{eq:approx_f2_cfg}, yield that $\tilde f\in\NN(L,W,S,B)$ with
\begin{align*}
    L \vee \log B \lesssim
      \log^2(e/\eta)\log(\sdata^{-2}R),
    \quad
    \|W\|_\infty\vee S
    &\lesssim
      \bigl(\log(e/\eta)\bigr)^{2d+5}
      (R^2\sdata^{-2})^d
      \log(\sdata^{-2}R).
\end{align*}
The proof is now complete.

\myendproof

\subsection{Proof of Lemma \ref{lem:psi_norm_sup}}
\label{sec:lem_psi_norm_sup_proof}

We aim to find $u > 0$ such that
\begin{align*}
    \E \sup_{\psi \in \cF_{\NN}} \exp\left\{ \frac{1}{2u} \big|\log\sfp_T^\psi(Y) - \log\sfp_T(Y)\big| \right\} \leq 2 .
\end{align*}
Using Jensen's inequality, we obtain that
\begin{align}
\label{eq:sup_psi_decomp}
    \notag
    &\E \sup_{\psi \in \cF_{\NN}} \exp\left\{ \frac{1}{2u} \big|\log\sfp_T^\psi(Y) - \log\sfp_T(Y)\big| \right\}
    \\&
    \leq \left( \E\sup_{\psi \in \cF_{\NN}} \exp\left\{ \frac{1}{u} \log\left(\frac{\sfp_T^\psi(Y)}{\sfp_T(Y)}\right) \right\}
    + \E \sup_{\psi \in \cF_{\NN}} \exp\left\{\frac{1}{u} \log\left(\frac{\sfp_T(Y)}{\sfp_T^\psi(Y)}\right)\right\}
    \right)^{1/2} .
\end{align}
We now analyze each term individually.

\noindent
\textbf{Step 1: analyzing the first term.}\quad
We establish the lower bound on the OU operator $\cT_{||}[e^\varphi](x)$.
Define the random vector $X = e^{-bT} x + \sqrt{\gamma} \sigma_T \xi \sim \sfq_T(\cdot \, | \, x)$, where $\xi \sim \cN(0, I_d)$.
Hence, using the lower bound on $\varphi$ and evaluating the Gaussian integral, we obtain
\begin{align*}
    \cT_{||}[e^\varphi](x)
    = \E[e^{\varphi(X)}]
    \geq \E [e^{-A(1 + \|X\|^2)}]
    = e^{-A}(1 + 2\gamma \sigma_T^2 A)^{-d/2} \exp\left\{-\frac{A e^{-2bT} \|x\|^2}{1 + 2\gamma\sigma_T^2 A}\right\} .
\end{align*}
Consequently, \eqref{eq:sup_psi_norm_ld_density} implies that
\begin{align*}
    \sfp_{||}^\varphi(u)
    &\leq
    e^{2M+A}(1 + 2\gamma\sigma_T^2 A)^{d/2}
    \int
    \exp\left\{\frac{A e^{-2bT} \|x\|^2}{1 + 2\gamma\sigma_T^2 A}\right\}
    \sfq_{||}(u\mid x)\rho_{||}(x) \, \dd x.
\end{align*}
Since $\rho_{||}$ and $\sfq_{||}(u \, | \, x)$ are Gaussian densities, as specified in Lemma \ref{lem:log_pot_decomp}, completing the square yields
\begin{equation}
\label{eq:sup_psi_p_parallel_v}
    \sfp_{||}^\varphi(u)
    \leq
    e^{2M+A}(2\pi \mathsf{v})^{-d/2}
    \exp\left\{\left(A - \frac1{2 \mathsf{v}}\right)\|u\|^2\right\},
\end{equation}
where
\[
    \mathsf{v} = \frac{\gamma\sigma_T^2 (1 + 2\gamma\sigma_T^2A) + e^{-2bT}}{(1 + 2\gamma\sigma_T^2 A)^2}.
\]
Using the decomposition \eqref{eq:sup_psi_log_p_decomp}, together with $\sigma \geq \sigma_{\min}$, and dropping the negative quadratic terms, we conclude that
\begin{align}
\label{eq:sup_psi_norm_p_psi_ub}
    \log\sfp_T^\psi(y)
    \leq 2M + A(1 + \|V^\top y\|^2)
    -\frac d2\log(2\pi \mathsf{v})
    -\frac{D-d}{2}\log(2\pi\sigma_{\min}^2).
\end{align}
On the other hand, by Assumption \ref{asn:linear_subspace}, we may represent $Y \sim \sfp_T$ as $Y = GZ + \sdata\xi$.
Jensen's inequality yields
\begin{align*}
    \sfp_T(y)
    =(2\pi\sdata^2)^{-D/2}
    \E\exp\left\{-\frac{\|y-GZ\|^2}{2\sdata^2}\right\}
    \geq
    (2\pi\sdata^2)^{-D/2}
    \exp\left\{-\frac{\|y\|^2+\E\|Z\|^2}{2\sdata^2}\right\},
\end{align*}
where we used $\E Z = 0$ and $G^\top G = I_d$.
Combining this with \eqref{eq:sup_psi_norm_p_psi_ub}, for every $\psi \in \cF_{\NN}$, we have that
\begin{align*}
    \log\frac{\sfp_T^\psi(y)}{\sfp_T(y)}
    \le C_0 + \left(A+\frac1{2\sdata^2}\right)\|y\|^2,
\end{align*}
where
\begin{align*}
    C_0=
    2M + A - \frac{d}{2}\log(2\pi\mathsf{v}) - \frac{D - d}{2} \log(2\pi\sigma_{\min}^2)
    + \frac{D}{2}\log(2\pi\sdata^2) + \frac{\E \|Z\|^2}{2\sdata^2} .
\end{align*}
Absorbing the sub-Gaussian parameter of $Z$ into the implicit constants and using \cite[Proposition 2.5.2]{vershynin18}, we have $\E\|Z\|^2\lesssim d$.
Furthermore, by \cite[Proposition 2.7.1]{vershynin18}, it follows that $\left\| \|Y\|^2 \right\|_{\psi_1} \lesssim d + \sdata^2 D$.
Therefore, assuming that $u \geq \|\|Y\|^2\|_{\psi_1} (A + (2\sdata^2)^{-1})$ and using Jensen's inequality, we deduce that
\begin{align*}
    \E\sup_{\psi\in\cF_{\NN}} \exp\left\{\frac1u\log\frac{\sfp_T^\psi(Y)}{\sfp_T(Y)}\right\}
    &\leq e^{C_0/u}\E \exp\left\{ \frac{\|Y\|^2}{\| \|Y\|^2\|_{\psi_1}} \cdot \frac{\|\|Y\|^2\|_{\psi_1} (A + (2\sdata^2)^{-1}) }{u}  \right\}
    \\&
    \leq e^{C_0/u} \exp\left\{ \frac{\|\|Y\|^2\|_{\psi_1} (A + (2\sdata^2)^{-1}) \log 2 }{u} \right\} .
\end{align*}
Thus, setting $u \gtrsim C_0 + \| \|Y\|^2 \|_{\psi_1} (A + (2\sdata^2)^{-1})$ guarantees that
\begin{align}
\label{eq:sup_psi_first_term}
    \E\sup_{\psi\in\cF_{\NN}} \exp\left\{\frac1u\log\frac{\sfp_T^\psi(Y)}{\sfp_T(Y)}\right\} \leq 2 .
\end{align}
Next, by Definition \ref{def:log_pot_class}, we have $A \lesssim \sigma_{\min}^{-2}$.
Moreover, \eqref{eq:sup_psi_p_parallel_v} suggests that $\log(e \vee \mathsf{v}^{-1}) \lesssim \log(e\sigma_{\min}^{-2})$.
Thus, it follows that
\begin{align*}
    C_0 + \| \|Y\|^2 \|_{\psi_1} (A + (2\sdata^2)^{-1})
    \lesssim M + D\sigma_{\min}^{-2} + \log(e \vee \mathsf{v}^{-1})
    \lesssim M + D\sigma_{\min}^{-2} .
\end{align*}
Hence, setting $u \gtrsim M + D\sigma_{\min}^{-2}$ guarantees the desired bound \eqref{eq:sup_psi_first_term}.

\noindent
\textbf{Step 2: analyzing the second term.}\quad
We first establish a lower bound on $\sfp_T^\psi$.
Since $\varphi(z) \leq 2M$, we have $\cT_{||}[e^\varphi](x)\leq e^{2M}$.
Combining this with $\varphi(z)\geq -A(1 + \|z\|^2)$ and \eqref{eq:sup_psi_norm_ld_density}, we obtain
\begin{align*}
    \log \sfp_{||}^\varphi(z)
    \geq -2M-A(1+\|z\|^2) + \log \int \sfq_{||}(z\mid x)\rho_{||}(x)\,\dd x ,
    \quad \text{for all $z \in \R^d$} .
\end{align*}
Set $h^2 = e^{-2bT}+\gamma\sigma_T^2$.
Since $\rho_{||}$ is the standard Gaussian density, the last integral is the density of $\cN(0,h^2 I_d)$ at $z$.
Hence, we have that
\begin{align*}
    \log\sfp_{||}^\varphi(z)
    \geq -2M-A-\frac{d}{2}\log(2\pi h^2) -\left(A+\frac1{2 h^2}\right)\|z\|^2.
\end{align*}
Using \eqref{eq:sup_psi_log_p_decomp}, it follows that
\begin{align}
\label{eq:sup_psi_norm_p_psi_lb}
    \notag
    \log\sfp_T^\psi(y)
    &\geq -2M-A-\frac d2\log(2\pi h^2) - \frac{D-d}{2}\log(2\pi\sigma^2)
    \\&\quad
    -\left(A + \frac1{2h^2}\right)\|V^\top y\|^2 -\frac{\|(I_D-VV^\top)y\|^2}{2\sigma^2}.
\end{align}
We now aim to obtain an upper bound on $\log \sfp_T(y)$.
In view of Assumption \ref{asn:linear_subspace}, the Gaussian mixture representation of the target density yields
\begin{align*}
    \sfp_T(y)
    = (2\pi\sdata^2)^{-D/2} \E\exp\left\{-\frac{\|y - GZ\|^2}{2\sdata^2}\right\}
    \leq(2\pi\sdata^2)^{-D/2}.
\end{align*}
Combining this with \eqref{eq:sup_psi_norm_p_psi_lb}, we obtain that, for all $y \in \R^D$,
\begin{align*}
    \log\frac{\sfp_T(y)}{\sfp_T^\psi(y)}
    &\leq 2M + A + \frac{d}{2} \log h^2
    + \frac{D-d}{2}\log\sigma^2+\frac D2\log\sdata^{-2}
    \\&\quad
    + \left(A+\frac1{2h^2}\right)\|V^\top y\|^2
    + \frac{\|(I_D-VV^\top)y\|^2}{2\sigma^2}.
\end{align*}
Consequently, we obtain the uniform bound
\begin{align}
\label{eq:sup_psi_log_dr_aux}
    \sup_{\psi\in\cF_{\NN}} \log\frac{\sfp_T(y)}{\sfp_T^\psi(y)}
    \leq C_1 + \left( A + \frac{1}{2 h^2} + \frac{1}{2\sigma_{\min}^2} \right)\|y\|^2 ,
\end{align}
where
\begin{align*}
    C_1 = 2M + A + \frac{d}{2}\log h^2 +\frac D2\log\sdata^{-2} .
\end{align*}
As in Step 1, for $u\geq (A + (2h^2)^{-1} + (2\sigma_{\min}^2)^{-1}) \| \|Y\|^2 \|_{\psi_1}$, Jensen's inequality implies that
\begin{align*}
    \E\sup_{\psi\in\cF_{\NN}}
    \exp\left\{\frac{1}{u}\log\frac{\sfp_T(Y)}{\sfp_T^\psi(Y)}\right\}
    &\leq e^{C_1/u}\E \exp\left\{ \left(A + \frac{1}{2h^2} + \frac{1}{2\sigma_{\min}^2} \right) \frac{\| \|Y\|^2 \|_{\psi_1}}{u} \frac{\|Y\|^2}{\| \|Y\|^2 \|_{\psi_1}} \right\}
    \\&
    \leq\exp\left\{\frac{C_1 + (A + (2h^2)^{-1} + (2\sigma_{\min}^2)^{-1}) \|\|Y\|^2\|_{\psi_1} \log2}{u}\right\}.
\end{align*}
Using the bounds $A \lesssim \sigma_{\min}^{-2}$ and $2M\lesssim M + D\sigma_{\min}^{-2}$ together with $\|\|Y\|^2\|_{\psi_1} \lesssim d+\sdata^2D$, we conclude that
\begin{align*}
    C_1 + \left(A + \frac{1}{2h^2} + \frac{1}{2\sigma_{\min}^2} \right) \|\|Y\|^2\|_{\psi_1}
    \lesssim M + D\sigma_{\min}^{-2}.
\end{align*}
Therefore, choosing $u \gtrsim M + D\sigma_{\min}^{-2}$ guarantees
\begin{align}
\label{eq:sup_psi_second_term}
    \E\sup_{\psi\in\cF_{\NN}}
    \exp\left\{\frac1u\log\frac{\sfp_T(Y)}{\sfp_T^\psi(Y)}\right\}
    \leq2.
\end{align}
Furthermore, the bound in \eqref{eq:sup_psi_log_dr_aux} reduces to
\begin{align*}
    \sup_{\psi\in\cF_{\NN}} \log\frac{\sfp_T(y)}{\sfp_T^\psi(y)}
    \lesssim M + D\sigma_{\min}^{-2} + \sigma_{\min}^{-2} \|y\|^2 .
\end{align*}
This completes the proof of claim $(ii)$.

\noindent
\textbf{Step 3: deriving the final bound.}\quad
Combining \eqref{eq:sup_psi_second_term}, \eqref{eq:sup_psi_first_term}, and the decomposition \eqref{eq:sup_psi_decomp}, we obtain
\begin{align*}
    \E\exp\left\{ \frac{1}{2u} \sup_{\psi\in\cF_{\NN}} \left|\log\sfp_T^\psi(Y) - \log\sfp_T(Y)\right| \right\}
    \leq \sqrt{2 + 2} = 2 .
\end{align*}
Consequently, we conclude that
\begin{align*}
    \left\|\sup_{\psi\in\cF_{\NN}} \left|\log\sfp_T^\psi(Y)-\log\sfp_T(Y)\right| \right\|_{\psi_1}
    \lesssim M + D\sigma_{\min}^{-2}.
\end{align*}
The implicit constant may depend on $\gamma$, $b$, $\tau$, and $d$, but not on $bT$, $D$, or $\sigma_{\min}$.
The proof is now complete.

\myendproof

\subsection{Proof of Lemma \ref{lem:bernstein_log_ratio}}
\label{sec:lem_bernstein_log_ratio_proof}

We follow the same argument as in Step 3 of the proof of \cite[Theorem 1]{puchkin25}, using \cite[Lemmata C.1 and C.2]{puchkin25}.
Fix $\psi \in \cF_{\NN}$ and define
\[
    L_\psi(y)=\log\frac{\sfp_T(y)}{\sfp_T^\psi(y)}.
\]
All expectations below are taken under $Y \sim \sfp_T$, unless another distribution is specified.
Since the proof is quite technical, we divide it into several steps.

\noindent
\textbf{Step 1: sub-Gaussian variance proxy.}\quad
In view of Assumption \ref{asn:linear_subspace}, there exists a positive constant $\nu_Z \lesssim \tau$ such that
\begin{align*}
    \E \exp\{u^\top Z\}
    \leq \exp\{\nu_Z^2\|u\|^2 / 2\},
    \quad \text{for all $u \in \R^d$}.
\end{align*}
In addition, Assumption \ref{asn:linear_subspace} implies that $Y = GZ + \sdata \xi$, where $Z$ and $\xi$ are independent.
Hence, for every $w \in \R^D$,
\begin{align*}
    \E \exp\{w^\top Y\}
    &
    = \E \exp\{w^\top G Z\} \; \E \exp\{\sdata w^\top \xi\}
    \\&
    \leq \exp\left\{ \frac{\nu_Z^2\|G^\top w\|^2 + \sdata^2\|w\|^2}{2} \right\}
    \\&
    \leq \exp\left\{\frac{\nu^2\|w\|^2}{2}\right\},
\end{align*}
where $\nu^2 = \nu_Z^2 + 1$.
Thus, $\nu^2$ is a variance proxy for the sub-Gaussian random vector $Y$ satisfying $\nu^2 \lesssim 1$.

\noindent
\textbf{Step 2: comparison with a mixture density.}\quad
For $\omega\in(0,1)$, let us introduce
\begin{align*}
    \sfp_{\omega}^\psi(y)
    = (1-\omega)\sfp_T^\psi(y) + \omega\sfp_T(y),
    \quad \text{and} \quad
    R_\psi(\omega)
    =\E\left( \log\frac{\sfp_{\omega}^\psi(Y)}{\sfp_T^\psi(Y)} \right)^2.
\end{align*}
Applying \cite[Lemma C.1]{puchkin25} with $\mathsf p=\sfp_T$ and $\sfq = \sfp_T^\psi$, we obtain
\begin{align}
\label{eq:bernstein_mixture_comparison}
    \Var [L_\psi(Y)]
    \leq\E (L_\psi(Y))^2
    \leq2\log(1/\omega)\, \KL(\sfp_T, \sfp_T^\psi) + 2R_\psi(\omega).
\end{align}
The quadratic bound established in claim $(ii)$ of Lemma \ref{lem:psi_norm_sup} and the variance proxy from Step 1 verify the assumptions of \cite[Lemma C.2]{puchkin25}.
Consequently, for $\omega\in(0,1/2]$, it follows that
\begin{align}
\label{eq:bernstein_mixture_R}
    R_\psi(\omega)
    \lesssim
    \omega^2 + e^\beta\omega
    + 6^D\left(\log(1/\omega) + \alpha \right) e^{\beta/(16\alpha\nu^2)}
    \omega^{1/(16\alpha\nu^2)} ,
\end{align}
where we choose $0 < \alpha \lesssim \sigma_{\min}^{-2}$ and $0 \leq \beta \lesssim M + D\sigma_{\min}^{-2}$.

\noindent
\textbf{Step 3: choice of $\omega$.}\quad
Fix $n \geq 2$ and set $\omega = e^{-\beta} (n^{-1} \wedge (6^D n)^{-16 \alpha \nu^2}) \in (0, 1/2]$.
This choice ensures that
\begin{align*}
    e^\beta \omega \vee 6^D e^{\beta / (16\alpha\nu^2)} \omega^{1 / (16\alpha\nu^2)} \leq \frac{1}{n} .
\end{align*}
Moreover, we have that
\begin{align*}
    \log(1 / \omega)
    \lesssim \beta + \log n + \alpha (D + \log n)
    \lesssim M + D \sigma_{\min}^{-2} + \sigma_{\min}^{-2} \log n .
\end{align*}
Substituting these bounds into \eqref{eq:bernstein_mixture_R}, we arrive at
\begin{align*}
    R_\psi(\omega)
    \lesssim \frac{1}{n^2} + \frac{\log(1 / \omega) + \alpha}{n}
    \lesssim \frac{(M + D\sigma_{\min}^{-2}) \log n}{n} .
\end{align*}
Combining this with \eqref{eq:bernstein_mixture_comparison}, we conclude that
\begin{align*}
    \Var[L_\psi(Y)]
    \leq \E(L_\psi(Y))^2
    \lesssim (M + D\sigma_{\min}^{-2}) \log n \, \left(\KL(\sfp_T, \sfp_T^\psi) + \frac{1}{n} \right)
\end{align*}
thereby completing the proof.

\myendproof

\subsection{Proof of Lemma \ref{lem:impr_loss_prox}}
\label{sec:lem_impr_loss_prox_proof}

In view of \eqref{eq:sup_psi_log_p_decomp}, we have that
\begin{align}
\label{eq:loss_prox_decomp}
    \notag
    \left|\log \sfp_T^{\psi_1}(y) - \log \sfp_T^{\psi_2}(y) \right|
    &\leq \left| \log \sfp_{||}^{\varphi_1}(V_1^\top y) - \log \sfp_{||}^{\varphi_2}(V_2^\top y) \right|
    + \frac{D - d}{2} \left|\log \sigma_1^2 - \log \sigma_2^2 \right|
    \\&\quad
    + \left| \frac{\|(I_D - V_1V_1^\top)y\|^2}{2\sigma_1^2} - \frac{\|(I_D - V_2V_2^\top)y\|^2}{2\sigma_2^2} \right| .
\end{align}
From \eqref{eq:sup_psi_norm_ld_density} we further deduce that
\begin{align}
\label{eq:loss_prox_density_ld_prox}
    \notag
    &\left| \log \sfp_{||}^{\varphi_1}(V_1^\top y) - \log \sfp_{||}^{\varphi_2}(V_2^\top y) \right|
    \leq \left| \varphi_1(V_1^\top y) - \varphi_2(V_2^\top y) \right|
    \\&\quad
    + \left| \log \int \frac{\sfq_{||}(V_1^\top y \, | \, x) \rho_{||}(x) \, \dd x}{\cT_T[e^{\varphi_1}](x)}
    -  \log \int \frac{\sfq_{||}(V_2^\top y \, | \, x) \rho_{||}(x) \, \dd x}{\cT_T[e^{\varphi_2}](x)}
    \right| .
\end{align}
We now bound the Lipschitz constant of $\varphi_1$ and $\varphi_2$ using the fact that the Lipschitz constant of a composition is at most the product of the individual constants.
Specifically, it follows that
\begin{align*}
    |\varphi_1(u) - \varphi_1(v)|
    \leq \sqrt{d}(\|W\|_\infty + 1)^{L}(B \vee 1)^L \|u - v\|,
    \quad \text{for all $u, v \in \R^d$} .
\end{align*}
In particular, $|\varphi_1(u) - \varphi_1(v)| \leq \Lambda_\varphi \|u - v\|$, where $\Lambda_\varphi$ is given by \eqref{eq:loss_prox_lambda_def}.
Using the fact that $\cT_T[e^{\varphi_1}](x) = \E\exp(\varphi_1(e^{-bT}x + \sqrt{\gamma}\sigma_T \xi))$ with $\xi \sim \cN(0, I_d)$ and using Lipschitzness of $\varphi_1$, we deduce that, for all $x, x' \in \R^d$,
\begin{align*}
    \exp\{-e^{-bT}\Lambda_\varphi \|x - x'\|\}\cT_T[e^{\varphi_1}](x')
    \leq \cT_T[e^{\varphi_1}](x)
    \leq \exp\{e^{-bT}\Lambda_\varphi \|x - x'\|\} \cT_T[e^{\varphi_1}](x') .
\end{align*}
Next, taking into account that both $\sfq_{||}(\cdot \, | \, x)$ and $\rho_{||}(\cdot)$ are Gaussian densities, for all $u, v \in \R^d$, we obtain
\begin{align*}
    &\left| \log \int \frac{\sfq_{||}(u \, | \, x) \rho_{||}(x) \, \dd x}{\cT[e^{\varphi_1}](x)}
    - \log \int \frac{\sfq_{||}(v \, | \, x) \rho_{||}(x) \, \dd x}{\cT[e^{\varphi_1}](x)}
    \right|
    \\&
    \leq \left| \log \E \frac{1}{\cT[e^{\varphi_1}](\tilde\sigma \xi + \mu_u)} - \log \E \frac{1}{\cT[e^{\varphi_1}](\tilde\sigma \xi + \mu_v)}  \right|
    + \frac{\left| \|u\|^2 - \|v\|^2 \right|}{2(e^{-2bT} + \gamma\sigma_T^2)} .
\end{align*}
where $\xi \sim \cN(0, I_d)$, $\tilde\sigma^2 = \gamma\sigma_T^2 / (\gamma\sigma_T^2 + e^{-2bT})$, $\mu_u =  e^{-bT} u / (e^{-2bT} + \gamma\sigma_T^2)$, and $\mu_v$ is defined similarly to $\mu_u$ .
Consequently, we conclude that, for all $u,v \in \R^d$,
\begin{align*}
    \left| \log \int \frac{\sfq_{||}(u \, | \, x) \rho_{||}(x) \, \dd x}{\cT[e^{\varphi_1}](x)}
    - \log \int \frac{\sfq_{||}(v \, | \, x) \rho_{||}(x) \, \dd x}{\cT[e^{\varphi_1}](x)}
    \right|
    \leq \frac{\Lambda_\varphi e^{-2bT}\|u - v\|}{e^{-2bT} + \gamma\sigma_T^2}
    + \frac{\left| \|u\|^2 - \|v\|^2 \right|}{2(e^{-2bT} + \gamma\sigma_T^2)} .
\end{align*}
Combining this with \eqref{eq:loss_prox_decomp} and \eqref{eq:loss_prox_density_ld_prox}, it follows that
\begin{align*}
    \left|\log \sfp_T^{\psi_1}(y) - \log \sfp_T^{\psi_2}(y) \right|
    &\lesssim \Lambda_\varphi\|V_1 - V_2\| \cdot \|y\| + |\varphi_1(V_2^\top y) - \varphi_2(V_2^\top y)|
    \\&\quad
    + \Lambda_\varphi e^{-2bT} \|V_1 - V_2\| \cdot \|y\| + \|V_1 - V_2\| \cdot \|y\|^2
    \\&\quad
    +  \left| \log \int \frac{\sfq_{||}(V_2^\top y \, | \, x) \rho_{||}(x) \, \dd x}{\cT_T[e^{\varphi_1}](x)}
    -  \log \int \frac{\sfq_{||}(V_2^\top y \, | \, x) \rho_{||}(x) \, \dd x}{\cT_T[e^{\varphi_2}](x)}
    \right|
    \\&\quad
    + D |\log(\sigma_1^2 / \sigma_2^2)|
    + \left| \frac{\|(I_D - V_1V_1^\top)y\|^2}{2\sigma_1^2} - \frac{\|(I_D - V_2V_2^\top)y\|^2}{2\sigma_2^2} \right| .
\end{align*}
Since $\sigma_1 \wedge \sigma_2 \geq \sigma_{\min}$, we have that $\left|\log(\sigma_1^2 / \sigma_2^2) \right| \lesssim \sigma_{\min}^{-2}|\sigma_1^2 - \sigma_2^2|$ and
\begin{align*}
    \left| \frac{\|(I_D - V_1V_1^\top)y\|^2}{2\sigma_1^2} - \frac{\|(I_D - V_2V_2^\top)y\|^2}{2\sigma_2^2} \right|
    \lesssim \left(\sigma_{\min}^{-4}|\sigma_1^2 - \sigma_2^2| + \sigma_{\min}^{-2}\|V_1 - V_2\| \right) \|y\|^2 .
\end{align*}
Therefore, the above bound simplifies to
\begin{align}
\label{eq:loss_prox_almost_final}
    \notag
    \left|\log \sfp_T^{\psi_1}(y) - \log \sfp_T^{\psi_2}(y) \right|
    &\lesssim (\Lambda_\varphi + \sigma_{\min}^{-2}) \|V_1 - V_2\| (1 + \|y\|^2)
    + (D \sigma_{\min}^{-2} + \sigma_{\min}^{-4} \|y\|^2) |\sigma_1^2 - \sigma_2^2|
    \\&\quad
    + |\varphi_1(V_2^\top y) - \varphi_2(V_2^\top y)|
    \\&\quad
    \notag
    + \left| \log \int \frac{\sfq_{||}(V_2^\top y \, | \, x) \rho_{||}(x) \, \dd x}{\cT_T[e^{\varphi_1}](x)}
    -  \log \int \frac{\sfq_{||}(V_2^\top y \, | \, x) \rho_{||}(x) \, \dd x}{\cT_T[e^{\varphi_2}](x)}
    \right| .
\end{align}
Apply Lemma~\ref{lem:log_integr_diff_aux} with upper bound $2M$, quadratic coefficient $A$, and intercept $A\vee 2M$. The normalization follows from Definition~\ref{def:log_pot_class}. We obtain
\begin{align}
\label{eq:loss_prox_log_conv_b2_cor}
    \notag
    &\left| \log \int \frac{\sfq_{||}(V_2^\top y \, | \, x) \rho_{||}(x) \, \dd x}{\cT_T[e^{\varphi_1}](x)}
    -  \log \int \frac{\sfq_{||}(V_2^\top y \, | \, x) \rho_{||}(x) \, \dd x}{\cT_T[e^{\varphi_2}](x)}
    \right|
    \\&
    \lesssim \left(\cT_\infty|\varphi_1 - \varphi_2| \right)^{1 / \cK(T)}
    (d^2 + \gamma^{-1}\|y\|^2)^{1 - 1 / \cK(T)}
    \\&\quad\cdot
    \notag
    \exp\left\{\cO\left(d+M+\sqrt d\,e^{-bT}(M+\log(e+A+M))+e^{-bT}(1+\|y\|^2)\right)\right\} .
\end{align}
Here, $\cK(T)$ is given by
\begin{align*}
    1 \leq \cK(T)
    = (1 - e^{-2bT})^{-5e^2\sqrt{d}} \exp\left\{ 2e^2 \sqrt{d} \arcsin(e^{-bT}) \right\}
    = 1 + \cO(e^{-bT}) .
\end{align*}
It remains to bound the term $|\varphi_1(V_2^\top y) - \varphi_2(V_2^\top y)|$.
Proceeding as in the proof of \cite[Lemma 5]{schmidt2020}, we obtain
\begin{align*}
    \left|\varphi_1(u) - \varphi_2(u)\right|
    \leq (d + 1) L (B \vee 1)^{L} (\|W\|_\infty + 1)^{L}(1 + \|u\|) \eps,
    \quad \text{for all $u \in \R^d$} .
\end{align*}
In particular, we have $|\varphi_1(u) - \varphi_2(u)| \leq \Lambda_\varphi(1 + \|u\|)\eps$, where $\Lambda_\varphi$ is given by \eqref{eq:loss_prox_lambda_def}.
Taking into account that $|\sigma_1^2 - \sigma_2^2| \leq \eps$ and $\|V_1 - V_2\|_\infty \leq \eps$, we obtain from \eqref{eq:loss_prox_almost_final} and \eqref{eq:loss_prox_log_conv_b2_cor} that
\begin{align*}
    \left|\log \sfp_T^{\psi_1}(y) - \log \sfp_T^{\psi_2}(y) \right|
    &\leq D (\Lambda_\varphi + \sigma_{\min}^{-4})(1 + \|y\|^2) \eps
    \\&\quad
    + (\cT_\infty|\varphi_1 - \varphi_2|)^{1 / \cK(T)}
    \\&\qquad\cdot \exp\{ \cO(M + e^{-bT}(M + \log(e+A+M) + \|y\|^2)) \} .
\end{align*}
For a sufficiently large constant in $bT\gtrsim1$, we have
$\cK(T)\leq2$ and hence $1/\cK(T)\geq1/2$.
Since $0<\eps<1$ and $\Lambda_\varphi\geq1$, this implies
$\eps^{1/\cK(T)}\leq\eps^{1/2}$ and
$\Lambda_\varphi^{1/\cK(T)}\leq\Lambda_\varphi$. Therefore,

\begin{align*}
    &\left|\log \sfp_T^{\psi_1}(y) - \log \sfp_T^{\psi_2}(y) \right|
    \lesssim D(\Lambda_\varphi + \sigma_{\min}^{-4})(1 + \|y\|^2) \eps
    \\&\quad
    + \exp\{ \cO(M + e^{-bT}(M + \log(e+A+M) + \|y\|^2)) \} \; \Lambda_\varphi \eps^{1/2} .
\end{align*}
The proof is now complete.

\myendproof

\section{Proof of Lemma \ref{lem:factorizable_log_pot_decomp}}
\label{sec:lem_factorizable_log_pot_decomp_proof}

The proof proceeds in three steps.

\medskip
\noindent
\textbf{Step 1: factorization of the Gaussian density and the OU kernel.}\quad
Let $S = (S_1^\top, \ldots, S_J^\top)^\top \in \R^{D \times D}$.
Since $S_j S_k^\top = 0$ for $j \neq k$ and $S_j S_j^\top = I_{d_j}$ for every $j$, we have $S S^\top = I_D$.
As $S$ is square, it is orthogonal. Therefore, for every $v \in \R^D$,
\begin{align*}
    \|v\|^2 = \|S v\|^2 = \sum_{j=1}^J \|S_j v\|^2 .
\end{align*}
Applying this identity with $v = x$ and $v = y - e^{-bT} x$, and using $d_1 + \ldots + d_J = D$, we obtain
\begin{align*}
    &
    \sfp_0(x)
    = (2\pi)^{-D / 2} \exp\left\{-\frac{\|x\|^2}{2}\right\}
    = \prod_{j=1}^J (2\pi)^{-d_j / 2} \exp\left\{-\frac{\|S_j x\|^2}{2}\right\}
    = \prod_{j=1}^J \rho^{(j)}(S_j x),
    \\&
    \sfq_T(y \, | \, x)
    = \prod_{j=1}^J \sfq_T^{(j)}(S_j y \, | \, S_j x),
    \qquad
    \sfp_T(y) = \prod_{j=1}^J \sfp^{(j)}_T(S_j y) .
\end{align*}

\medskip
\noindent
\textbf{Step 2: a product ansatz solves the full Schr\"odinger system.}\quad
For each $j \in \{1, \ldots, J\}$, the pair $(\nu_0^{(j)}, \nu_T^{(j)})$ satisfies the Schr\"odinger system for the $d_j$-dimensional subproblem, namely
\begin{align*}
    \rho^{(j)}(u)
    &= \nu_0^{(j)}(u) \integral{\R^{d_j}} \sfq_T^{(j)}(v \, | \, u) \nu_T^{(j)}(v) \, \dd v,
    \quad u \in \R^{d_j},
    \\
    \sfp^{(j)}_T(v)
    &= \nu_T^{(j)}(v) \integral{\R^{d_j}} \sfq_T^{(j)}(v \, | \, u) \nu_0^{(j)}(u) \, \dd u,
    \quad v \in \R^{d_j} .
\end{align*}
Define
\begin{align*}
    \hat\nu_0(x) = \prod_{j=1}^J \nu_0^{(j)}(S_j x),
    \qquad
    \hat\nu_T(y) = \prod_{j=1}^J \nu_T^{(j)}(S_j y) .
\end{align*}
Consider the induced joint density
\begin{align*}
    \hat\pi(x, y) = \hat\nu_0(x) \, \sfq_T(y \, | \, x) \, \hat\nu_T(y),
    \quad x, y \in \R^D .
\end{align*}
Since all factors are non-negative, Tonelli's theorem applies.
Using the orthogonal change of variables $(u_1, \ldots, u_J) = S x$, whose Jacobian equals one, we get
\begin{align*}
    \integral{\R^D} \hat\pi(x, y) \, \dd x
    &= \prod_{j=1}^J \left[
        \nu_T^{(j)}(S_j y)
        \integral{\R^{d_j}} \sfq_T^{(j)}(S_j y \, | \, u_j) \nu_0^{(j)}(u_j) \, \dd u_j
    \right]
    \\
    &= \prod_{j=1}^J \sfp^{(j)}_T(S_j y)
    = \sfp_T(y) .
\end{align*}
Similarly, using the orthogonal change of variables $(v_1, \ldots, v_J) = S y$, we obtain
\begin{align*}
    \integral{\R^D} \hat\pi(x, y) \, \dd y
    &= \prod_{j=1}^J \left[
        \nu_0^{(j)}(S_j x)
        \integral{\R^{d_j}} \sfq_T^{(j)}(v_j \, | \, S_j x) \nu_T^{(j)}(v_j) \, \dd v_j
    \right]
    \\
    &= \prod_{j=1}^J \rho^{(j)}(S_j x)
    = \sfp_0(x) .
\end{align*}
Hence, $(\hat\nu_0, \hat\nu_T)$ is a pair of Schr\"odinger potentials for the original problem on $\R^D$.

\medskip
\noindent
\textbf{Step 3: identification with $(\nu_0^*, \nu_T^*)$.}\quad
By Step 2, $(\hat\nu_0, \hat\nu_T)$ is a non-negative pair of Schr\"odinger potentials for the original problem on $\R^D$.
By the uniqueness property (see \cite[Theorem 2.12]{leonard13}), any such pair coincides with $(\nu_0^*, \nu_T^*)$ up to reciprocal multiplicative constants.
Hence there exists $c > 0$ such that
\begin{align*}
    \hat\nu_0 = c \, \nu_0^*,
    \qquad
    \hat\nu_T = \nu_T^* / c .
\end{align*}
Replacing $(\nu_0^{(1)}, \nu_T^{(1)})$ with $(\nu_0^{(1)}/c,\,c\nu_T^{(1)})$ leaves the first subproblem unchanged, so we may absorb the constant $c$ into the first factor.
Taking logarithms then yields
\begin{align*}
    \log \nu_T^*(y) = \sum_{j=1}^J \log \nu_T^{(j)}(S_j y),
    \qquad
    \log \nu_0^*(x) = \sum_{j=1}^J \log \nu_0^{(j)}(S_j x) .
\end{align*}
This completes the proof.

\myendproof

\section{Proofs of the results from Section \ref{sec:proof_insights}}

\subsection{Proof of Lemma \ref{lem:log_density_der}}
\label{sec:lem_log_density_der_proof}

The proof is quite technical, so we divide it into two steps.

\noindent
\textbf{Step 1: exact expression for the $k$-th derivative.}\quad
Recalling the definition of $\sfp_{||}$ (see \eqref{eq:rho_par_def}), we have that
\begin{align*}
    \nabla^k\left(\log \sfp_{||}(u) + \frac{\|u\|^2}{2\sdata^2}\right)
    = \nabla^k \log\integral{\R^d} \exp\left\{ \frac{u^\top z}{\sdata^2} - \frac{\|z\|^2}{2\sdata^2} \right\} \dd \mu(z) .
\end{align*}
For $u$ in a fixed compact set, every polynomial in $z$ multiplied by
$\exp\{u^\top z/\sdata^2-\|z\|^2/(2\sdata^2)\}$ is bounded uniformly in $z$.
Since $\mu$ is a probability measure, differentiation under the integral
is therefore justified at every order.
Proceeding as in Step 1 of the proof of Lemma \ref{lem:log_int_der_gaus},
we deduce that, for every integer $k\geq2$,
\begin{align*}
    &\nabla^k \log\integral{\R^d} \exp\left\{ \frac{u^\top z}{\sdata^2} - \frac{\|z\|^2}{2\sdata^2}\right\} \dd \mu(z)
    \\&
    = \left(\sdata^2 Z(u) \right)^{-k} \integral{\R^d} \dots \integral{\R^d} P_k \exp\left\{ \frac{u^\top(z_1 + \ldots + z_k)}{\sdata^2} - \sum_{j = 1}^k \frac{\|z_j\|^2}{2\sdata^2} \right\} \prod_{j = 1}^k \dd \mu(z_j),
\end{align*}
where
\[
    Z(u) = \integral{\R^d} \exp\left\{ \frac{u^\top z}{\sdata^2} - \frac{\|z\|^2}{2\sdata^2} \right\} \dd\mu(z)
\]
and $P_k(z_1, \dots, z_k)$ satisfies the recurrence
\[
    P_1 = z_1,
    \quad
    P_2 = \frac12 (z_1 - z_2) \otimes (z_1 - z_2),
\]
and
\begin{equation}
\label{eq:log_dens_der_pk_rec}
    P_{k + 1} = P_k \otimes(z_1 + \ldots + z_k - k z_{k + 1}),
    \quad k \geq 2.
\end{equation}
Unrolling \eqref{eq:log_dens_der_pk_rec} yields the closed form
\[
    P_k = 
    \frac12 (z_1 - z_2) \otimes (z_1 - z_2) \otimes \ldots
    \otimes \biggl( \, \sum_{j=1}^{k-1} z_j - (k-1)\,z_k\biggr).
\]
Equivalently, defining
\begin{align}
    \label{eq:cond_marg_def}
    \dd \pi_u(z) = \frac{1}{Z(u)} \exp\left\{ \frac{u^\top z}{\sdata^2} - \frac{\|z\|^2}{2\sdata^2}\right\} \dd \mu(z),
\end{align}
we observe that 
\begin{align}
    \label{eq:log_dens_deriv_exp_red}
    &\notag
    \nabla^k \log\integral{\R^d} \exp\left\{ \frac{u^\top z}{\sdata^2} - \frac{\|z\|^2}{2\sdata^2}\right\} \dd \mu(z)
    \\&
    = \frac{1}{2\sdata^{2k}} \, \E \left[ (X_1 - X_2) \otimes (X_1 - X_2) \otimes \ldots
    \otimes \biggl( \, \sum_{j=1}^{k-1} X_j - (k-1) \, X_k\biggr) \right],
\end{align}
where $X_1, \dots, X_k$ are independent samples from $\pi_u$.

\noindent
\textbf{Step 2: reduction to sub-Gaussian conditional.}\quad We now reformulate the task using the notation above. Completing the exponent of \eqref{eq:cond_marg_def} gives
\begin{align*}
    \dd \pi_u(z)
    = \frac{1}{Z(u)} \exp\left\{ \frac{u^\top z}{\sdata^2} - \frac{\|z\|^2}{2\sdata^2}\right\} \dd \mu(z)
    \propto \exp\left\{ - \frac{\|u - z\|^2}{2\sdata^2}\right\} \dd \mu(z).
\end{align*}
Let us introduce $S = \eta + Z$, where $\eta \sim \cN(0, \sdata^2 I_d)$ is independent of $Z \sim \mu$.
Then $\pi_u$ can be considered as the conditional Law of $Z$ given $S = u$. Applying Lemma \ref{lem:cond_density_bound} to the pair $(\eta, Z)$ with $R = \|u\| + 2\tau\sqrt{d}$, one obtains
\begin{align*}
    \bigl\|\eta - \E[\eta \mid S = u]\bigr\|_{\psi_2(\cdot \mid S = u)} \leq 6R + 6\sdata,
    \qquad
    \bigl\|\E[\eta \mid S = u]\bigr\| \le 3R + 3\sdata.
\end{align*}
Here and further in the proof, $\|\cdot\|_{\psi_2(\cdot \mid S = u)}$ stands for the Orlicz norm with respect to the conditional distribution given $S = u$.
Let us denote
\[
    \mu_T = \E[Z \mid S = u]
    \quad \text{and} \quad \tilde \sigma = 6(R+\sdata)>0.
\]
Since $Z = u - \eta$ whenever $S = u$, we observe that
\begin{align}
\label{eq:cond_Z_bounds}
    \bigl\|Z-\mu_T\bigr\|_{\psi_2(\cdot\mid S=u)}\leq\tilde\sigma,
    \quad \text{and} \quad
    \|\mu_T\| \leq \|u\| + 3R + 3\sdata \leq 4 R + 4\sdata.
\end{align}
Introducing $Z_j = (X_j - \mu_T) / \tilde\sigma$, $j \in \{1, \dots, k\}$, we find that
\begin{align*}
    &
    \E (X_1 - X_2) \otimes (X_1 - X_2) \otimes \dots \otimes \left(\sum_{j = 1}^{k - 1}X_j - (k - 1) X_k \right)
    \\&
    = \tilde\sigma^k \; \E (Z_1 - Z_2) \otimes (Z_1 - Z_2) \otimes \dots \otimes \left(\sum_{j = 1}^{k - 1}Z_j - (k - 1)Z_k\right).
\end{align*}
Since $Z_1, \dots, Z_k$ are independent, centered-sub-Gaussian with the conditional Orlicz norm at most $1$, we can apply Lemma \ref{lem:gaus_expect} and obtain that
\begin{align*}
    \left\| \E (X_1 - X_2) \otimes (X_1 - X_2) \otimes \dots \otimes \left(\sum_{j = 1}^{k - 1}X_j - (k - 1) X_k \right) \right\|_\infty
    \leq 
    \left( 6 \tilde\sigma \sqrt{C} \right)^k k! ,
\end{align*}
where $C \geq 1$ is an absolute constant.
Substituting this bound into \eqref{eq:log_dens_deriv_exp_red} and using \eqref{eq:cond_Z_bounds}, we deduce that
\begin{align*}
    \left\| \nabla^k \log\integral{\R^d} \exp\left\{ \frac{u^\top z}{\sdata^2} - \frac{\|z\|^2}{2\sdata^2}\right\} \dd \mu(z) \right\|_\infty
    &
    \leq \left( \frac{36 (R + \sdata) \sqrt{C}}{\sdata^2} \right)^k k!
    \\&
    = \left( \frac{36 (\|u\| + 2\tau \sqrt{d} + \sdata) \sqrt{C}}{\sdata^2} \right)^k k!.
\end{align*}
\myendproof

\subsection{Proof of Lemma \ref{lem:log_int_der_gaus}}
\label{sec:lem_log_int_der_gaus_proof}

Let us fix arbitrary $i_1, \dots, i_k \in \{1, \dots, d\}$.
By the definition of the $\ell_\infty$-norm, it is enough to show that
\begin{align*}
    \left| \left( \nabla^k \log \integral{\R^d} \sfq_{||}(y \,|\, x) \frac{\rho_{||}(x) \, \dd x}{\cT_{||}[\nu_{||}](x)} \right)_{i_1, \dots, i_k} \right|
    &
    \leq \exp\left\{ k (2 M + 3) + \frac{k e^{-2bT} \|y\|^2}{2 \big(\gamma \sigma_T^2 \big)^2}\right\}
    \\&\quad
    \cdot \left( \frac{3\cdot2^{2+d/2}e^{-bT}\sqrt{C}}{\gamma\sigma_T^2} \right)^k k!.
\end{align*}
We divide the proof into six steps.
The Gaussian envelope proved in Step 3, together with $\alpha(T)\leq1/2$
as verified in Step 6, dominates every $y$-derivative of $\Phi(x,y)$
locally uniformly in $y$ by an integrable polynomial times a Gaussian
in $x$. Thus all differentiations under the integral below are justified.

\noindent
\textbf{Step 1: exact expression for the $k$-th derivative.}
\quad
Let us introduce
\[
    \Phi(x, y) = \exp\left\{ \frac{\|y\|^2}{2 \gamma \sigma_T^2} \right\} \cdot \frac{\sfq_{||}(y \,|\, x) \rho_{||} (x)}{\cT_{||}[\nu_{||}](x)},
    \quad
    Z(y) = \int\limits_{\R^d} \Phi(x, y) \, \dd x,
    \quad x, y \in \R^d,
\]
and show that, for any $y \in \R^d$,
\begin{align*}
    &
    \nabla^k \left( \frac{\|y\|^2}{2 \gamma \sigma_T^2} + \log \integral{\R^d} \sfq_{||}(y \,|\, x) \frac{\rho_{||}(x) \, \dd x}{\cT_{||}[\nu_{||}](x)}  \right)
    \\&
    = \left( \frac{e^{-bT} / (\gamma \sigma_T^2)}{Z(y)} \right)^k \integral{\R^d} \ldots \integral{\R^d} P_k \prod\limits_{j = 1}^k \Phi(x_j, y) \, \dd x_1 \ldots \dd x_k,
\end{align*}
where
\[
    P_1=x_1,\qquad
    P_2=P_2(x_1,x_2)=\frac12(x_1-x_2)\otimes(x_1-x_2),
\]
and
\begin{equation}
    \label{eq:pk_recursion}
    P_{k + 1} = P_{k + 1}(x_1, \dots, x_{k + 1})
    = P_k(x_1, \dots, x_k) \otimes (x_1 + \ldots + x_k - k x_{k + 1}),\qquad k\geq2.
\end{equation}
Indeed, for $k = 1$, it holds that
\begin{align*}
    \nabla \left( \frac{\|y\|^2}{2 \gamma \sigma_T^2} + \log \integral{\R^d} \sfq_{||}(y \,|\, x) \frac{\rho_{||}(x) \, \dd x}{\cT_{||}[\nu_{||}](x)}  \right)
    &
    = \nabla \log \integral{\R^d} \Phi(x, y) \, \dd x
    \\&
    = \left( \; \integral{\R^d} \frac{\partial \Phi(x, y) }{\partial y} \, \dd x \right) \Big/ \left( \; \integral{\R^d} \Phi(x, y) \, \dd x \right)
    \\&
    = \left( \frac{e^{-bT} / (\gamma \sigma_T^2)}{Z(y)} \right) \integral{\R^d} x \; \Phi(x, y) \, \dd x.
\end{align*}
Differentiating the expression in the right-hand side once again, we obtain that
\begin{align*}
    &
    \nabla^2 \left( \frac{\|y\|^2}{2 \gamma \sigma_T^2} + \log \integral{\R^d} \sfq_{||}(y \,|\, x) \frac{\rho_{||}(x) \, \dd x}{\cT_{||}[\nu_{||}](x)}  \right)
    \\&
    = \left( \frac{e^{-bT} / (\gamma \sigma_T^2)}{Z(y)} \right)^2 \integral{\R^d} \integral{\R^d} x_1 \otimes (x_1 - x_2) \prod\limits_{j = 1}^2 \Phi(x_j, y) \, \dd x_1 \dd x_2.
\end{align*}
Due to exchangeability of $x_1$ and $x_2$ in the expression in the right-hand side, it holds that
\begin{align*}
    &
    \nabla^2 \left( \frac{\|y\|^2}{2 \gamma \sigma_T^2} + \log \integral{\R^d} \sfq_{||}(y \,|\, x) \frac{\rho_{||}(x) \, \dd x}{\cT_{||}[\nu_{||}](x)}  \right)
    \\&
    = \frac12 \left( \frac{e^{-bT} / (\gamma \sigma_T^2)}{Z(y)} \right)^2 \integral{\R^d} \integral{\R^d} x_2 \otimes (x_2 - x_1) \prod\limits_{j = 1}^2 \Phi(x_j, y) \, \dd x_1 \dd x_2
    \\&\quad
    + \frac12 \left( \frac{e^{-bT} / (\gamma \sigma_T^2)}{Z(y)} \right)^2 \integral{\R^d} \integral{\R^d} x_1 \otimes (x_1 - x_2) \prod\limits_{j = 1}^2 \Phi(x_j, y) \, \dd x_1 \dd x_2
    \\&
    = \frac12 \left( \frac{e^{-bT} / (\gamma \sigma_T^2)}{Z(y)} \right)^2 \integral{\R^d} \integral{\R^d} (x_1 - x_2) \otimes (x_1 - x_2) \prod\limits_{j = 1}^2 \Phi(x_j, y) \, \dd x_1 \dd x_2.
\end{align*}
Proceeding by the induction, we observe that, for any $k \geq 2$, 
\begin{align*}
    &
    \nabla^{k + 1} \left( \frac{\|y\|^2}{2 \gamma \sigma_T^2} + \log \integral{\R^d} \sfq_{||}(y \,|\, x) \frac{\rho_{||}(x) \, \dd x}{\cT_{||}[\nu_{||}](x)}  \right)
    \\&
    = \nabla_y\left[ \left( \frac{e^{-bT} / (\gamma \sigma_T^2)}{Z(y)} \right)^k \integral{\R^d} \ldots \integral{\R^d} P_k \prod\limits_{j = 1}^k \Phi(x_j, y) \; \dd x_1 \ldots \dd x_k \right]
    \\&
    = \left( \frac{e^{-bT} / (\gamma \sigma_T^2)}{Z(y)} \right)^k \Bigg[ \; \integral{\R^d} \ldots \integral{\R^d} P_k \otimes \frac{\partial}{\partial y} \left( \prod\limits_{j = 1}^k \Phi(x_j, y) \right) \, \dd x_1 \ldots \dd x_k 
    \\&\quad
    - \frac{k}{Z(y)} \integral{\R^d} \ldots \integral{\R^d} P_k \prod\limits_{j = 1}^k \Phi(x_j, y) \, \dd x_1 \ldots \dd x_k \otimes \nabla Z(y) \Bigg].
\end{align*}
Since
\begin{align*}
    &
    \integral{\R^d} \ldots \integral{\R^d} P_k \otimes \frac{\partial}{\partial y} \left( \prod\limits_{j = 1}^k \Phi(x_j, y) \right) \, \dd x_1 \ldots \dd x_k
    \\&
    = \left( \frac{e^{-bT}}{\gamma \sigma_T^2} \right) \integral{\R^d} \ldots \integral{\R^d} P_k \otimes (x_1 + \ldots + x_k) \prod\limits_{j = 1}^k \Phi(x_j, y) \, \dd x_1 \ldots \dd x_k
    \\&
    = \frac{e^{-bT} / (\gamma \sigma_T^2)}{Z(y)} \integral{\R^d} \ldots \integral{\R^d} P_k \otimes (x_1 + \ldots + x_k) \prod\limits_{j = 1}^{k + 1} \Phi(x_j, y) \, \dd x_1 \ldots \dd x_{k + 1}
\end{align*}
and
\begin{align*}
    &
    \frac{1}{Z(y)} \integral{\R^d} \ldots \integral{\R^d} P_k \prod\limits_{j = 1}^k \Phi(x_j, y) \, \dd x_1 \ldots \dd x_k \otimes \nabla Z(y)
    \\&
    = \frac{1}{Z(y)} \integral{\R^d} \ldots \integral{\R^d} P_k \prod\limits_{j = 1}^k \Phi(x_j, y) \, \dd x_1 \ldots \dd x_k \otimes \integral{\R^d} \frac{\partial \Phi(x_{k + 1}, y)}{\partial y} \, \dd x_{k + 1}
    \\&
    = \frac{e^{-bT} / (\gamma \sigma_T^2)}{Z(y)} \integral{\R^d} \ldots \integral{\R^d} P_k \otimes x_{k + 1} \prod\limits_{j = 1}^{k + 1} \Phi(x_j, y) \, \dd x_1 \ldots \dd x_{k + 1},
\end{align*} 
we conclude that
\begin{align*}
    &
    \nabla^{k + 1} \left( \frac{\|y\|^2}{2 \gamma \sigma_T^2} + \log \integral{\R^d} \sfq_{||}(y \,|\, x) \frac{\rho_{||}(x) \, \dd x}{\cT_{||}[\nu_{||}](x)} \right)
    \\&
    = \left( \frac{e^{-bT} / (\gamma \sigma_T^2)}{Z(y)} \right)^{k + 1} \integral{\R^d} \ldots \integral{\R^d} P_k \otimes (x_1 + \ldots + x_k - k x_{k + 1}) \prod\limits_{j = 1}^{k + 1} \Phi(x_j, y) \, \dd x_1 \ldots \dd x_{k + 1}
    \\&
    = \left( \frac{e^{-bT} / (\gamma \sigma_T^2)}{Z(y)} \right)^{k + 1} \integral{\R^d} \ldots \integral{\R^d} P_{k + 1} \prod\limits_{j = 1}^{k + 1} \Phi(x_j, y) \, \dd x_1 \ldots \dd x_{k + 1}.
\end{align*}
It remains to note that $\nabla^k \|y\|^2$ equals to zero for all $k \geq 3$. For this reason,
\begin{align}
    \label{eq:kth_derivative}
    &\notag
    \nabla^k \left( \log \integral{\R^d} \sfq_{||}(y \,|\, x) \frac{\rho_{||}(x) \, \dd x}{\cT_{||}[\nu_{||}](x)}  \right)
    \\&
    = \left( \frac{e^{-bT} / (\gamma \sigma_T^2)}{Z(y)} \right)^k \integral{\R^d} \ldots \integral{\R^d} P_k \prod\limits_{j = 1}^k \Phi(x_j, y) \, \dd x_1 \ldots \dd x_k
\end{align}
for all $k \geq 3$ and $y \in \R^d$.

\medskip

\noindent
\textbf{Step 2: lower bound on $Z(y)$.} Let us recall that, according to the definition,
\begin{align*}
    Z(y)
    &
    = \int\limits_{\R^d} \Phi(x, y) \, \dd x
    = \int\limits_{\R^d} \exp\left\{ \frac{\|y\|^2}{2 \gamma \sigma_T^2} \right\} \cdot \frac{\sfq_{||}(y \,|\, x) \rho_{||} (x)}{\cT_{||}[\nu_{||}](x)} \, \dd x
    \\&
    = (2\pi)^{-d} (\gamma \sigma_T^2)^{-d/2} \int\limits_{\R^d} \exp\left\{ \frac{e^{-bT} y^\top x}{\gamma \sigma_T^2} - \frac{\|x\|^2}2 \left(1 + \frac{e^{-2bT}}{\gamma \sigma_T^2} \right) \right\} \frac{\dd x}{\cT_{||}[\nu_{||}](x)}.
\end{align*}
Taking into account that $\nu_{||} \leq e^{M}$ due to the conditions of the lemma, we obtain that $\cT_{||}[\nu_{||}](x) \leq e^{M}$ for all $x \in \R^d$, and then
\[
    Z(y) \geq (2\pi)^{-d} (\gamma \sigma_T^2)^{-d/2} e^{-M} \int\limits_{\R^d} \exp\left\{ \frac{e^{-bT} y^\top x}{\gamma \sigma_T^2} - \frac{\|x\|^2}2 \left(1 + \frac{e^{-2bT}}{\gamma \sigma_T^2} \right) \right\} \, \dd x.
\]
The integral in the right-hand side admits a closed form. Namely, it straightforward to check that
\begin{align*}
    &
    \int\limits_{\R^d} \exp\left\{ \frac{e^{-bT} y^\top x}{\gamma \sigma_T^2} - \frac{\|x\|^2}2 \left(1 + \frac{e^{-2bT}}{\gamma \sigma_T^2} \right) \right\} \, \dd x
    \\&
    = (2\pi)^{d/2} \left(1 + \frac{e^{-2bT}}{\gamma \sigma_T^2} \right)^{-d/2} \exp\left\{ \frac{e^{-2bT} \|y\|^2}{2 \gamma \sigma_T^2 (\gamma \sigma_T^2 + e^{-2bT})} \right\}.
\end{align*}
Hence, it holds that
\begin{equation}
    \label{eq:z_lower_bound}
    Z(y) \geq (2\pi)^{-d/2} \left(\gamma \sigma_T^2 + e^{-2bT} \right)^{-d/2} e^{-M}  \exp\left\{ \frac{e^{-2bT} \|y\|^2}{2 \gamma \sigma_T^2 (\gamma \sigma_T^2 + e^{-2bT})} \right\}.
\end{equation}

\medskip

\noindent
\textbf{Step 3: upper bound on $\Phi(x,y)$.}\quad
Let
\begin{equation}
    \label{eq:ck}
    \cK(T)=(1-e^{-2bT})^{-5e^2\sqrt d}
    \exp\{2e^2\sqrt d\,\arcsin(e^{-bT})\}
\end{equation}
and
\begin{equation}
    \label{eq:alpha}
    \alpha(T)=\frac{2be^2\cK(T)}{\gamma\sqrt d}\arcsin(e^{-bT}).
\end{equation}
Write $a=\gamma/(2b)$, $r=e^{-bT}$, $q=a(1-r^2)=\gamma\sigma_T^2$,
and $K=\cK(T)$, and let $m=\cN(0,aI_d)$ and $Q_x=\cN(rx,qI_d)$.
For $f=\log\nu_{||}$, the assumptions give $f\in L^1(m)$,
$\int f\,\dd m=0$, and $M\geq0$.
Since $\log(\dd Q_x/\dd m)$ is an integrable quadratic function,
Jensen's inequality gives
\begin{align*}
    \log\cT_{||}[\nu_{||}](x)
    &=\log\int\exp\left\{f+\log\frac{\dd Q_x}{\dd m}\right\}\,\dd m
      \geq-\KL(m,Q_x)
      =-C_T-\frac{r^2\|x\|^2}{2q},
\end{align*}
where
\[
    C_T=\frac d2\left(\frac a q-1+\log\frac q a\right)
       =\frac d2\left(\frac{r^2}{1-r^2}+\log(1-r^2)\right).
\]
The second condition in \eqref{eq:der_bt_cond} implies $r^2\leq1/2$ and
\[
    2e^2\sqrt d\,r\leq\log K
    \leq10e^2\sqrt d\,r^2+4e^2\sqrt d\,r
    \leq14e^2\sqrt d\,r\leq\log2.
\]
Consequently,
\[
    C_T\leq dr^2\leq\frac{(\log K)^2}{4e^4}\leq2K\log K,
    \qquad
    \frac{r^2}{q}\leq\frac{2r^2}{a}
    \leq\frac{e^2r}{a\sqrt d}\leq\alpha(T),
\]
where we used $\sqrt d\,r\leq\log2/(2e^2)<e^2/2$ and
$\alpha(T)=e^2K\arcsin(r)/(a\sqrt d)$.
Since $K\geq1$ and $M\geq0$, we conclude that
\[
    \frac{e^{M}}{\cT_{||}[\nu_{||}](x)}
    \leq\exp\left\{M+C_T+\frac{r^2\|x\|^2}{2q}\right\}
    \leq\exp\left\{\frac{\alpha(T)\|x\|^2}2
             +\bigl(M+2\log\cK(T)\bigr)\cK(T)\right\}.
\]
It follows that
\begin{align}
    \label{eq:phi_upper_bound}
    &\notag
    (2\pi)^d(\gamma\sigma_T^2)^{d/2}e^{M}\Phi(x,y)
    \\
    &\leq\exp\left\{
      \frac{e^{-bT}y^\top x}{\gamma\sigma_T^2}
      -\frac12\left(1-\alpha(T)+\frac{e^{-2bT}}{\gamma\sigma_T^2}\right)\|x\|^2
      +\bigl(M+2\log\cK(T)\bigr)\cK(T)
    \right\}.
\end{align}

\medskip
\noindent\textbf{Step 4: summing up the two upper bounds.}
\quad The expression \eqref{eq:kth_derivative} yields that
\begin{align*}
    &
    \left| \left( \nabla^k \log \integral{\R^d} \sfq_{||}(y \,|\, x) \frac{\rho_{||}(x) \, \dd x}{\cT_{||}[\nu_{||}](x)} \right)_{i_1, \dots, i_k} \right|
    \\&
    = \left| \left( \frac{e^{-bT} / (\gamma \sigma_T^2)}{Z(y)} \right)^k \integral{\R^d} \ldots \integral{\R^d} (P_k)_{i_1, \dots, i_k} \, \prod\limits_{j = 1}^k \Phi(x_j, y) \, \dd x_1 \ldots \dd x_k \right|
    \\&
    \leq \left( \frac{e^{-bT} / (\gamma \sigma_T^2)}{Z(y)} \right)^k \integral{\R^d} \ldots \integral{\R^d} \big|(P_k)_{i_1, \dots, i_k} \big| \prod\limits_{j = 1}^k \Phi(x_j, y) \, \dd x_1 \ldots \dd x_k.
\end{align*}
In view of \eqref{eq:z_lower_bound} and \eqref{eq:phi_upper_bound}, it holds that
\begin{align*}
    &
    \left| \left( \nabla^k  \log \integral{\R^d} \sfq_{||}(y \,|\, x) \frac{\rho_{||}(x) \, \dd x}{\cT_{||}[\nu_{||}](x)}  \right)_{i_1, \dots, i_k} \right|
    \\&
    \leq (2\pi)^{-kd/2} \left( \frac{\gamma \sigma_T^2 + e^{-2bT}}{\gamma \sigma_T^2} \right)^{kd/2}
    \\&\quad\cdot\exp\left\{ k \big(M + 2 \log \cK(T) \big) \cK(T) -\frac{k e^{-2bT} \|y\|^2}{2 \gamma \sigma_T^2 (\gamma \sigma_T^2 + e^{-2bT})}\right\}
    \\&\quad
    \cdot \left(\frac{e^{-bT}}{\gamma\sigma_T^2}\right)^k \integral{\R^d} \ldots \integral{\R^d} \big|(P_k)_{i_1, \dots, i_k} \big|
    \\&\quad\cdot \prod\limits_{j = 1}^k \exp\left\{ \frac{e^{-bT} y^\top x_j}{\gamma \sigma_T^2} - \frac12 \left(1 - \alpha(T) + \frac{e^{-2bT}}{\gamma \sigma_T^2} \right) \|x_j\|^2 \right\} \, \dd x_1 \ldots \dd x_k.
\end{align*}
Let us note that, due to the definition of $P_k$ (see \eqref{eq:pk_recursion}), it is invariant under the substitutions
\begin{equation}
    \label{eq:breve_sigma}
    x_j := x_j - \frac{e^{-bT} y}{\breve\sigma^2},
    \quad 1 \leq j \leq k,
    \quad \text{where} \quad
    \breve\sigma^2 = \gamma \sigma_T^2 + e^{-2bT} - \gamma \sigma_T^2 \, \alpha(T).
\end{equation}
This implies that
\begin{align*}
    &
    \left| \left( \nabla^k \log \integral{\R^d} \sfq_{||}(y \,|\, x) \frac{\rho_{||}(x) \, \dd x}{\cT_{||}[\nu_{||}](x)} \right)_{i_1, \dots, i_k} \right|
    \\&
    \leq (2\pi)^{-kd/2} \left( \frac{\gamma \sigma_T^2 + e^{-2bT}}{\gamma \sigma_T^2} \right)^{kd/2}
    \\&\quad\cdot\exp\left\{ k \big(M + 2 \log \cK(T) \big) \cK(T) -\frac{k e^{-2bT} \|y\|^2}{2 \gamma \sigma_T^2 (\gamma \sigma_T^2 + e^{-2bT})} \right\}
    \\&
    \cdot \left( \frac{e^{-bT}}{\gamma \sigma_T^2} \right)^k \integral{\R^d} \ldots \integral{\R^d} \big|(P_k)_{i_1, \dots, i_k} \big|
    \\&\quad\cdot \prod\limits_{j = 1}^k \exp\left\{ \frac{e^{-2bT} \|y\|^2}{2 \gamma \sigma_T^2 \breve\sigma^2} - \frac{\breve\sigma^2 \|x_j\|^2 }{2 \gamma \sigma_T^2} \right\} \, \dd x_1 \ldots \dd x_k.
\end{align*}
Using the identity
\[
    \frac{e^{-2bT} \|y\|^2}{2 \gamma \sigma_T^2 \breve\sigma^2} - \frac{e^{-2bT} \|y\|^2}{2 \gamma \sigma_T^2 (\gamma \sigma_T^2 + e^{-2bT})}
    = \frac{e^{-2bT} \alpha(T) \|y\|^2}{2 (\gamma \sigma_T^2 + e^{-2bT}) \breve\sigma^2}
\]
and rearranging the terms, we obtain that 
\begin{align}
    \label{eq:kth_derivative_bound}
    &\notag
    \left| \left( \nabla^k \log \integral{\R^d} \sfq_{||}(y \,|\, x) \frac{\rho_{||}(x) \, \dd x}{\cT_{||}[\nu_{||}](x)} \right)_{i_1, \dots, i_k} \right|
    \\&
    \leq \exp\left\{ k \big(M + 2 \log \cK(T) \big) \cK(T) + \frac{k e^{-2bT} \alpha(T) \|y\|^2}{2 (\gamma \sigma_T^2 + e^{-2bT}) \breve\sigma^2}\right\}
    \\&\quad\notag
    \cdot \left( \frac{e^{-bT}}{\gamma \sigma_T^2} \right)^k \left( \frac{\gamma \sigma_T^2 + e^{-2bT}}{\breve\sigma^2} \right)^{kd/2} \, \cJ_k,
\end{align}
where
\begin{align}
    \label{eq:jk}
    \cJ_k &= (2\pi)^{-kd/2} \left( \frac{\gamma \sigma_T^2}{\breve\sigma^2} \right)^{-kd/2} \integral{\R^d} \ldots \integral{\R^d} \big|(P_k)_{i_1, \dots, i_k} \big|
    \\&\quad\cdot\notag \prod\limits_{j = 1}^k \exp\left\{ - \frac{\breve \sigma^2 \|x_j\|^2}{2 \gamma \sigma_T^2} \right\} \, \dd x_1 \ldots \dd x_k.
\end{align}

\noindent\textbf{Step 5: evaluation of $\cJ_k$.}
\quad
The integral $\cJ_k$ defined in \eqref{eq:jk} can be considered as the expectation of a function of $k$ i.i.d. Gaussian random vectors. Indeed, let $\eta_1, \dots, \eta_k \sim \cN(0, \gamma \sigma_T^2 \breve\sigma^{-2} I_d)$, where $\sigma_T$ and $\breve\sigma$ are given by \eqref{eq:markov_kernel} and \eqref{eq:breve_sigma}, respectively.
Unrolling the recursion \eqref{eq:pk_recursion} for $P_k$, we arrive at 
\[
    \cJ_k = \frac12\,\E \left| (\eta_1 - \eta_2)_{i_1} (\eta_1 - \eta_2)_{i_2} (\eta_1 + \eta_2 - 2\eta_3)_{i_3} \dots \big(\eta_1 + \ldots + \eta_{k - 1} - (k - 1) \eta_k \big)_{i_k} \right|.
\]
Let $\xi_1, \dots, \xi_k$ be independent standard Gaussian random vectors in $\R^d$, and let us introduce
\[
    S_j = \sum\limits_{i = 1}^j \xi_i,
    \quad j \in \{1, \dots, k\}.
\]
Then
\[
    \cJ_k = \frac12\left( \frac{\gamma \sigma_T^2}{\breve\sigma^2} \right)^{k/2} \E \left| (S_1 - \xi_2)_{i_1} (S_1 - \xi_2)_{i_2} (S_2 - 2\xi_3)_{i_3} \dots \big(S_{k - 1} - (k - 1) \xi_k \big)_{i_k} \right|.
\]
For $1\leq j\leq k-1$, put $D_j=S_j-j\xi_{j+1}$.
These are independent centered Gaussian vectors with
$\operatorname{Cov}(D_j)=j(j+1)I_d$: their cross-covariances vanish.
Independence and the Cauchy--Schwarz inequality therefore yield
\begin{align*}
    \E\left|D_{1,i_1}D_{1,i_2}
          \prod_{j=2}^{k-1}D_{j,i_{j+1}}\right|
    &\leq 2\prod_{j=2}^{k-1}\sqrt{j(j+1)}
      =\sqrt{2(k-1)!\,k!}\leq k!.
\end{align*}
Consequently,
$\cJ_k\leq\frac12(\gamma\sigma_T^2/\breve\sigma^2)^{k/2}k!$.
Due to \eqref{eq:kth_derivative_bound}, this implies that
\begin{align}
    \label{eq:deriv_almost_fin}
    &\notag
    \left| \left( \nabla^k \log \integral{\R^d} \sfq_{||}(y \,|\, x) \frac{\rho_{||}(x) \, \dd x}{\cT_{||}[\nu_{||}](x)} \right)_{i_1, \dots, i_k} \right|
    \\&
    \leq \exp\left\{ k \big(M + 2 \log \cK(T) \big) \cK(T) + \frac{k e^{-2bT} \alpha(T) \|y\|^2}{2 (\gamma \sigma_T^2 + e^{-2bT}) \breve\sigma^2}\right\}
    \\&\quad\notag
    \cdot \frac12\left( \frac{e^{-bT}}{\sqrt{\gamma\sigma_T^2\,\breve\sigma^2}} \right)^k \left( \frac{\gamma \sigma_T^2 + e^{-2bT}}{\breve\sigma^2} \right)^{kd/2} k!.
\end{align}

\noindent
\textbf{Step 6: simplifying the bound \eqref{eq:deriv_almost_fin}.}\quad 
On this step, we aim to make the bound \eqref{eq:deriv_almost_fin} more user-friendly using the inequalities \eqref{eq:der_bt_cond}.
First, note that
\[
    \arcsin x \leq 2x
    \quad \text{for all $x \in [0, 1]$}
    \quad \text{and} \quad
    -\log(1 - x) \leq 2x
    \quad \text{for all $x \in [0, 1/2]$.}
\]
Then, in view of the definition of $\cK(T)$ (see \eqref{eq:ck}), we have that
\begin{align}
    \label{eq:cK_T_bound}
    1 \leq \cK(T)
    &\notag
    \leq \exp\left\{ -5e^2\sqrt{d} \log(1 - e^{-2bT}) + 2e^2\sqrt{d} \arcsin(e^{-bT}) \right\}
    \\&
    \leq \exp\{14 e^2 \sqrt{d} e^{-bT}\}
    \leq 2,
\end{align}
where the last inequality uses the conditions on $bT$ from \eqref{eq:der_bt_cond}. Since the normalization and the upper bound imply $M\geq0$, this yields
\[
    \big( M + 2 \log \cK(T) \big) \cK(T)
    \leq 2 M + 4 \log 2
    \leq 2 M + 3.
\]
On the other hand, $\alpha(T)$, given by \eqref{eq:alpha}, satisfies
\[
    \alpha(T)
    = \frac{e^2 \cK(T) (1 - e^{-2bT}) \arcsin(e^{-bT})}{\gamma \sigma_T^2 \sqrt{d}}
    \leq \frac{4e^{2 - bT}}{\gamma \sigma_T^2 \sqrt{d}}
    \leq \frac12.
\]
The last inequality is also due to \eqref{eq:der_bt_cond}. This implies that
\begin{align}
    \label{eq:alpha_bound_corollary_1}
    \frac{\alpha(T)}{(\gamma \sigma_T^2 + e^{-2bT}) \breve\sigma^2}
    &
    = \frac{\alpha(T)}{(\gamma \sigma_T^2 + e^{-2bT}) \big(\gamma \sigma_T^2 + e^{-2bT} - \alpha(T) \gamma \sigma_T^2 \big)}
    \\&
    \leq \frac{\alpha(T)}{1 - \alpha(T)} \cdot \frac1{\big( \gamma \sigma_T^2 \big)^2}
    \leq \frac1{\big( \gamma \sigma_T^2 \big)^2}
\end{align}
and that
\begin{equation}
    \label{eq:alpha_bound_corollary_2}
    \frac{\gamma \sigma_T^2 + e^{-2bT}}{\breve \sigma^2}
    = \frac{\gamma \sigma_T^2 + e^{-2bT}}{\gamma \sigma_T^2 + e^{-2bT} - \gamma \sigma_T^2 \alpha(T)}
    \leq \frac1{1 - \alpha(T)}
    \leq 2.
\end{equation}
Since $\breve\sigma^2\geq\gamma\sigma_T^2/2$, we also have
\[
    \frac{e^{-bT}}{\sqrt{\gamma\sigma_T^2\,\breve\sigma^2}}
    \leq\frac{\sqrt2\,e^{-bT}}{\gamma\sigma_T^2}.
\]
Substituting this bound, \eqref{eq:alpha_bound_corollary_1}, and
\eqref{eq:alpha_bound_corollary_2} into \eqref{eq:deriv_almost_fin}
gives the following (weaker) bound for any absolute constant $C\geq1$:
\begin{align*}
    \left| \left( \nabla^k \log \integral{\R^d} \sfq_{||}(y \,|\, x) \frac{\rho_{||}(x) \, \dd x}{\cT_{||}[\nu_{||}](x)} \right)_{i_1, \dots, i_k} \right|
    &
    \leq \exp\left\{ k (2 M + 3) + \frac{k e^{-2bT} \|y\|^2}{2 \big(\gamma \sigma_T^2 \big)^2}\right\}
    \\&\quad
    \cdot \left( \frac{3 \cdot 2^{2 + d/2} e^{-bT} \sqrt{C}}{\gamma\sigma_T^2} \right)^k k!.
\end{align*}
The proof is now complete.

\myendproof

\subsection{Proof of Lemma \ref{lem:cond_density_bound}}
\label{sec:lem_cond_density_bound_proof}

\textbf{Step 1: Gaussian conjugacy.}
Let us introduce
\[
    f_S(t) = \E \varphi_\sigma(t - Y),
    \quad \text{where} \quad
    \varphi_\sigma(u) = (2\pi\sigma^2)^{-d/2} e^{-\|u\|^2/(2\sigma^2)}
\]
Note that $f_S(s) > 0$ for all $s \in \R^d$.
For any $\lambda \in \R^d$ and a test function $g : \R^d \to \R$, the change of variables $u = x + Y$ gives
\[
    \E \left[ g(S)e^{\lambda^\top X} \right]
    = \E \left[ \; \integral{\R^d} e^{\lambda^\top x} g(x+Y) \varphi_\sigma(x) \, \dd x \right]
    = \integral{\R^d} g(u) \, \E \left[ e^{ \lambda^\top (u - Y)} \varphi_\sigma(u - Y) \right] \, \dd u .
\]
Hence,
\begin{align*}
    f_S(u) \, \E \left[ e^{\lambda^\top X} \mid S = u \right]
    = \E\left[ e^{\lambda^\top (u - Y)} \varphi_\sigma(u-Y) \right] .
\end{align*}
Using the identity
\begin{align*}
    e^{\lambda^\top v} \varphi_\sigma(v)
    = e^{\sigma^2\|\lambda\|^2/2} \varphi_\sigma(v - \sigma^2\lambda),
\end{align*}
we obtain that
\begin{equation}
\label{eq:mgf_subgaussian_conditional}
    \E[e^{\lambda^\top X} \mid S = s]
    = e^{\sigma^2\|\lambda\|^2/2} \frac{f_S(s - \sigma^2\lambda)}{f_S(s)},
    \quad \lambda \in \R^d.
\end{equation}

\textbf{Step 2: two-sided control of $f_S$.}
Trivially,
\begin{align*}
    f_S(t) \le \|\varphi_\sigma\|_\infty = (2\pi\sigma^2)^{-d/2} .
\end{align*}
On the other hand, by Markov's inequality,
\begin{align*}
    \p(\|Y\| \le 2\tau\sqrt{d})
    = \p(\|Y\|^2 \le 4d\tau^2)
    \ge 3/4 .
\end{align*}
On the event $\{ \|Y\| \le 2\tau\sqrt{d} \}$, we have $\|s - Y\| \le \|s\| + 2\tau\sqrt{d} = R$. Because $\varphi_\sigma$ is radially decreasing,
\begin{align*}
    \varphi_\sigma(s - Y)
    \ge (2\pi\sigma^2)^{-d/2} e^{-R^2/(2\sigma^2)} .
\end{align*}
Thus,
\[
    f_S(s)
    \ge \E \left[ \varphi_\sigma(s - Y) \1\{\|Y\| \le 2\sqrt{d}\tau\} \right]
    \ge \frac{3}{4} (2\pi\sigma^2)^{-d/2} e^{-R^2/(2\sigma^2)} .
\]
Plugging both bounds into \eqref{eq:mgf_subgaussian_conditional} yields
\begin{align*}
    \E[e^{\lambda^\top X} \mid S = s]
    \le \frac{4}{3} e^{\sigma^2\|\lambda\|^2/2 + R^2/(2\sigma^2)}
\end{align*}
for all $\lambda \in \R^d$.

\textbf{Step 3: uncentered $\psi_2$-norm via Gaussian integration.}
Fix an arbitrary unit vector $v \in S^{d-1}$ and let $X_v = v^\top X$. Taking $\lambda = \alpha v$ gives the one-dimensional marginal bound
\begin{align*}
    \E[e^{\alpha X_v} \mid S=s]
    \le \frac{4}{3} e^{R^2/(2\sigma^2)} e^{\sigma^2\alpha^2/2}
\end{align*}
for all $\alpha \in \R$. To convert this inequality into an upper bound on the $\psi_2$-norm, we introduce a standard Gaussian random variable $\xi \sim \cN(0,1)$, which is independent of $X$ and $S$. Using the Gaussian integral identity
\[
    e^{X_v^2/(8\sigma^2)} = \E_\xi \left[e^{\xi X_v / (2\sigma)} \right],
\]
we can rewrite the conditional expectation as
\[
    \E \left[ e^{X_v^2/(8\sigma^2)} \mid S=s \right]
    = \E_{X|S} \left[ \E_\xi \, e^{\xi X_v / (2\sigma)} \mid S=s \right]
    = \E_\xi \, \E_{X|S} \left[ e^{\xi X_v / (2\sigma)} \mid S=s \right] .
\]
Using our moment generating function bound conditionally on $\xi$ with $\alpha = \xi / (2\sigma)$, so that $\sigma^2 \alpha^2/2 = \xi^2/8$, we obtain
\begin{align*}
    \E_{X|S} \left[ e^{\xi X_v / (2\sigma)} \mid S = s \right]
    \le \frac{4}{3} e^{R^2/(2\sigma^2)} e^{\xi^2 / 8} .
\end{align*}
Taking the outer expectation over $\xi$, we get
\[
    \E[e^{X_v^2/(8\sigma^2)} \mid S=s]
    \le \frac{4}{3} e^{R^2/(2\sigma^2)} \E_\xi \left[ e^{\xi^2 / 8} \right]
    = \frac{4 e^{R^2/(2\sigma^2)}}{3 \sqrt{1 - 1/4}} 
    = \left(\frac{4}{3}\right)^{3/2} e^{R^2/(2\sigma^2)}.
\]

Let $u \ge 1$ be a constant. Since the mapping $x \mapsto x^{1/u^2}$ is concave for $x \ge 0$, Jensen's inequality yields that
\begin{align*}
    \E[e^{X_v^2 / (8\sigma^2 u^2)} \mid S=s]
    \le \left( \E[e^{X_v^2 / (8\sigma^2)} \mid S=s] \right)^{1/u^2}
    \le \exp\left\{ \frac{R^2/\sigma^2 + 3\ln(4/3)}{2u^2} \right\} .
\end{align*}
We demand this bound to be at most $e^{1/2} < 2$, which requires $u^2 \ge R^2/\sigma^2 + 3\ln(4/3)$. Since $3\ln(4/3) \le 1$, we choose $u^2 = R^2/\sigma^2 + 1$, giving $8\sigma^2 u^2 = 8(R^2 + \sigma^2)$. Consequently,
\begin{align*}
    \E[e^{X_v^2 / (8(R^2+\sigma^2))} \mid S=s]
    \le e^{1/2} \le 2 .
\end{align*}
By the definition of the Orlicz norm, the uncentered variable obeys
\begin{align*}
    \|X_v\|_{\psi_2(\cdot \mid S=s)}
    \le \sqrt{8(R^2+\sigma^2)}
    \le \sqrt{8}(R+\sigma)
    \le 3R + 3\sigma .
\end{align*}

\textbf{Step 4: centering via the triangle inequality.}
Let $\mu_v = \E[X_v \mid S=s]$ denote the projection of the conditional mean. 
Let us take $K = 3R + 3\sigma$. By applying Jensen's inequality to the strictly convex mapping $x \mapsto \exp(x^2/K^2)$, we have
\begin{align*}
    \exp(\mu_v^2 / K^2)
    \le \E[\exp(X_v^2/K^2) \mid S=s]
    \le 2,
\end{align*}
which implies $|\mu_v| \le K \sqrt{\log 2}$. As a consequence, we have
\[
    \|\mu(s)\| = \sup\limits_{\|v\| = 1} |\mu_v| \leq K \sqrt{\log 2} \leq K.
\]
Since the $\psi_2$-norm of any constant $c$ is given by $|c|/\sqrt{\ln 2}$, we find that
\[
    \|\mu_v\|_{\psi_2(\cdot\mid S=s)} \le K = 3R + 3\sigma.
\]
The triangle inequality yields that
\begin{align*}
    \|X_v - \mu_v\|_{\psi_2(\cdot\mid S=s)}
    \le \|X_v\|_{\psi_2(\cdot\mid S=s)} + \|\mu_v\|_{\psi_2(\cdot\mid S=s)}
    \le 2K = 6R + 6\sigma.
\end{align*}
Taking the supremum over unit vectors $v$ establishes the final sub-Gaussian vector estimate:
\begin{align*}
    \|X - \mu(s)\|_{\psi_2(\cdot\mid S = s)}
    = \sup\limits_{\|v\| = 1} \|X_v - \mu_v\|_{\psi_2(\cdot\mid S=s)}
    \le 6R + 6\sigma .
\end{align*}
This completes the proof.

\myendproof

\section{Auxiliary results}

\begin{Lem}
\label{lem:potential_lower_bound}
    Grant Assumptions \ref{asn:linear_subspace} and \ref{asn:bounded_log_pot}, and assume \eqref{eq:der_bt_cond}. Let us choose $\nu_{||}^*$ as in Lemma \ref{lem:log_pot_decomp}. Then
    \[
        \cT_\infty[\log\nu_{||}^*] = 0
    \]
    and, for every $z \in \R^d$,
    \[
        -\frac{\|z\|^2}{2\sdata^2}-B
        \leq\log\nu_{||}^*(z)\leq M,
        \quad
        \text{where}
        \quad
        B \lesssim \frac{d\tau^2}{\sdata^2}
        + d \bigl| \log(\gamma\sigma_T^2) \bigr| + d
    \]
    and the hidden constant is absolute.
\end{Lem}

\begin{Lem}
\label{lem:gaus_expect}
    Let $k\geq2$ be an integer and let $Z_1, \dots, Z_k$ be independent centered random vectors in $\R^d$ satisfying $\|Z_j\|_{\psi_2} \leq 1$ for all $j \in \{1, \dots, k\}$. Then there exists an absolute constant $C \geq 1$ such that for any $i_1, \dots, i_k \in \{1, \dots, d\}$ the following inequality holds:
    \begin{align*}
        \E \left| (Z_{1, i_1} - Z_{2, i_1}) (Z_{1, i_2} - Z_{2, i_2}) \dots \left( \sum_{j = 1}^{k - 1} Z_{j, i_k} - (k - 1) Z_{k, i_k} \right) \right|
        \leq \left( 6 \sqrt{C}\right)^k k!.
    \end{align*}   
\end{Lem}

\subsection{Proof of Lemma \ref{lem:potential_lower_bound}}
\label{sec:lem_potential_lower_bound_proof}
 
The normalization $\cT_\infty[\log\nu_{||}^*] = 0$ follows from Lemma \ref{lem:log_pot_decomp}.
Let $G_\perp$ have orthonormal columns spanning the orthogonal complement of $\operatorname{Im}(G)$. Average its decomposition at $y=Gz+G_\perp v$ over
$v\sim\cN(0,\gamma\sigma_\infty^2 I_{D-d})$.
The perpendicular term has mean zero, so
$\log\nu_{||}^*(z)\leq M$ by Assumption
\ref{asn:bounded_log_pot}. When $D = d$, the potentials $\nu_T^*$ and $\nu_{||}^*$ coincide, and the bound becomes trivial.

For the lower bound, Lemma \ref{lem:log_pot_decomp} gives
\begin{equation}
\label{eq:potlb_par_id}
    -\log\nu_{||}^*(z)
    =-\log\sfp_{||}(z)
      +\log\int_{\R^d}
       \frac{\sfq_{||}(z\mid x)\rho_{||}(x)}
            {\cT_{||}[\nu_{||}^*](x)}\,\dd x.
\end{equation}
Since $\E Z=0$, Jensen's inequality in \eqref{eq:rho_par_def}
yields that
\begin{equation}
\label{eq:potlb_negp_par_final}
    -\log\sfp_{||}(z)
    \leq\frac d2\log(2\pi\sdata^2)
       +\frac{\|z\|^2+\E\|Z\|^2}{2\sdata^2}.
\end{equation}
On the other hand, Jensen's inequality also implies that $\log\cT_{||}[\nu_{||}^*](x)$ is not less than minus $\KL$-divergence between the Gaussian measures $\cN(0, \gamma \sigma_\infty^2 I_d)$ and $\cN(e^{-bT} x, \gamma \sigma_T^2 I_d)\bigr)$. Computing this $\KL$-divergence and taking into account the normalization from  Lemma \ref{lem:log_pot_decomp}, we obtain that
\begin{align*}
    \log\cT_{||}[\nu_{||}^*](x)
    &
    \geq -\frac d2\left\{\frac{e^{-2bT}}{1-e^{-2bT}}
    +\log(1-e^{-2bT})\right\}
    -\frac{e^{-2bT}\|x\|^2}{2\gamma\sigma_T^2}
    \\&
    \geq -d e^{-2bT} -\frac{e^{-2bT} \|x\|^2}{2\gamma \sigma_T^2}.
\end{align*}
Here \eqref{eq:der_bt_cond} implies $e^{-2bT}\leq1/2$.
Completing the square therefore yields the following. Combining the lower bound
for $\log \cT_{||}[\nu_{||}^*]$ with $\rho_{||}(x)=(2\pi)^{-d/2}e^{-\|x\|^2/2}$, as given in \eqref{eq:rho_par_def}, and
writing $r=e^{-bT}$, we obtain
\begin{align*}
    &\int_{\R^d}\frac{\sfq_{||}(z\mid x)\rho_{||}(x)}
    {\cT_{||}[\nu_{||}^*](x)}\,\dd x
    \\&
    \leq e^{d e^{-2bT}}(2\pi \gamma\sigma_T^2)^{-d/2}
    \int_{\R^d}(2\pi)^{-d/2}
    \exp\Bigl\{-\frac{\|z-rx\|^2}{2\gamma\sigma_T^2}-\frac{\|x\|^2}{2}
    +\frac{r^2\|x\|^2}{2\gamma\sigma_T^2}\Bigr\}\,\dd x .
\end{align*}
Expanding $\|z-rx\|^2=\|z\|^2-2rz^\top x+r^2\|x\|^2$, the two terms
quadratic in $x$ cancel, and the exponent reduces to
\[
    -\frac{\|z\|^2}{2\gamma\sigma_T^2}+\frac{rz^\top x}{\gamma\sigma_T^2}-\frac{\|x\|^2}{2}
    =-\frac{\|z\|^2}{2\gamma\sigma_T^2}+\frac{r^2\|z\|^2}{2(\gamma\sigma_T^2)^2}
    -\frac12\Bigl\|x-\frac{rz}{\gamma\sigma_T^2}\Bigr\|^2 ,
\]
which is the completed square. Since the Gaussian density
$(2\pi)^{-d/2}\exp\{-\tfrac12\|x-rz/(\gamma\sigma_T^2)\|^2\}$ integrates to one,
\begin{align*}
    \int_{\R^d}\frac{\sfq_{||}(z\mid x)\rho_{||}(x)}
    {\cT_{||}[\nu_{||}^*](x)}\,\dd x
    &\leq e^{d e^{-2bT}}(2\pi \gamma\sigma_T^2)^{-d/2}
    \exp\left\{-\frac{\|z\|^2}{2\gamma\sigma_T^2}+\frac{r^2\|z\|^2}{2(\gamma\sigma_T^2)^2}\right\}
    \\
    &= e^{d e^{-2bT}}(2\pi \gamma\sigma_T^2)^{-d/2}
    \exp\left\{-\frac{(\gamma\sigma_T^2-e^{-2bT})\|z\|^2}{2(\gamma\sigma_T^2)^2}\right\}
    \\&
    \leq e^{d e^{-2bT}}(2\pi \gamma\sigma_T^2)^{-d/2},
\end{align*}
because $\gamma\sigma_T^2\geq e^{-2bT}$ by \eqref{eq:der_bt_cond}.
Combining this estimate with \eqref{eq:potlb_par_id} and
\eqref{eq:potlb_negp_par_final}, we obtain that
\[
 -\log\nu_{||}^*(z)
 \leq\frac{\|z\|^2+\E\|Z\|^2}{2\sdata^2}
       +\frac d2\log\frac{\sdata^2}{\gamma\sigma_T^2}
       +d e^{-2bT}.
\]
The assertion follows from $\E\|Z\|^2\lesssim d\tau^2$ and $0<\sdata<1$.

\myendproof

\subsection{Proof of Lemma \ref{lem:gaus_expect}}
\label{sec:lem_gaus_expect_proof}

Let us fix an arbitrary multi-index $(i_1, \dots, i_k) \in \{1, \dots, d\}^k$. Due to the triangle inequality, it holds that
\begin{align}
    \label{eq:deriv_expect_reduct}
    &
    \E \left| (Z_{1, i_1} - Z_{2, i_1}) (Z_{1, i_2} - Z_{2, i_2}) \dots \left( \sum_{j = 1}^{k - 1} Z_{j, i_k} - (k - 1) Z_{k, i_k} \right) \right|
    \\&\notag
    \leq \E \sum_{(\omega_0, \dots, \omega_{k-1}) \in \{0, 1\}^{k}} |Z_{1, i_1}|^{\omega_0} |Z_{2, i_1}|^{1 - \omega_0} \cdot \prod_{j = 1}^{k - 1} \left( \left|\sum_{s = 1}^{j} Z_{s, i_{j + 1}}\right|^{\omega_j} \cdot j^{1 - \omega_j} \cdot |Z_{j + 1, i_{j + 1}}|^{1 - \omega_j} \right).
\end{align}
For a fixed $(\omega_0, \dots, \omega_{k-1}) \in \{0, 1\}^k$, by the H\"older inequality, we have that
\begin{align}
\label{eq:deriv_omega_term_aux}
    &
    \E |Z_{1, i_1}|^{\omega_0} |Z_{2, i_1}|^{1 - \omega_0} \cdot \prod_{j = 1}^{k - 1} \left( \left|\sum_{s = 1}^{j} Z_{s, i_{j + 1}}\right|^{\omega_j} \cdot j^{1 - \omega_j} \cdot |Z_{j + 1, i_{j + 1}}|^{1 - \omega_j} \right)
    \\&\notag
    \leq \prod_{j = 1}^{k - 1} j^{1 - \omega_j} \left(\E |Z_{1, i_1}|^{3 \omega_0} |Z_{2, i_1}|^{3 - 3 \omega_0} \right)^{1/3}  \left(\E \prod_{j = 1}^{k - 1} |Z_{j + 1, i_{j + 1}}|^{3(1 - \omega_j)} \right)^{1/3}
    \left( \E \prod_{j = 1}^{k - 1} \left|\sum_{s = 1}^{j} Z_{s, i_{j + 1}}\right|^{3\omega_j} \right)^{1/3}
    \\&\notag
    = \prod_{j = 1}^{k - 1} j^{1 - \omega_j} \left(\E |Z_{1, i_1}|^{3 \omega_0} \E |Z_{2, i_1}|^{3 - 3 \omega_0} \right)^{1/3}  \left( \prod_{j = 1}^{k - 1} \E |Z_{j + 1, i_{j + 1}}|^{3(1 - \omega_j)} \right)^{1/3}
    \left( \E \prod_{j = 1}^{k - 1} \left|\sum_{s = 1}^{j} Z_{s, i_{j + 1}}\right|^{3\omega_j} \right)^{1/3}.
\end{align}
Using properties of sub-Gaussian random variables (see, for instance, \cite[Proposition 2.5.2]{vershynin18}), we obtain that 
\[
    \left( \E |Z_{1, i_1}|^{3 \omega_0} \right)^{1/3} \leq (3C)^{\omega_0 / 2},
    \quad
    \left( \E |Z_{2, i_1}|^{3 - 3 \omega_0} \right)^{1/3} \leq (3C)^{(1 - \omega_0) / 2},
\]
and
\[
    \left( \E |Z_{j + 1, i_{j + 1}}|^{3(1 - \omega_j)} \right)^{1/3} \leq (3C)^{(1 - \omega_j) / 2},
    \quad \text{for all $1 \leq j \leq k-1$},
\]
where $C \geq 1$ is an absolute constant.
Furthermore, let $\Omega=\omega_1+\cdots+\omega_{k-1}$.
When $\Omega=0$, the product involving the partial sums is empty and equals one.
For $\Omega\geq1$, applying H\"older's inequality and using independence and
centering, which imply that $\|Z_{1, i_{j + 1}} + \ldots + Z_{j, i_{j + 1}}\|_{\psi_2} \lesssim \sqrt{j}$, we deduce that
\begin{align*}
    \left( \E \prod_{j = 1}^{k - 1} \left|\sum_{s = 1}^{j} Z_{s, i_{j + 1}}\right|^{3\omega_j} \right)^{1/3}
    &
    =
    \left( \E \prod_{j: \omega_j = 1} \left|\sum_{s = 1}^{j} Z_{s, i_{j + 1}}\right|^{3\omega_j} \right)^{1/3}
    \\&
    \leq \left( \prod_{j: \omega_j = 1} \E  \left|\sum_{s = 1}^{j} Z_{s, i_{j + 1}}\right|^{3\Omega \omega_j} \right)^{1/ (3 \Omega)}
    \\&
    \leq \prod_{j: \omega_j = 1} \big( 3C \Omega j \big)^{\omega_j / 2}
    = (3C \Omega)^{\Omega / 2} \prod_{j = 1}^{k-1} j^{\omega_j / 2}.
\end{align*}
Using the convention $0^0=1$ when $\Omega=0$ and substituting these bounds into \eqref{eq:deriv_omega_term_aux}, we conclude that
\begin{align*}
    &
    \E |Z_{1, i_1}|^{\omega_0} |Z_{2, i_1}|^{1 - \omega_0} \cdot \prod_{j = 1}^{k - 1} \left( \left|\sum_{s = 1}^{j} Z_{s, i_{j + 1}}\right|^{\omega_j} \cdot j^{1 - \omega_j} \cdot |Z_{j + 1, i_{j + 1}}|^{1 - \omega_j} \right)
    \\&
    \leq \sqrt{3C} \; \prod\limits_{j = 1}^{k - 1} \left( j^{1 - \omega_j} \cdot (3C)^{(1 - \omega_j) / 2} \right) \cdot (3C \Omega)^{\Omega / 2} \prod_{j = 1}^{k-1} j^{\omega_j / 2}
    \\&
    = (3C)^{k/2} \sqrt{(k - 1)!} \, \cdot \Omega^{\Omega / 2} \prod\limits_{j = 1}^{k - 1} j^{(1 - \omega_j) / 2}.
\end{align*}
Let us note that the latter term in the right-hand side does not exceed
\[
    \prod\limits_{j = 1}^{k - 1} j^{(1 - \omega_j) / 2}
    \leq \sqrt{\frac{(k - 1)!}{\Omega!}}.
\]
Since $l! \geq (l / e)^l$ for all $l \in \N$, we obtain that
\[
    \Omega^{\Omega / 2} \prod\limits_{j = 1}^{k - 1} j^{(1 - \omega_j) / 2}
    \leq \Omega^{\Omega / 2} \sqrt{\frac{(k - 1)!}{\Omega!}} \leq e^{k / 2} \sqrt{(k - 1)!},
\]
and then
\[
    \E |Z_{1, i_1}|^{\omega_0} |Z_{2, i_1}|^{1 - \omega_0} \cdot \prod_{j = 1}^{k - 1} \left( \left|\sum_{s = 1}^{j} Z_{s, i_{j + 1}}\right|^{\omega_j} \cdot j^{1 - \omega_j} \cdot |Z_{j + 1, i_{j + 1}}|^{1 - \omega_j} \right)
    \leq (3Ce)^{k/2} (k - 1)!
\]
Taking into account that the right-hand side of
\eqref{eq:deriv_expect_reduct} includes $2^k$ terms, we conclude that
\begin{align*}
    \E \left| (Z_{1, i_1} - Z_{2, i_1}) (Z_{1, i_2} - Z_{2, i_2}) \dots \left( \sum_{j = 1}^{k - 1} Z_{j, i_k} - (k - 1) Z_{k, i_k} \right) \right|
    &
    \leq \left( 2 \sqrt{3Ce}\right)^k k!
    \\&
    \leq \left( 6 \sqrt{C}\right)^k k!.
\end{align*}
The proof is finished.

\myendproof

\newpage
\tableofcontents

\newpage
\section*{NeurIPS Paper Checklist}

\begin{enumerate}

\item {\bf Claims}
    \item[] Question: Do the main claims made in the abstract and introduction accurately reflect the paper's contributions and scope?
    \item[] Answer: \answerYes{} 
    \item[] Justification: Main claims made in the abstract and introduction accurately reflect the paper's contributions and scope.

\item {\bf Limitations}
    \item[] Question: Does the paper discuss the limitations of the work performed by the authors?
    \item[] Answer: \answerYes{} 
    \item[] Justification: We discuss the assumptions, their validity and restrictiveness in the main section. Besides that, we provide a separate section for the discussion.

\item {\bf Theory assumptions and proofs}
    \item[] Question: For each theoretical result, does the paper provide the full set of assumptions and a complete (and correct) proof?
    \item[] Answer: \answerYes{} 
    \item[] Justification: All proofs can be found in separate Appendix sections. Additionally one can found proof sketches in the Proof Insights section of the main text.

    \item {\bf Experimental result reproducibility}
    \item[] Question: Does the paper fully disclose all the information needed to reproduce the main experimental results of the paper to the extent that it affects the main claims and/or conclusions of the paper (regardless of whether the code and data are provided or not)?
    \item[] Answer: \answerNA{} 
    \item[] Justification: The main contribution is theoretical. The paper does not include experiments.

\item {\bf Open access to data and code}
    \item[] Question: Does the paper provide open access to the data and code, with sufficient instructions to faithfully reproduce the main experimental results, as described in supplemental material?
    \item[] Answer: \answerNA{} 
    \item[] Justification: The paper does not include experiments requiring code.

\item {\bf Experimental setting/details}
    \item[] Question: Does the paper specify all the training and test details (e.g., data splits, hyperparameters, how they were chosen, type of optimizer) necessary to understand the results?
    \item[] Answer: \answerNA{} 
    \item[] Justification: The paper does not include experiments.

\item {\bf Experiment statistical significance}
    \item[] Question: Does the paper report error bars suitably and correctly defined or other appropriate information about the statistical significance of the experiments?
    \item[] Answer: \answerNA{} 
    \item[] Justification: The paper does not include experiments.

\item {\bf Experiments compute resources}
    \item[] Question: For each experiment, does the paper provide sufficient information on the computer resources (type of compute workers, memory, time of execution) needed to reproduce the experiments?
    \item[] Answer: \answerNA{} 
    \item[] Justification: The paper does not include experiments.
    
\item {\bf Code of ethics}
    \item[] Question: Does the research conducted in the paper conform, in every respect, with the NeurIPS Code of Ethics \url{https://neurips.cc/public/EthicsGuidelines}?
    \item[] Answer: \answerYes{} 
    \item[] Justification: The research conducted in the paper conform with the NeurIPS Code of Ethics.

\item {\bf Broader impacts}
    \item[] Question: Does the paper discuss both potential positive societal impacts and negative societal impacts of the work performed?
    \item[] Answer: \answerNA{} 
    \item[] Justification: There is no societal impact of the work performed. The paper is theoretical, we are unaware of any negative impacts.
    
\item {\bf Safeguards}
    \item[] Question: Does the paper describe safeguards that have been put in place for responsible release of data or models that have a high risk for misuse (e.g., pre-trained language models, image generators, or scraped datasets)?
    \item[] Answer: \answerNA{} 
    \item[] Justification: The paper poses no such risks.

\item {\bf Licenses for existing assets}
    \item[] Question: Are the creators or original owners of assets (e.g., code, data, models), used in the paper, properly credited and are the license and terms of use explicitly mentioned and properly respected?
    \item[] Answer: \answerNA{} 
    \item[] Justification: The paper does not use existing assets.

\item {\bf New assets}
    \item[] Question: Are new assets introduced in the paper well documented and is the documentation provided alongside the assets?
    \item[] Answer: \answerNA{} 
    \item[] Justification: The paper does not release new assets.

\item {\bf Crowdsourcing and research with human subjects}
    \item[] Question: For crowdsourcing experiments and research with human subjects, does the paper include the full text of instructions given to participants and screenshots, if applicable, as well as details about compensation (if any)? 
    \item[] Answer: \answerNA{} 
    \item[] Justification: The paper does not involve crowdsourcing nor research with human subjects.

\item {\bf Institutional review board (IRB) approvals or equivalent for research with human subjects}
    \item[] Question: Does the paper describe potential risks incurred by study participants, whether such risks were disclosed to the subjects, and whether Institutional Review Board (IRB) approvals (or an equivalent approval/review based on the requirements of your country or institution) were obtained?
    \item[] Answer: \answerNA{} 
    \item[] Justification: The paper does not involve crowdsourcing nor research with human subjects.

\item {\bf Declaration of LLM usage}
    \item[] Question: Does the paper describe the usage of LLMs if it is an important, original, or non-standard component of the core methods in this research? Note that if the LLM is used only for writing, editing, or formatting purposes and does \emph{not} impact the core methodology, scientific rigor, or originality of the research, declaration is not required.
    \item[] Answer: \answerNA{}
    \item[] Justification: The core method development in this research does not involve LLMs as any important, original, or non-standard components. We use LLMs for verification of the proofs, grammar check, and text polishing only.

\end{enumerate}

\end{document}